\documentclass[letterpaper, 10 pt, conference]{ieeeconf}  

\IEEEoverridecommandlockouts                              
\usepackage{dsfont}
\usepackage[utf8]{inputenc}
\usepackage[T1]{fontenc}
\usepackage{graphicx}
\usepackage{subfigure}
\usepackage{dcolumn}
\usepackage{bm}
\usepackage{amsmath}
\usepackage{comment} 
\usepackage{amssymb}
\usepackage{mathrsfs} 
\usepackage[normalem]{ulem}
\usepackage{url}
\usepackage{hyperref}
\usepackage{amsmath}
\usepackage{amssymb}
\usepackage{bbold}
\usepackage[dvipsnames]{xcolor}
\usepackage{xspace}

\usepackage[english]{babel} 
\usepackage{caption}
\renewcommand{\thetable}{\arabic{table}}
\AtBeginDocument{%
  
}

\title{\LARGE \bf
Which Neural Network to Choose for Post-Fault Localization, Dynamic State Estimation  \& Optimal Measurement Placement in Power Systems?
}

\author{Andrei Afonin and Michael Chertkov
\thanks{This work was supported by M. Chertkov's seed funding at UArizona.}
\thanks{A. Afonin was with Moscow Institute of Physics and Technology, Dolgoprudny, 141700, Russia (early/Russian version of the paper was reported in AA's MIPT B.Sc. diploma defended in July of 2020) and is currently at
École Polytechnique Fédérale de Lausanne (EPFL), Lausanne, CH 1015, Switzerland (M.Sc. student) 
{\tt\small andrei.afonin@epfl.ch}
}
\thanks{M. Chertkov is the Chair of the Program in Applied Mathematics at the University of Arizona, Tucson, AZ 85721, USA and he is also an adjunct professor at the Skolkovo Institute of Science and Technology, Moscow, 121205, Russia {\tt\small chertkov@arizona.edu}}
}

\begin{document}

\maketitle
\thispagestyle{plain}
\pagestyle{plain}

\begin{abstract}
We consider a power transmission system monitored with Phasor Measurement  Units (PMUs) placed at significant, but not all, nodes of the system. Assuming that a sufficient number  of distinct single-line faults, specifically pre-fault state and (not cleared) post-fault state, are recorded by the PMUs and are  available for training, we, first, design a comprehensive sequence of Neural Networks (NNs) locating the faulty line. Performance of different NNs in the sequence, including Linear Regression, Feed-Forward  NN, AlexNet, Graphical Convolutional NN,  Neural Linear ODE and Neural Graph-based ODE, ordered according to the type and amount of the power flow physics involved, are compared for different levels of observability. Second, we build a sequence of advanced Power-System-Dynamics-Informed and Neural-ODE based Machine  Learning schemes  trained, given pre-fault state, to predict the post-fault  state and  also, in parallel, to estimate system parameters.  Finally, third, and continuing to work with the first (fault localization) setting we design a (NN-based) algorithm which discovers optimal PMU placement.  
\end{abstract}

\section{Machine Learning to Identify a Power System Failure: Setting the Stage}

We consider the following three settings which  are relevant for the transmission level PS monitoring of faults which are not cleared but which are also not system critical, i.e. result in the post-fault transient, typically occurring in the course of 5-20 seconds and leading to  a post-fault steady state which is distinct from the  pre-fault steady  state \footnote{Notice, in passing, that  the three settings  are  most relevant to the post-fault control decisions made by the system operator, which are, however, not discussed  in the manuscript directly.}
\begin{itemize}
	\item (I) Given a set of samples, each consistent of (a) a pre-fault state  and (b) post-fault state, both recorded at the nodes of  the system equipped with PMUs, and (c) faulty line identified/localized -- we aim to find  a function which  {\bf predicts post-fault state} (at  the PMU locations), i.e.  maps a mismatch between (a) and (b), considered as an input, to (c), considered as an output.
	
	\item (II)  The same as above in (I), however not utilizing (c), and aiming at -- finding a {\bf universal dynamical model} which maps (a)  to (b). \footnote{The model is universal in the sense that it stays the same regardless of the location where (c) has occurred. This setting is of a special relevance to the situations where PMU placement is relatively sparse and the faults considered are not severe.}

	\item (III)  Given a limited budget on the number of active PMUs available for the system monitoring (which is normally a fraction of all the system's nodes), and given an algorithm available for (I) above which can be applied to any PMU placement -- to find the {\bf optimal placement} of PMUs.
	
\end{itemize}

While addressing the outlined problems we will not only present a modern algorithmic Machine Learning solution, but also suggest for each of the problems a number of solutions/algorithms organized in a sequence.  The sequence(s) will be ordered, according to the amount and type of  the power flow information used in this particular solution.  Therefore,  first algorithm(s) in the sequence will be PS-physics-agnostic,  that is dealing with the PMU data as if it would be any other data stream. We will see that these algorithms may be quite successful in the regime(s) of routine fault which is not stressing the entire PS too much and/or  in the regime of a very detailed PMU coverage, when all or almost all nodes of the system are monitored. On  the other extreme of the spectrum we will be discussing very demanding regime(s) when either the fault is  severe or observability is very limited, or possibly both. In this high stress regime we expect the PS-agnostic schemes to perform very poorly,  and will thus be focusing on injecting some (not all)  PS-guidance in the algorithm.  In general,  we are interested to build a road map towards fault detection, localization and interpretation, which would help the system operator to have a  choice of a wide variety of tools,  to select one depending on the current operational needs.

Application of  ML to the problems related to localization of faulty line (I) in our list of  the  tasks above and also the challenge of the  PMU placement for better detection were already discussed in \cite{2019WentingLi},  which is thus a starting point for our analysis. This manuscript is linked to \cite{2019WentingLi} in a number of ways, some already mentioned above, but also we continue to work in here with the same data-source and the same model.  
We generate data by Power system toolbox \cite{PST}  and work through out the mansucript with the same exemplary model  -- IEEE 68-bus electrical network -- see Section \ref{sec:exp-static} for details. We apply similar measures of performance, e.g. the  cross-entropy \cite{CE} loss function to solve the classification problem of fault localization in Section \ref{sec:failure_static}, and the Mean Squared Error (MSE) \cite{MSE} loss function as we solve the regression problem of the dynamic state estimation in Section \ref{sec:dynamic} and the classification problem of the optimal PMU placement in Section \ref{sec:placement}.

As mentioned above, in this manuscript we describe Machine Learning (ML) models juxtaposed against each other and experimented with in the following Sections to establish their regime of optimal use. Our aim is four-fold.  First, we want to make description of the models simple and transparent. Second,  we attempt to clarify logic behind the models design/architecture focusing, in particular, on explaining why particular models are chosen to answer the power system learning problems (failure localization and/or state prediction). Third, we build the hierarchy of models,  in the sense that models introduced earlier are used as building blocks to construct more advanced models introduced later in the Section. Finally, fourth, the hierarchy of models will also be gauged and commented on in terms of the level of physics of the underlying power system processes involved in their construction.  

\section{Detection of Failure in the Static Regime}
\label{sec:failure_static}

This Section is split into Subsection as follows. We remind the basic elements of the Machine Learning (ML) architecture and training in Section \ref{sec:Parametrization}. We also use it to set the stage for other learning problems considered in the following. The experimental setup of the manuscript is detailed in Section \ref{sec:exp-static}. Linear Regression (LR), Feed-Forward Neural Network (FF-NN), AlexNet and Graph Convolution Neural Networks (GC-NN) are introduced in Sections \ref{sec:LR},\ref{sec:FFNN},\ref{sec:AlexNet},\ref{sec:AlexNet}. In Section \ref{sec:failure_static_discussion} we present and discuss results of our failure detection experiments with the NNs (and also other NNs related to Neural ODE, described in Section \ref{sec:dynamic}).

\subsection{Parameterization and Training}\label{sec:Parametrization}

A supervised ML-model is introduced as a map of the input,  normally denoted as $x$,  to the output, normally denoted as $y$, which is parameterized by the vector of parameters, $\phi$.  We use notation, $\text{ML}_\phi:x\to y$, emphasize that the ML model is of a general position. In the supervised learning setting,  which we are mainly focusing on in this manuscript, we are given $I$ samples of the input/output data, $i=1,\cdots,I:\ x^{(i)},y^{(i)}$, which we also call incidentally $I$ samples. 

In the \emph{fault localization} classification problem, aiming to detect a failed line, we follow the scheme of \cite{2019WentingLi}. We input a sample vector, $x_{{\cal V}_o}=(x_a|a\in{\cal V}_o)$, with as many components as the number of observed nodes, where ${\cal V}_o$ is the set of observed nodes of the power system network. Here ${\cal V}_0$ is a subset of the set of all nodes of the network, ${\cal V}_0\subset {\cal V}$. The output, $y=(y_{ab}|\{a,b\}\in {\cal E})$ is the vector of the dimensionality equal to the number of power lines in the system (number of edges, ${\cal E}$,  in the  power network,  where each line is connecting two neighboring nodes of the network). Each output vector is sparse, with only one nonzero (unity) element corresponding to the location of the fault.

A popular choice, see e.g. \cite{2019WentingLi} of the loss function for the case of a classification output,  e.g. of the fault localization of interest here, is the so-called Cross-Entropy (CE) loss function \cite{CE}: 
\begin{gather}\label{eq:L-CE}
L_{\mbox{\small CE}}(\phi;{\cal V}_o) = -\frac{1}{I}\sum\limits_{i=1}^I
\sum\limits_{\{a,b\}\in{\cal E}} y_{ab}^{(i)}  \log \left(\text{ML}_{\phi;ab}(x^{(i)}_{{\cal V}_o})\right),
\end{gather} 
where $\text{ML}_{\phi;ab}(x^{(i)}_{{\cal V}_o})=y^{(i)}_{\phi;ab}$ shows the $\{a,b\}\in{\cal E}$ component of the output vector for the $i$-th sample generated by the NN function with the fixed vector of the parameters, $\phi$; 
the sums in Eq.~(\ref{eq:L-CE}) corresponds to averaging over empirical probability associated with $I$ actual (true) observations of the faults at specific locations within the grid.

The process of training the ML model consists in solving the optimization problem
\begin{gather}\label{eq:opt}
\phi_{\text{trained}}({\cal V}_o)\doteq \mbox{arg}\min\limits_{\phi}L_{\mbox{\small CE}}(\phi;{\cal V}_o),
\end{gather}
where $\mbox{arg}\min$ means finding the argument of the minimum with respect to the vector of parameters, $\phi$, and $L_{\mbox{\small CE}}(\phi;{\cal V}_o)$ is defined in Eq.~(\ref{eq:L-CE}). Notice that the result (\ref{eq:opt}) depends on the set of the observed nodes, ${\cal V}_o$. Below, in Section \ref{sec:placement}, we analyze how the quality of learning (localization of the failure) depends on ${\cal V}_o$.

\subsection{Experimental Set Up} \label{sec:exp-static} 

We are conducting our experiments on the ground truth data, $(x,y)$, generated with the Power system toolbox \cite{PST} on the exemplary IEEE 68-bus electrical network, consisting of $n=68$ nodes and $m=87$ lines. 

We follow the supervised learning setup of \cite{2019WentingLi}:
\begin{itemize}
	\item The power network is described in terms of the known, symmetric $(n\times n)$ admittance matrix  with $2*m$ off-diagonal nonzero complex elements. 
	
	\item We limit our analysis to single line failures. To generate the ground truth data we pick the failed line i.i.d. at random from the $m=|{\cal E}|$ options. The fault is permanent (not cleared),  however we assume operating in the so-called $N-1$  safe regime,  with the system stabilized after any of the single-line faults to a new steady state (thus achieved in the regime of the corrected admittance matrix derived from the initial admitance matrix by removing a line,  i.e. forcing the admittance of the corrected line to zero). 
	
	\item Observations, before and after the fault, are available at $|{\cal V}_{o}|$ nodes assumed equipped with Phasor Measurement Units, or an alternative measurement equipment. We consider the cases with [5 \%, 10 \%, 20 \%, 40 \%, 70 \%, 100 \%] of observed nodes. Creating the initial training dataset (this approach will be improved in Section \ref{sec:placement}), we pick the  observed nodes at random. For each setting of the observed nodes we train each of the ML models (yet to be described). We repeat the training protocol $50$ times for each ML model in each case of partial observability and then present the averaged results. 
	
	\item Input (sample): $ x$ is generated by the Power System Toolbox (PST) \cite{PST} according to $x = Y \Delta U $, where $ Y \in C^{n \times s} $ is a $n\times s$, where $n=|{\cal V}|$ and $s=|{\cal V}_o|$, sub-matrix of the full $(n\times n)$ admittance matrix,  $ \Delta U \in C^{s} $ is the complex valued vector of changes, i.e. difference in readings before and after the incident, in the voltage potentials at the observed nodes. Here we assume that $Y$ is known. Notice, that each component of the $x$-vector is complex,  therefore represented in the NN modeling via two real channels.
	
	\item Output (sample): $m=|{\cal E}|$ is the binary vector of the empirical line failure probability, $ \forall \{a,b\}\in{\cal E}:\ y_{ab}\in\{0,1\}, \sum_{\{a,b\}\in{\cal E}} y_{ab}=1$. In our experiments each (of the 50) samples corresponds to a new randomly removed line of the system.
	
\end{itemize}

In the following Subsections we describe our choice of the ML models trained on the ground truth data generated according to the procedure described above. Training consists in minimizing the cross entropy loss function \cite{CE} with respect to the vector of parameters, $\phi$, over 1000 epochs. We use the Adam \cite{kingma2014adam} gradient optimization method. 

\begin{figure}
	\subfigure[\ 100\% observability]{
		\includegraphics[scale=0.2]{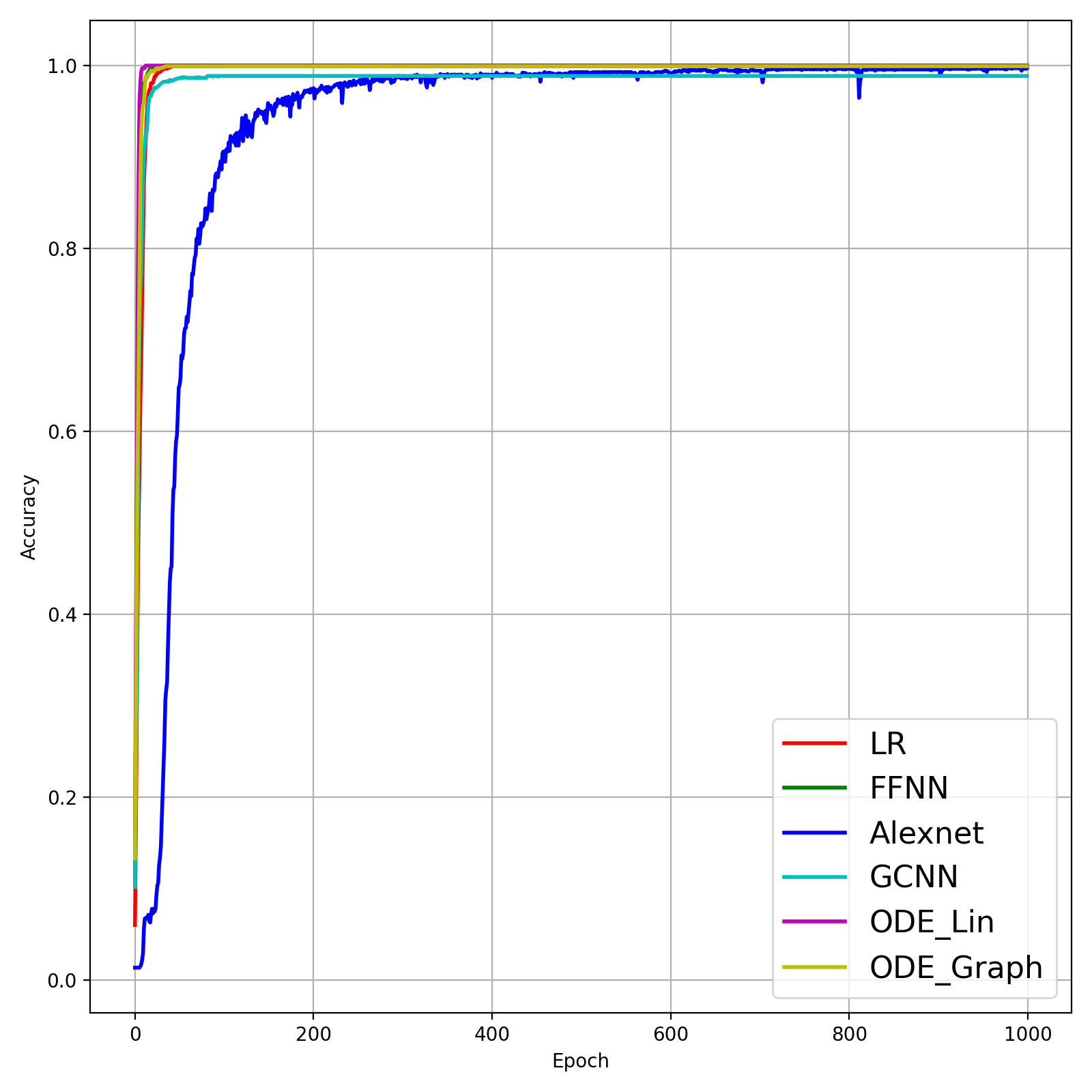}}
	\subfigure[\ 70\% observability]{
		\includegraphics[scale=0.2]{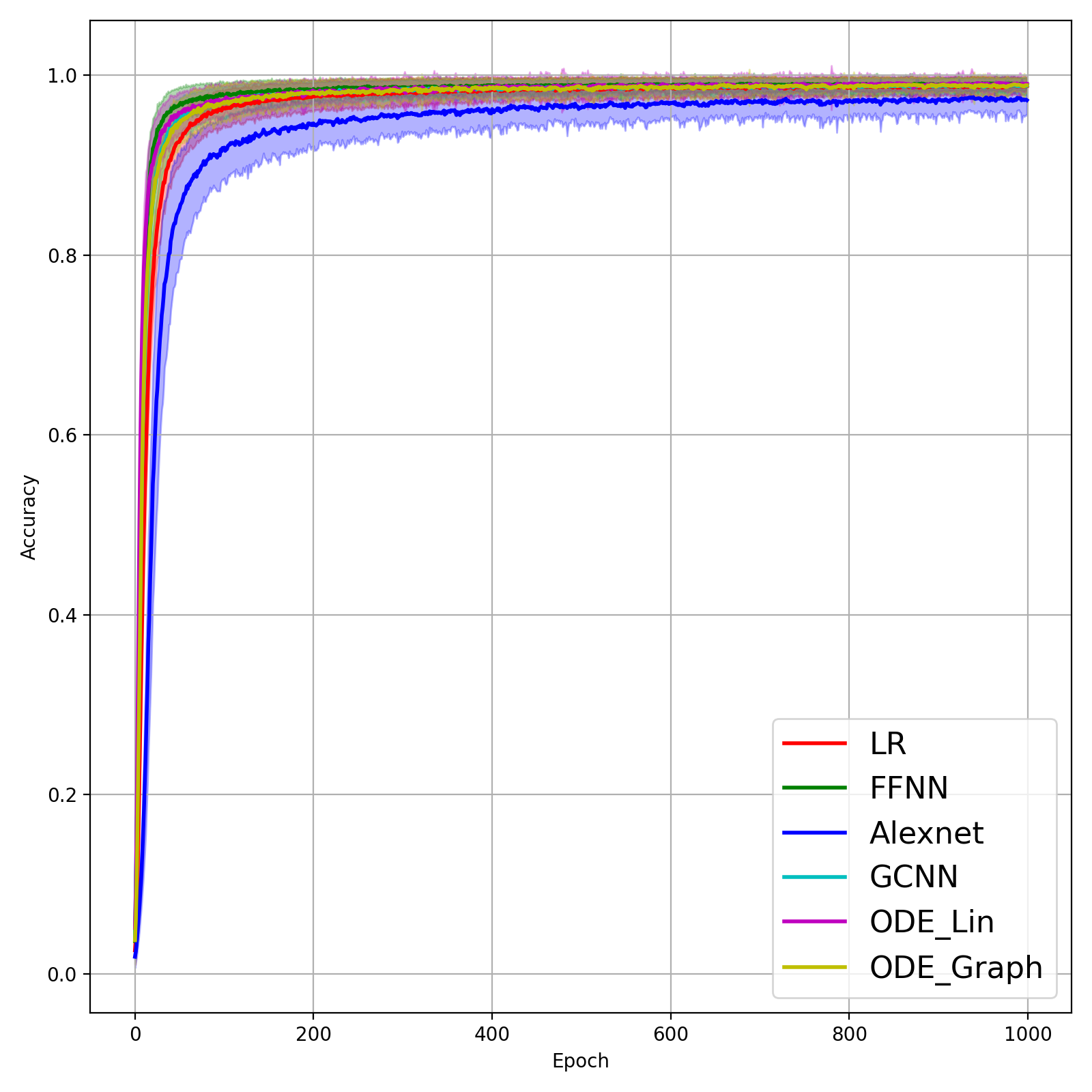}}
	\subfigure[\ 40\% observability]{
		\includegraphics[scale=0.2]{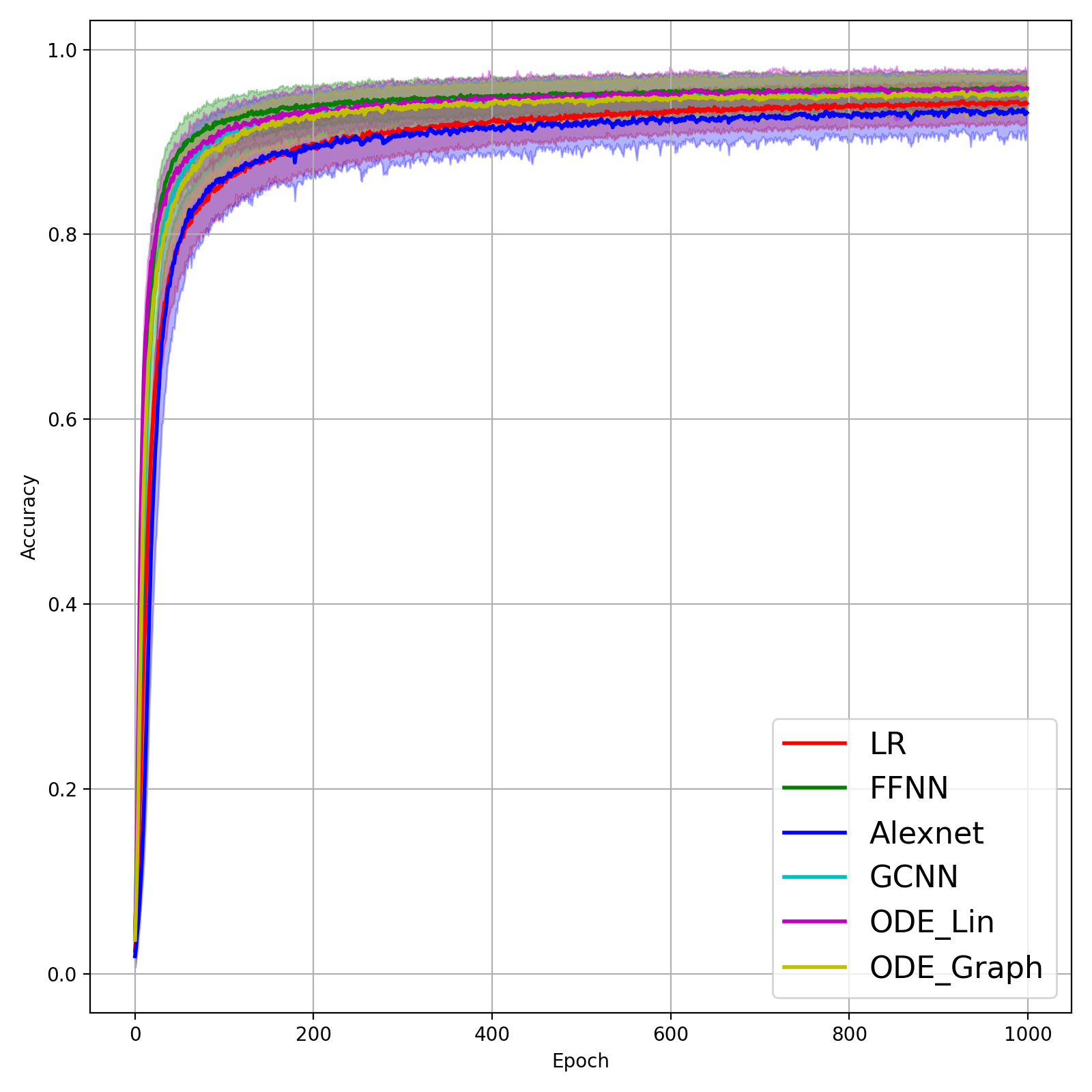}}
	\subfigure[\ 20\% observability]{
		\includegraphics[scale=0.2]{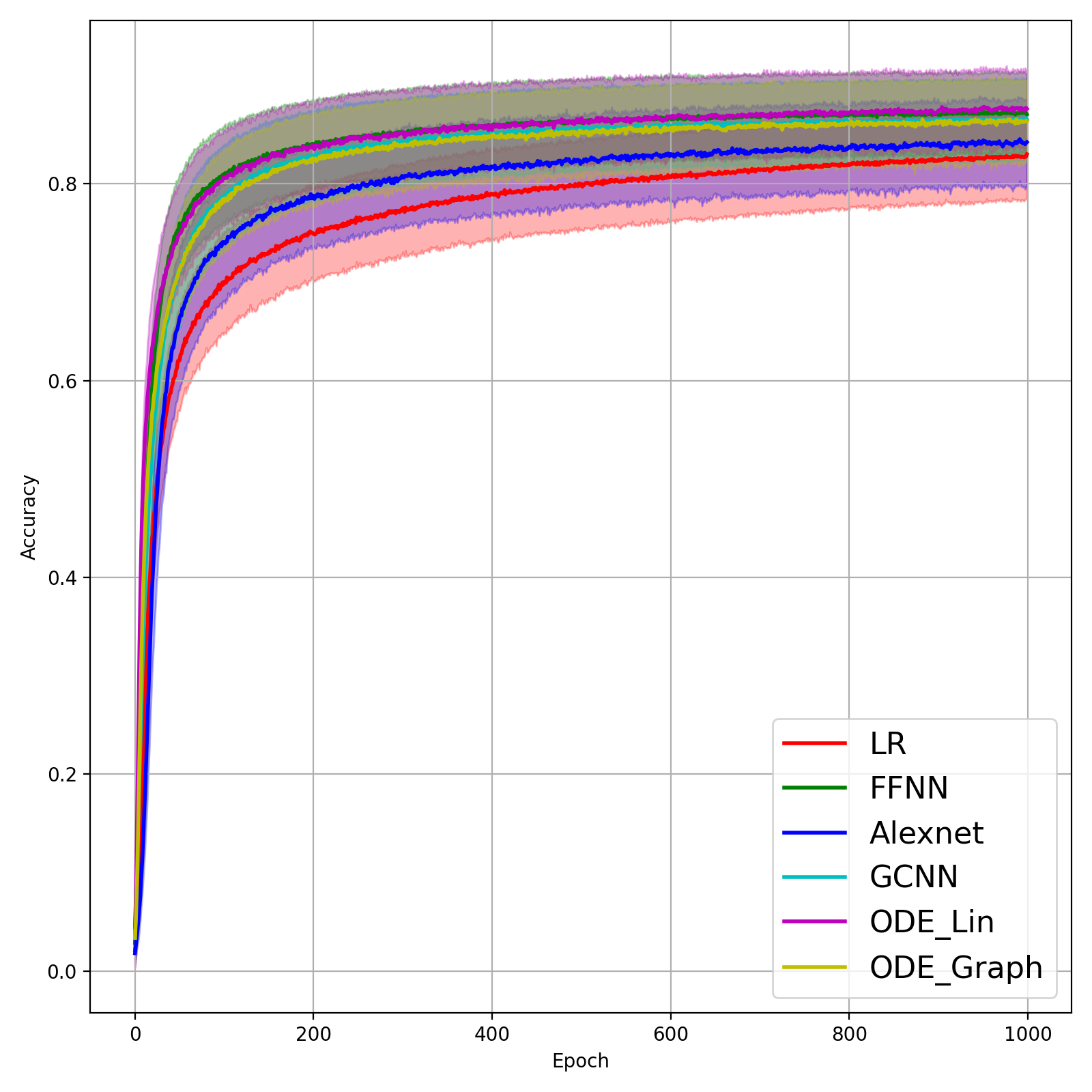}}
	\subfigure[\ 10\% observability]{
		\includegraphics[scale=0.2]{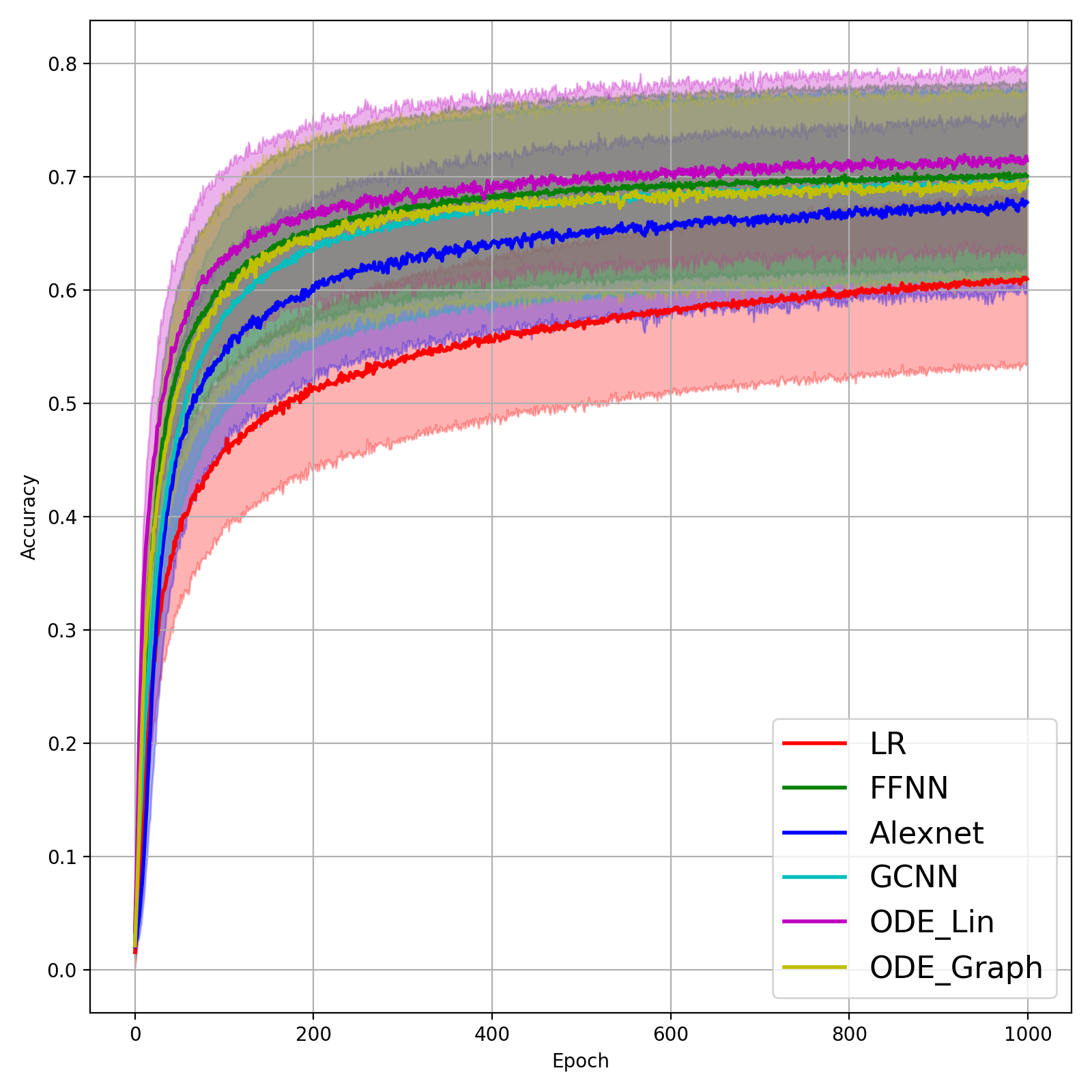}}
	\subfigure[\ 5\% observability]{
		\includegraphics[scale=0.2]{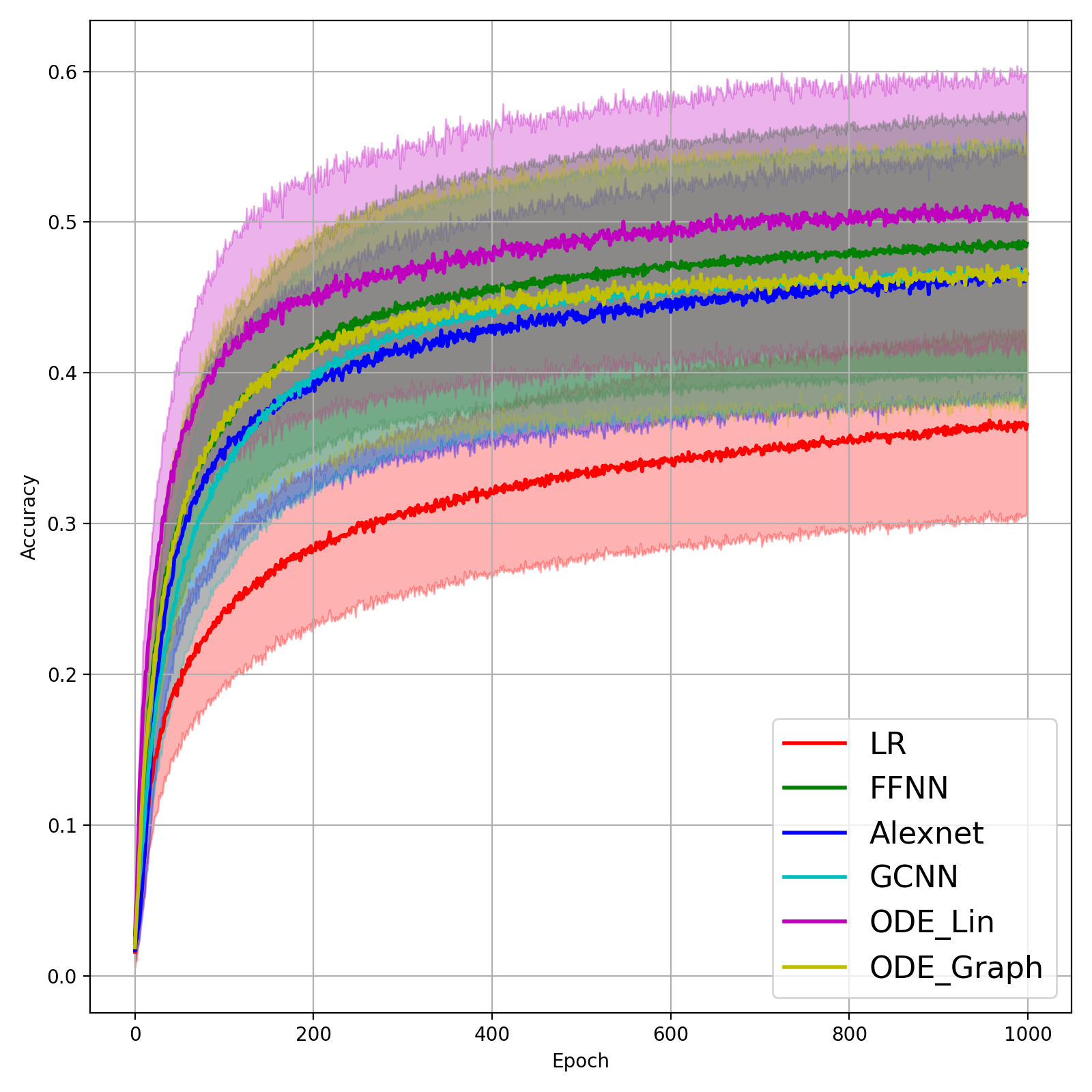}}
	\caption{Comparison of the learning model performance for detection of line failure in the static regime. \label{fig:failure_static}}
\end{figure}

Results of our experiments are shown in the Figure~\ref{fig:failure_static} and the Table~\ref{fig:failure_static}. Discussion of the results is presented in Section \ref{sec:failure_static_discussion}.

\begin{table}[]
	\centering
	\begin{tabular}{ |p{1.35cm}||p{1.2cm}|p{1.2cm}|p{1.2cm}| p{1.2cm}| }
		\hline
		Model & Quality  & Param. & CPU  & GPU\\
		\hline
		LR   & 0.4993  & 6,003 & 0.016 & 0.023\\
		FFNN &   0.6490  & 5,079  & 0.021 & 0.018\\
		AlexNet \cite{2019WentingLi} & 0.6229 & 2,071 & 0.100 & 0.048\\
		GCNN  & 0.6342 & 5,079 & 0.029 & 0.019\\
		ODE Lin &  0.6737 & 10,695 & 1.238 & 1.847\\
		ODE Graph & 0.6398 & 10,695 & 1.284 & 1.791\\
		\hline
	\end{tabular}
	\caption{Summary of the detection failure experiments. The columns show: models; quality of performance under 5 \% observability and 0 \% SNR; number of parameters; time per epoch (in sec), averaged over 1000 epochs for  CPU (Intel(R) Xeon(R) CPU \@ 2.20GHz) and GPU (12GB NVIDIA Tesla K80 GPU), respectively.\label{table:failure_static}}
\end{table}

\subsection{Linear Regression} \label{sec:LR}

Linear Regression (LR) is the simplest ML model, which is the benchmark for comparison in all of our experiments. If it performs well in a regime, other models will not be needed. It is also appropriate to mention that in the case of a small-to-mild perturbation,  power systems are well explained by linear equations (static or dynamic),  therefore providing additional (even though imprecise) legitimacy to the LR. 

Formally,  the LR model, maps input vector, $x \in \mathbb{R}^n$,  to the output vector, $ y \in \mathbb{R}^s$, according to, $y = W x + b$, where $W \in \mathbb{R}^{s \times n}$, $b \in \mathbb{R}^{s}$ are, respectively, multiplicative matrix and additive vector to be learned. $\phi\doteq (W,b)$ append $W$  and $b$ in one vector of parameters. We will also use the following (standard in ML literature) notation for the linear map,  
\begin{gather}\label{eq:LR}
\text{LR}_\phi:\quad  x \rightarrow  y=Wx+b.
\end{gather}

The fault-localization version of the LR-learning consists in solving Eqs.~(\ref{eq:L-CE},\ref{eq:opt}) with the generic function, $\text{ML}_\phi$ substituted by $\text{LR}_\phi$.

\subsection{Feed Forward Neural Network with two layers} \label{sec:FFNN}

The Feed Forward Neural Network (FFNN) with two layers is one of the simplest architectures of nonlinear NN. We use it in the regime of limited observability when we expect that due to severity of the perturbation the LR reconstruction may not be sufficient. FFNN is implemented with Rectified Linear Unit (ReLU) sandwiched by two LR layers:
\begin{gather}\label{eq:ReLu}
\text{FFNN}_\phi:\quad  x \rightarrow \text{LR}_\phi \rightarrow \text{ReLU} \rightarrow \text{LR}_\phi \rightarrow y,
\end{gather}
where $\phi$ is the vector of parameters on the left side is built by appending $W$ and $b$ parameters of the two $LR$ layers on the right (parameters associated with the two layers are independent), and therefore if  $x \in \mathbb{R}^n$ is the input vector and $y \in \mathbb{R}^s$ is output vector (as in the LR case), then $p$ is the dimension of the hidden ReLU layer. (Notice that the ReLU layer is fixed, that is there are no parameters associated with the layer.)

Training of the $\text{FFNN}_\phi$ is, like before in the case of the  $\text{LR}_\phi$, consists in solving Eqs.~(\ref{eq:L-CE},\ref{eq:opt})  with the generic function, $\text{ML}_\phi$ substituted by $\text{FFNN}_\phi$.

\subsection{AlexNet CNN} \label{sec:AlexNet}

AlexNet \cite{AlexNet} is a Convolutional Neural Network (CNN) which was cited most in the ML literature. The network was originally designed for ImageNet Large Scale Visual Recognition Challenge (and won it in 2012). It was later used in many other applications as a starting CNN option, in particular for the real-time faulted line detection reported in \cite{2019WentingLi}. Following \cite{2019WentingLi} we adapt here the classic AlexNet layout.  We use the 13 layer AlexNet CNN to reconstruct line failures. The CNN takes input at the observed nodes and output status of lines (in the form of sparse vector with unity at the position of the failure). The CNN has four convolutional layers and one fully connected layer. Every convolutional layer consists of the convolution sub-layer and the max-pooling sublayer. 
Training of the network 
to localize fault  consists in solving Eqs.~(\ref{eq:L-CE},\ref{eq:opt})  with the generic function, $\text{ML}_\phi$ substituted by the $\text{AlexNet}_\phi$.

\subsection{Graph Convolutional Neural Network}\label{sec:GCNN}

Graph Convolutional Neural Network (GCNN) is a NN which we build making relations between variables in the (hidden) layers based on the known graph of the power system. In this regards, GCNN is informed, at least in part, about physical laws and controls associated with the power system operations. Specifically,  we utilize in constructing GCNN, a sparse $n\times n$ matrix, $|Y|$, built from the absolute values of the impedances associated with power lines connecting $n$ nodes of the system to each other. (The matrix is sparse because degree of a typical node in a transmission level power system is somewhere in the $1-4$ range.) We follow construction of \cite{kipf2016} and use, $Y$, to build the convolutional layer of the GCNN. Let $ H $ be the input vector to the graph convolutional layer, then the output $f(H, A)$ of such a layer is $f(H, A) = \sigma\left(D^{-\frac{1}{2}}AD^{-\frac{1}{2}}H W\right)$, where  $W $ is a matrix of parameters; $A = |Y| + I$, where $I$ is the unit matrix and $ \sigma() $ is a nonlinear activation function. We normally use $ \text{ReLU}() $ for $\sigma()$. $D$ is the diagonal matrix built from the vector of the node degrees within the power system graph. $D^{-\frac{1}{2}}$ stands for the matrix derived from $D$ by taking components-wise inverse square-root.  We use $\text{GC}_\phi$ for the GC operation where $\phi$ denotes all the parameters needed to describe graph convolution map from  $n$-dimensional input to $p$-dimensional vector representing the hidden layer. With a minor abuse of notations the resulting map becomes:
\begin{gather}\label{eq:GraphConv}
\text{GCNN}_\phi:\quad  x \rightarrow \text{GC}_\phi \rightarrow \text{ReLU} \rightarrow \text{LR}_\phi\rightarrow y,
\end{gather}
where $x \in \mathbb{R}^n$ is the input vector to our model, $y \in \mathbb{R}^s$ is its output, $\text{GC}_\phi(x)$ is the $p$-dimensional vector of the intermediate layer $p$. As always, two independent vectors of parameters on the right hand side of Eq.~(\ref{eq:GraphConv}) are appended into the resulting vector of the parameters on the right hand side of Eq.~(\ref{eq:GraphConv}). 

Training of the $\text{GCNN}_\phi$ to localize fault  consists in solving Eqs.~(\ref{eq:L-CE},\ref{eq:opt})  with the generic function, $\text{ML}_\phi$ substituted by the $\text{GCNN}_\phi$.

\subsection{Discussion of Results} \label{sec:failure_static_discussion}

We have tested performance of all the models introduced so far on the problem of fault detection on the experimental setup described in Section \ref{sec:exp-static}. We also included in this comparative study experiments with two other models introduced below. Specifically, in Section \ref{sec:ODE} we have adapted two Neural ODE model,  which both have a broader applicability, to the case study of the fault detection in the case of the static observations. 

Results of the 100 experiments with different randomly initialized parameters for each model are shown in Fig.~(\ref{fig:failure_static}). Bold lines show mean accuracy curves for each model. We observe that in general Linear ODE model performs better than other models. Also, all models outperform linear regression in low-observability regimes. Finally, our proposed models outperform AlexNet based model, which was suggested for the problem in \cite{2019WentingLi}. We also observed that performance of the models depends dramatically on where the measurement (PMU) devices are placed. This observation motivated material of Section \ref{sec:placement} discussing NN approach to the optimal placement of PMUs.  

\section{Dynamic Models}\label{sec:dynamic}

This Section is devoted to introduction and discussion of the Dynamic Models, transitioning to the models gracefully from the topic (of fault localization) discussed in the preceding Section. We start with the discussion of a generic, and thus Power System (PS)- physics-agnostic Neural ODE model in Section \ref{sec:ODE}, and then start to add the PS-physics in steps progressing to the Physics-Informed Neural Networks (PINN) in Section \ref{sec:PINN}, to the Hamiltonian Neural Networks (HNN) in Section 
\ref{model:HNN}, and finally to the Direct ODE NN based on the Swing Equations in Section
\ref{sec:Swing}.

\subsection{Neural ODE}\label{sec:ODE}

Neural ODE is a modern NN method suggested in \cite{chen2018neural}. It builds an input-to-output map as if it would come from the temporal dynamics governed by the parameterized ODE, 
\begin{gather}\label{eq:ODE}
t\in [0,T]:\ \frac{dx(t)}{dt} =  f_\phi(x(t)),
\end{gather}
where $\phi$ is a (possibly time-dependent) vector parameterizing the "rhs"  of the ODE,  i.e. $f_{\phi(t)}(x(t))$,  by a NN. It is assumed that an  ODE solver,  taking $f$ as an input,  can be used in a black-box fashion to train the NN. When considered in discrete time Eq.~(\ref{eq:ODE}) becomes, $k=1,\cdots,K,\ t_k=\Delta k,\ \Delta=T/K$:
\begin{gather}\label{eq:ODE-disc}
x(t_{k+1}) = x(t_k) + \Delta f_\phi(x(t_k)),
\end{gather}
where $\Delta$ is the time step. Neural ODE are also naturally linked to the so-called ResNet (Residual Network) architecture discussed in \cite{he2015deep}. Consistently with notations used for other models:
\begin{gather}\label{eq:NeuralODE}
\text{NeuralODE}_\phi:\quad  x(0) \rightarrow x(T),
\end{gather}
where $x(0) \in \mathbb{R}^n$ is the input vector to our model, and $x(T) \in \mathbb{R}^n$ is the output which is of the same dimensionality, $n$, as the input. 
We will work in the following with an LR version of $f_\phi$ and with a Graph CNN version of $f_\phi$ in Eq.~(\ref{eq:ODE}),  then replacing $\text{NeuralODE}$ in Eq.~(\ref{eq:NeuralODE}) by $\text{LinODE}$ and $\text{GraphODE}$ respectively, where $\text{LinODE}$ and $\text{GraphODE}$ means that $f_{\phi(t)}(x(t))$ parameterized by linear layer and graph convolutional layer correspondingly. To make output of the $\text{LinODE}$ and $\text{GraphODE}$ versions of Eq.~(\ref{eq:NeuralODE})  consistent with the output of other (static) models discussed so far we will additionally map $x(T)$ to $y$, as discussed above, inserting additional ReLU function. (We remind that $y$ is the output vector which is, in the training stage, has only one nonzero component correspondent to the faulty line.)  We therefore add, as already discussed in Section \ref{sec:failure_static}, the $\text{LinODE}$ and $\text{GraphODE}$ augmented with the ReLU function to the list of other (static) schemes resolving the task of the failed line localization.

However, we may also consider NeuralODE~(\ref{eq:NeuralODE}) as a part of the Dynamic State Estimation (DSE) scheme. In this case we assume that $x(T)$ is the observed output and then we may train the NeuralODE by minimizing 
\begin{align}
\label{eq:l2-Loss-NeuralODE}
& \text{arg}\min\limits_{\phi}L_{\text{2;NeuralODE}}(\phi),\quad
L_{\text{2;NeuralODE}}(\phi)=\\ & \sum\limits_{i=1}^I \Biggr\|x^{(i)}(T)-\text{NeuralODE}_\phi(x^{(i)}(0))\Biggl\|^2.
\end{align}

Moreover, we may generalize Eq.~(\ref{eq:NeuralODE}) and consider the entire trajectory, which we will also call ``the path", $\{x(t)|t\in ]0,T]\}$, or (more realistically) its discretized version, $\{x(t_k)|k=1,\cdots,K\}$, as the output of the $\text{NeuralODE}_\phi$, i.e. 
\begin{gather}\label{eq:Path-NeuralODE}
\text{Path-NeuralODE}_\phi:\  x(0) \rightarrow \{x(t_k)|k=1,\cdots,K\}.
\end{gather}
Then the exemplary training problem finding the best (functional) map in the path-version of the NeuralODE becomes
\begin{align}
\label{eq:l2-Loss-NeuralODE-Path}
& \text{arg}\!\min\limits_{\phi}L_{\text{2;Path-NeuralODE}}(\phi),\quad L_{\text{2;Path-NeuralODE}}(\phi) =\\ & 
\text{arg}\min\limits_{\phi}\!\! \sum\limits_{i=1}^I\!\! \frac{1}{K}\!\!\sum\limits_{k=1}^K\Biggl\|x^{(i)}(t_k)\!-\!\text{Path-NeuralODE}_\phi(x^{(i)}(0);t_k)\Biggr\|^2.\nonumber
\end{align}

As will be argued in the remaining Subsections of this Section, we may project the formulation of Eqs.~(\ref{eq:Path-NeuralODE},\ref{eq:l2-Loss-NeuralODE-Path}) to the problems of interest to power system dynamics. Specifically, we may consider $x(t)$ corresponding to dynamics of the state of the power system measured as a function of time at the observed mode (e.g. $S(t)$ and/or $V(t)$) in the transient regime. In this case the training data,  i.e. $\{x^{(i)}(t)|i=1,\cdots,I,\ t\in[0,T]\}$,  can be generated by a dynamic power flow solver resolving many more degrees of freedom (at many more nodes) and therefore producing results much slower than the trained Path-NeuralODE reduced model.

\subsection{Physics-Informed Neural Net}\label{sec:PINN}

The structure of the so-called Physics Informed NN (PINN) is described in  \cite{2018Raissi}. It is based on some early ideas on tuning a NN to satisfy the output of a differential equation  \cite{Lagaris_1998}. We are seeking to use in the data fitting a concrete version of the ODE model (\ref{eq:ODE}),  with $f_\phi(x(t))$ replaced by a specified by "physics", $f_\psi(x(t))$, where thus $\psi$ stand for the vector of physics-meaningful (explainable or interpretable) parameters,
\begin{gather}\label{eq:ph-ODE}
\frac{dx(t)}{dt} =  f_\psi(x(t)),
\end{gather}
where $x(t)$ here stands for measurements changing in time $t$. We built a Neural Network, mapping $t$ to $\hat{x}_\phi(t)$.  We aim to search through the space of $\phi$ to minimize the difference between, $\hat{x}_\phi(t)$ and the actual measurements, $x$ at the time $t$. In the PINN of \cite{2018Raissi} the goal is achieved by minimizing the following Loss Function 
\begin{align}\nonumber 
& \mbox{arg}\min\limits_{\phi,\psi} L_{PINN},\quad L_{PINN}(\phi,\psi) = \lambda \sum\limits_{k=1}^K (\hat{x}_\phi(t_k) - x(t_k))^2 +\\ & \sum\limits_{k=1}^K \left(\hat{x}_\phi(t_{k+1})-\hat{x}_\phi(t_k) - \Delta f_\psi(t_k,\hat{x}_\phi(t_k) )\right)^2, \label{eq:PINN-Loss}
\end{align}
over $\phi$- represent the aforementioned NN, and also over $\psi$- which may be represented by a NN, or can also also include some "physical" parameters,  i.e. parameters which allow physical (power system in our case) interpretation \footnote{In the following we will combine $\phi$ and $\psi$ in one set of parameters, where some of the parameters may be physical, i.e. interpretable,  and some, normally represented by a NN, can be physics blind.}; $\lambda $ is a pre-set hyper-parameter; the entire data path, $\{x(t)\}_K=\{t_k,x_k|k=1,\cdots,K\}$, is assumed known.

A number of technical and terminology remarks are in order. First, the vector of physical parameters, which may describe $\psi$ or its part, should be turned to specifics of the power system and this is what will be done below in Sections \ref{model:HNN},\ref{sec:Swing}. 
Second, generalization of the scheme from ODE to PDE is straightforward. In fact, the Burgers PDE was presented as an enabling example in \cite{2018Raissi}.)  Finally, third, let us also mention that the PINN ideas \cite{2018Raissi} are similar to the approach known under the name of Learning Differential Equation (LDE), see e.g. \cite{2010LearningStochasticODE} and references therein, also discussed in the context of learning power system dynamics in \cite{2017Lokhov}. The similarity between the two approaches is in the form of the loss function, including differential equation via the $l_2$-term and also kind of similar, but not identical, $l_1$ regularization term. The difference between the PINN approach of \cite{2018Raissi} and the LDE approach of \cite{2010LearningStochasticODE} is two fold.  On one hand, no NN were used in \cite{2010LearningStochasticODE} to represent unknown functions,  while embedding NN into the scheme is the main novelty of \cite{2018Raissi}. 
On the other hand,  the LDE approach of  \cite{2010LearningStochasticODE} consisted in learning the stochastic differential equations, and specifically unknown physical parameters, $\psi$,  in  $f_\psi(t,u)$ (if we use the extension of PINN) just introduced above in the first remark). The stochastic component revealed itself in \cite{2010LearningStochasticODE} via appearance of the inverse covariance matrix (also called precision or concentration matrix) which may also be considered contributing, in full or partially, the vector of the physics-meaningful training paramaters, $\psi$. finally,  fourth let us also mention that the PINN scheme of \cite{2018Raissi} was adapted to dynamic parameter learning in the power system setting in \cite{2019Spyros}. See also related discussion below in Section \ref{sec:Swing}.

\subsection{Hamiltonian Neural Net}
\label{model:HNN}

As already mentioned above more structure, related to our understanding (or expectation) about the physics of the problem, can be embedded into NeuralODE and PINN.  Specifically, if the underlying ODE is of a conservative (Hamiltonian) type we can construct what is coined in \cite{Zhong2020Symplectic} \cite{zhong2020dissipative} the Hamiltonian NN. However, the system of equations describing the power system dynamics (which are yet to be introduced) are not conservative, therefore suggesting that a more general model than the bare Hamiltonian one, can be appropriate here. It seems  reasonable to consider the dynamical system described by the so-called Port-Hamiltonian system of equations \cite{PortHamilt}:
\begin{equation} \label{eq:port-Hamilt}
\begin{pmatrix} 
\dot{\text{q}} \\ 
\dot{\text{p}}
\end{pmatrix} = 
\left(\begin{pmatrix} 
0 & \text{I}\\ -\text{I} & 0\end{pmatrix}-\text{D}_\phi(p,q)\right) 
\begin{pmatrix} 
\frac{\partial H_\phi(p,q)}{\partial \text{q}} \\ 
\frac{\partial H_\phi(p,q)}{\partial \text{p}}
\end{pmatrix} +
\begin{pmatrix} 
0 \\ 
\text{F}_\phi(p,q)
\end{pmatrix},
\end{equation}
where coordinate vector, $p$, and the momentum vector, $q$, are of the same dimensionality, $m$, $\text{I} $ is the $m\times m$-dimensional identity matrix, $H_\phi(p,q)$ is the Hamiltonian (function), $\text{D}_\phi(q) $ is the symmetric positive-definite $m\times m$ dissipation matrix (function) and $F_\phi(p,q)$ is the source function.  

Obviously one may consider Eq.~(\ref{eq:port-Hamilt}) as a particular case of the general ODE Eq.~(\ref{eq:ODE}) where $x=(p,q)$. Then one can naturally introduce the (port-) Hamiltonian version of the Path-Neural ODE,  substituting $\text{Path-NeuralODE}_\phi$ in Eq.~(\ref{eq:Path-NeuralODE}) by $\text{path-HNN}_{\phi}$,
and then train it minimizing Eq.~(\ref{eq:l2-Loss-NeuralODE-Path} where the respective substitution is also made.

\subsection{Direct ODE NN based on the Swing Equations}
\label{sec:Swing}

A popular model of the power system, extending the static Power Flow equations
to the dynamical case, is the so-called nonlinear and dissipative swing equations governing dynamics of  the phase of the voltage potential, $\theta_a(t)$,
\begin{align}\nonumber 
& \forall a\in{\cal V}:\ m_a\ddot{\theta}_a+d_a \dot{\theta}_a=P_a-\!\!\!\!\!\!\sum\limits_{b\in{\cal V};\{a,b\}\in{\cal E}}\!\!\!\!\!\beta_{ab}v_a v_b\sin(\theta_a\!-\!\theta_b)\\ & -\sum\limits_{b\in{\cal V};\{a,b\}\in{\cal E}}g_{ab}v_a \left(v_a-v_b\cos(\theta_a-\theta_b)\right),
\label{eq:nonlinear-swing}
\end{align}
where $v_a$ is the absolute value of the voltage potential at a node, $a$, $\beta_{ab}$ and $g_{ab}$ are susceptance and conductance of the line $\{a,b\}$ defined as imaginary and real parts of the line admitance, $Y_{ab}=g_{ab}+i\beta_{ab}$; and $m_a$ and $d_a$ are inertia and the so-called droop coefficients at the node $a$ \footnote{In the DIRODENN scheme described below the $g=0$ version of the Eqs.~(\ref{eq:nonlinear-swing}), ignoring power line resistance,  was actually implemented.}.  

Normally (in the power system literature) the power network and line characteristics in Eq.~(\ref{eq:nonlinear-swing}) correspond to the actual physical lines, then line parameters, $m,d$, are dynamic physical parameters associated with devices (generators and loads) inertia and damping (also frequency control), respectively, while $\beta,g$ are static physical parameters of the respective lines and devices.  Here, we adapt this physical picture to a reduced model of power systems.  This adaptation is obviously blurred by limited observability. We assume, extending the static consideration of \cite{pagnier2021physicsinformed}, that Eq.~(\ref{eq:nonlinear-swing}) also applies to a reduced set of nodes where the PMU devices are located and measurements are available. In this setting we consider a complete graph connected the observed nodes and do not assume that the (effective) nodal and line parameters are known --- instead we aim to learn the effective parameters. Notice also that Eqs.~(\ref{eq:nonlinear-swing}) can be viewed as a particular, that is more structured, version of the port-Hamiltonian system of Eqs.~(\ref{eq:port-Hamilt}). Here, like in the case of Path-HNN,  we introduce the 
Direct ODE NN (DIRODENN) version of the Path-Neural ODE, substituting $\text{Path-NeuralODE}_\phi$ in Eq.~(\ref{eq:Path-NeuralODE}) by $\text{Path-DIRODENN}_\phi$,
and then train it minimizing Eq.~(\ref{eq:l2-Loss-NeuralODE-Path}) where the respective substitution is also made.

\section{Dynamic State Estimation} 
\label{sec:state_est_dynamic}

\begin{figure}
	\subfigure[\ 100\% observability]{
		\includegraphics[scale=0.2]{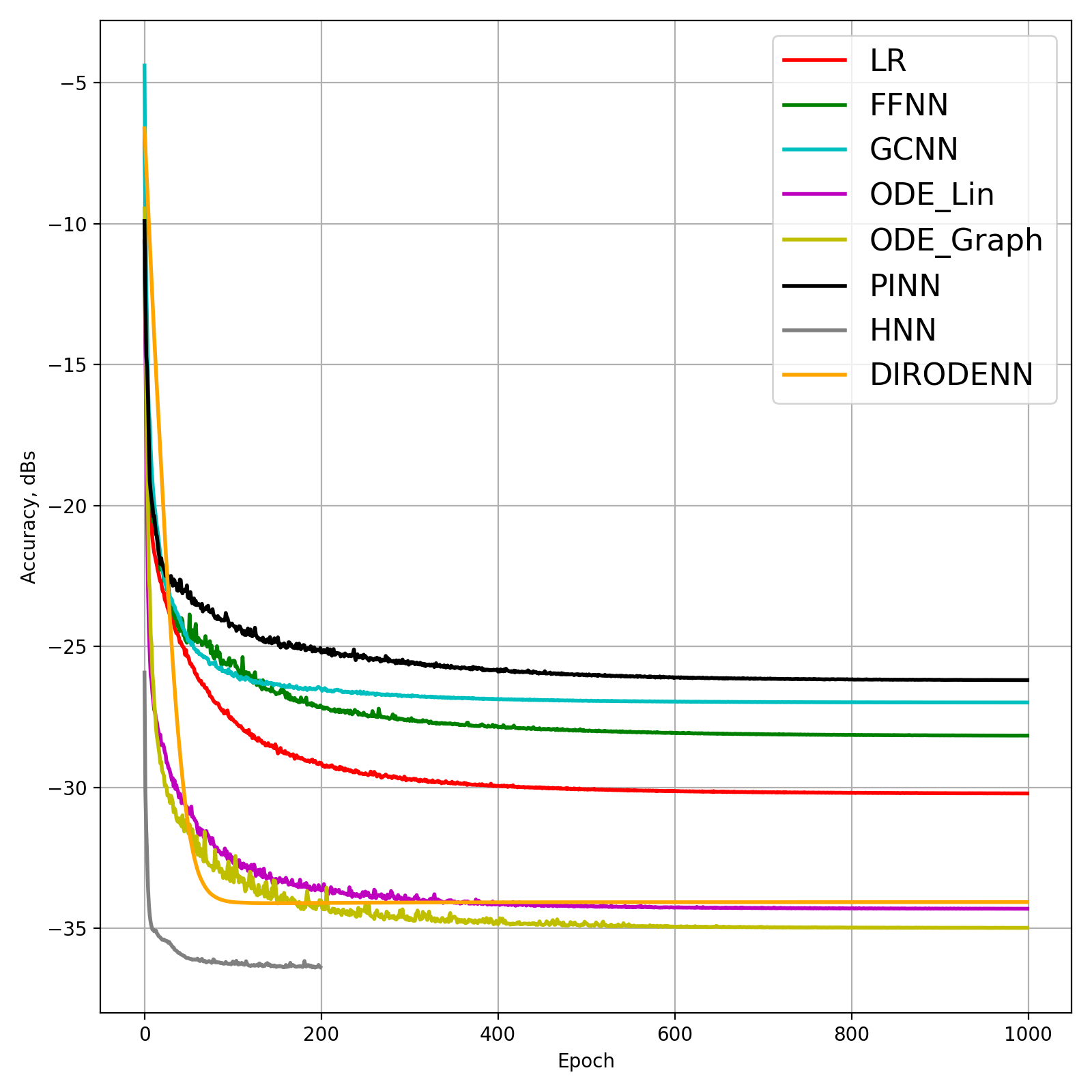}}
	\subfigure[\ 70\% observability]{
		\includegraphics[scale=0.2]{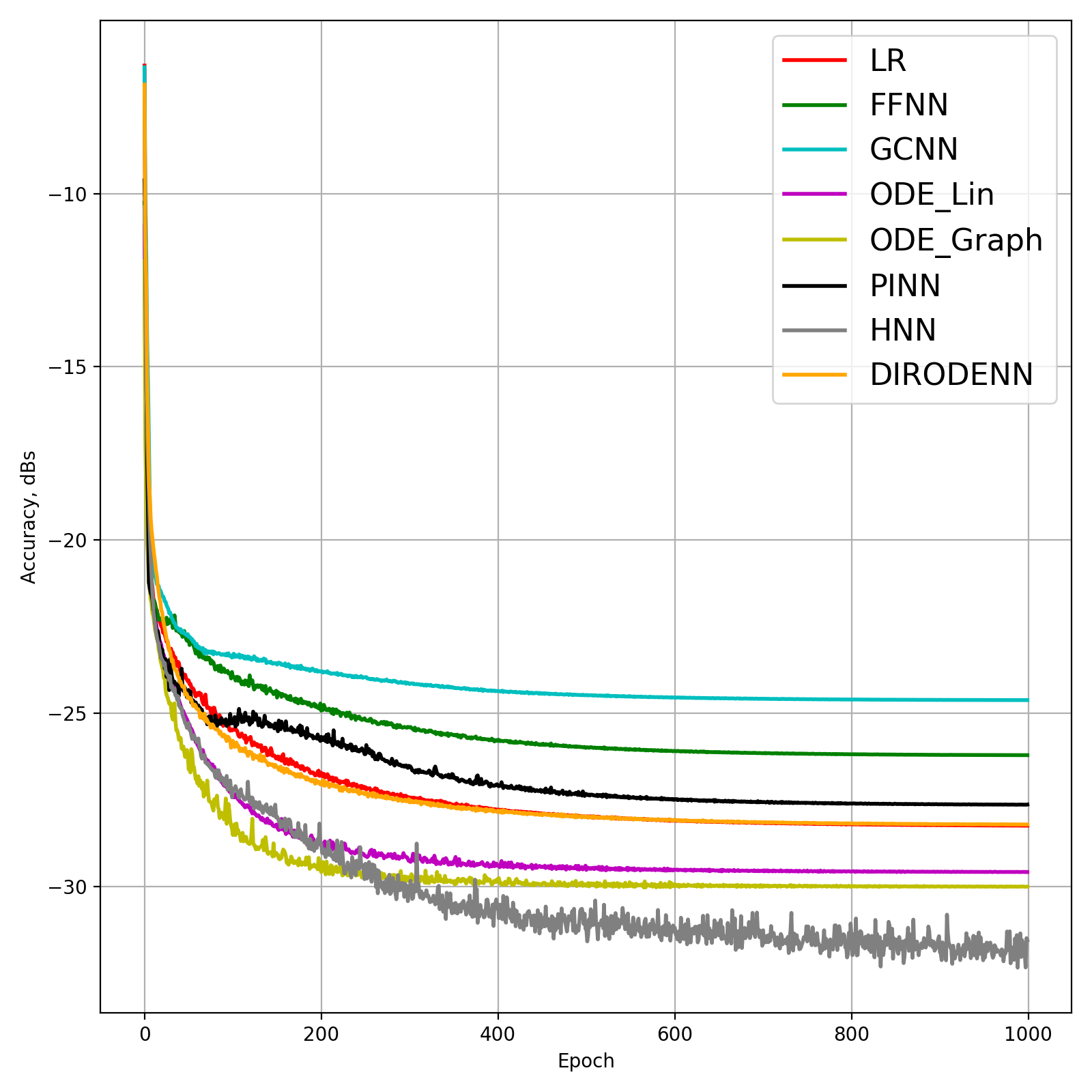}}
	\subfigure[\ 40\% observability]{
		\includegraphics[scale=0.2]{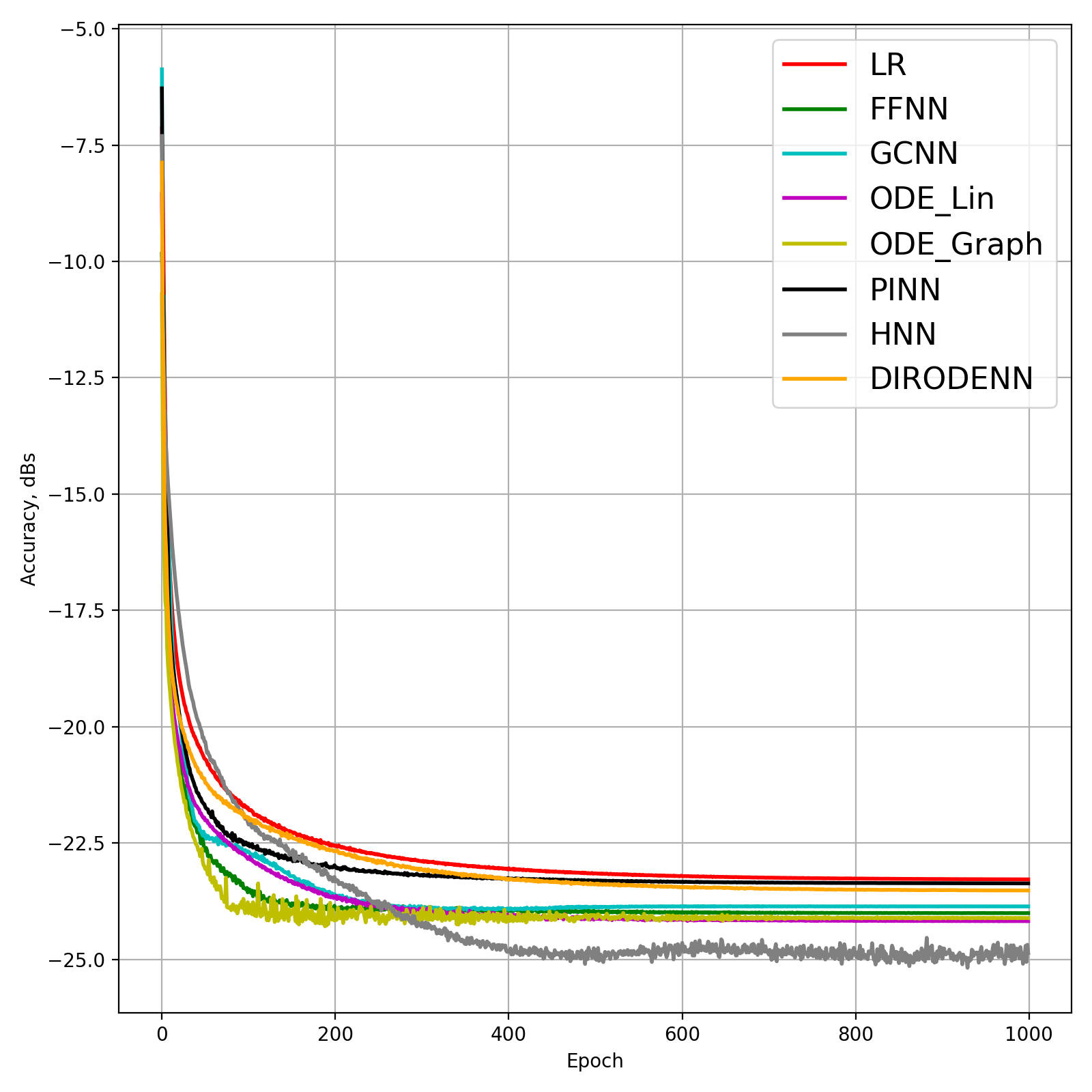}}
	\subfigure[\ 20\% observability]{
		\includegraphics[scale=0.2]{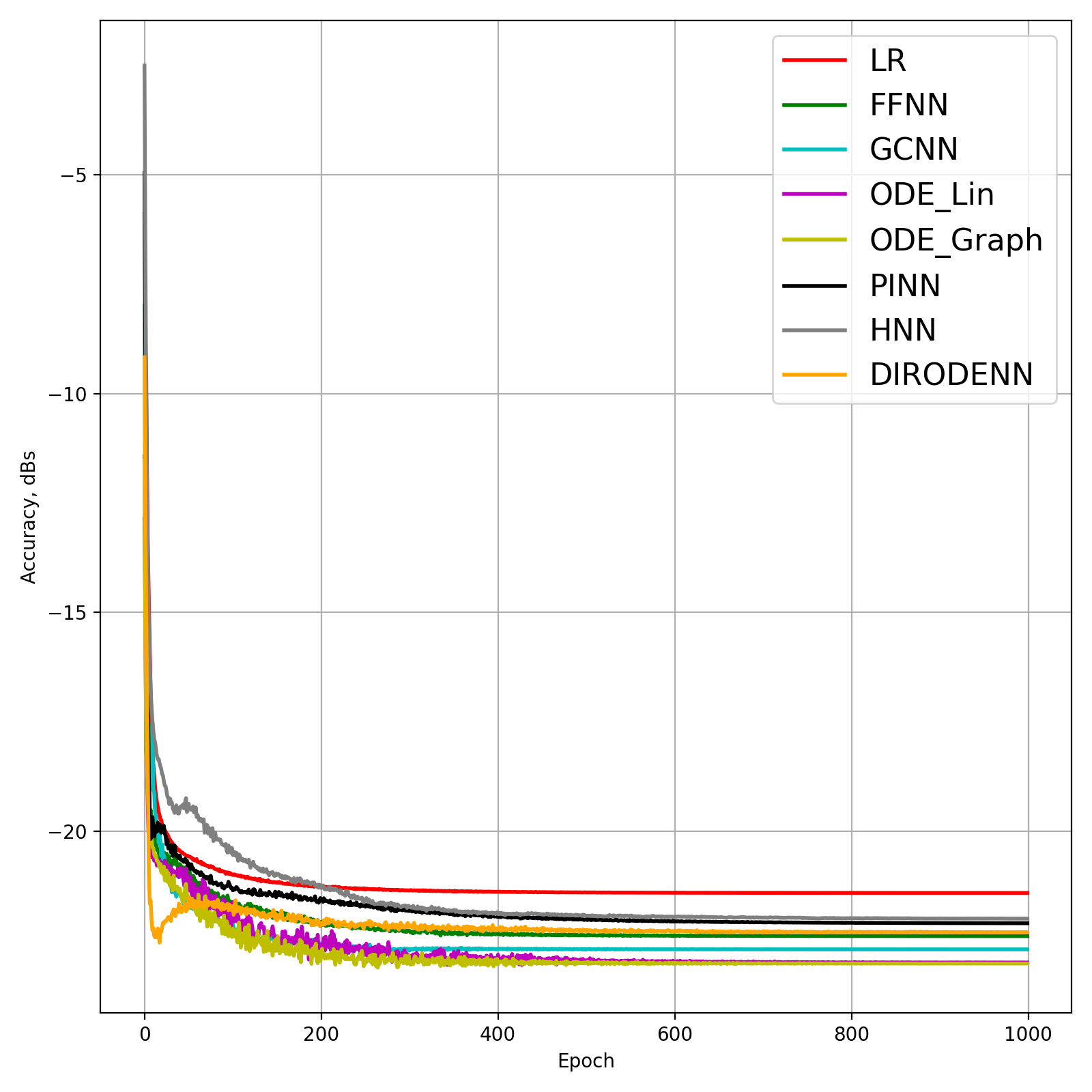}}
	\subfigure[\ 10\% observability]{
		\includegraphics[scale=0.2]{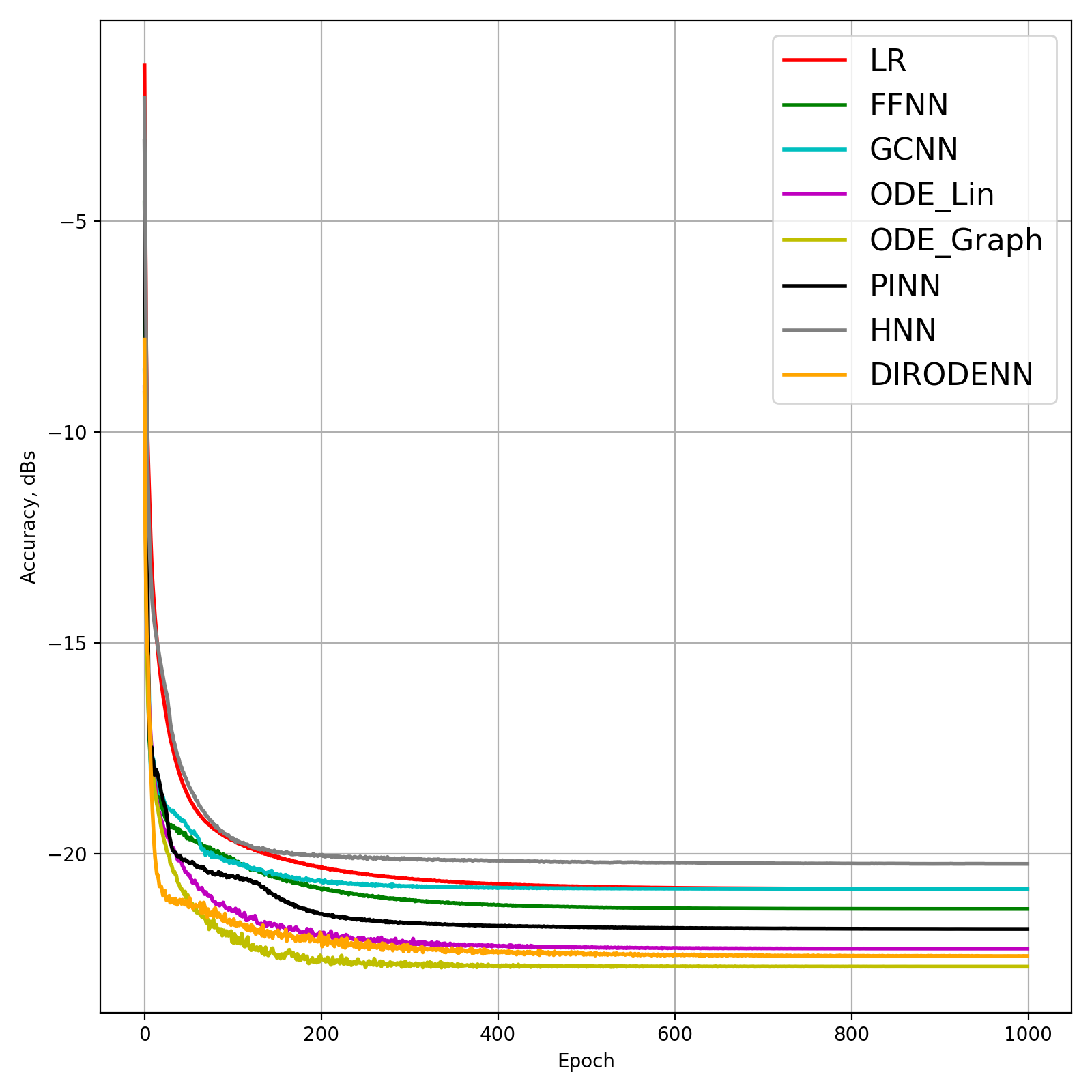}}
	\subfigure[\ 5\% observability]{
		\includegraphics[scale=0.2]{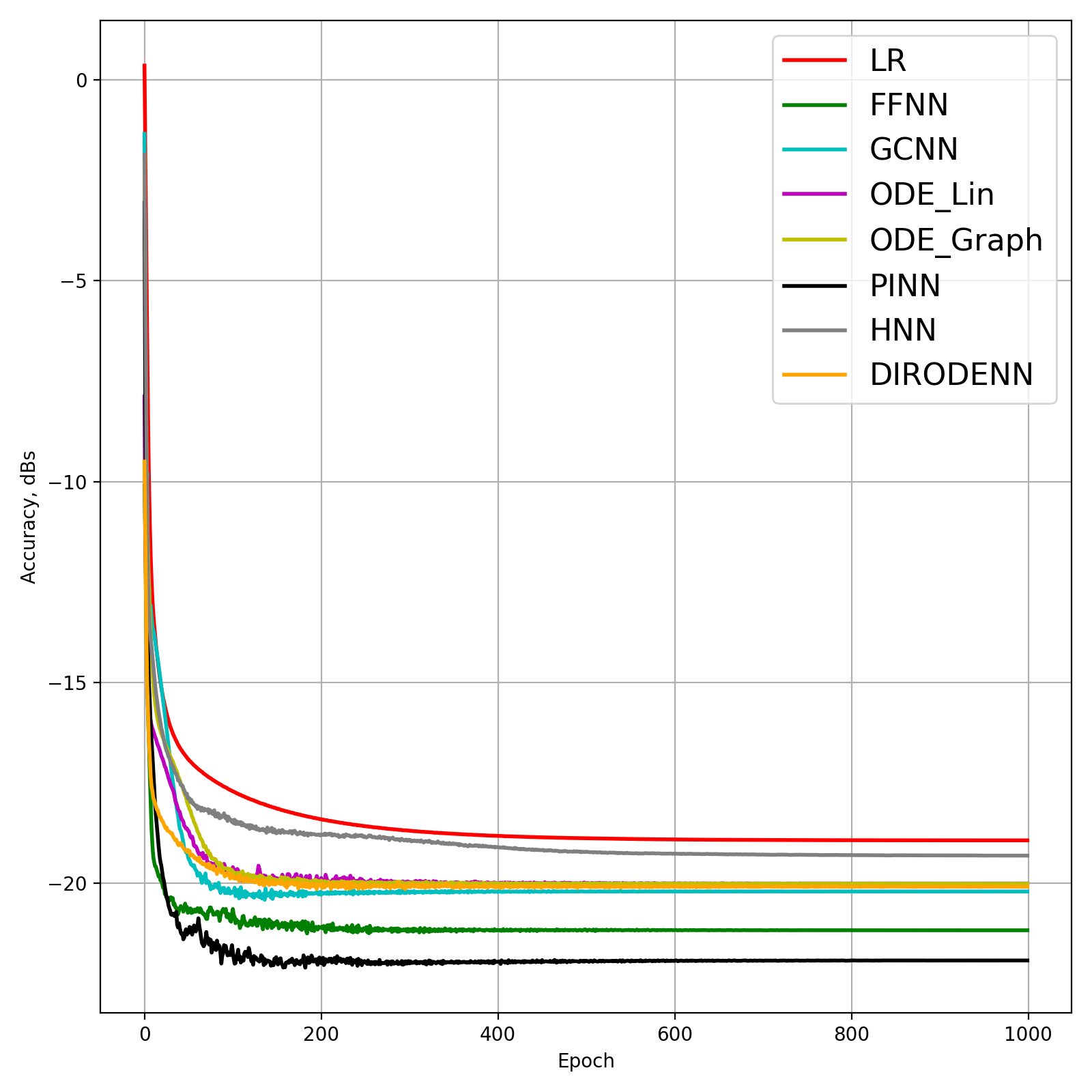}}
	\caption{Comparison of the learning model performance for detection of line failure in the dynamic regime. \label{fig:failure_dynamic}}
\end{figure}

In this Section we describe the setting, report and discuss results of our dynamic state estimation experiment with the dynamic models described in the preceding Section \ref{sec:dynamic}. As in the (static) fault detection setting, described in Section \ref{sec:exp-static}, we are conducting our dynamic experiments with the data generated in the Power system toolbox \cite{PST} on the IEEE 68-bus electrical network under sufficiently small dynamic perturbations \footnote{The authors are grateful to Wenting Li for providing temporal, PST generated data.}.  Changes in the dynamic setting (when compared with the static one of Section \ref{sec:exp-static}) are as follows. Input/output is the dynamic path, $\{x(t)\}_K$,  where at each $t_k$, $x(t_k)$ represents the voltage potential (absolute value and phase) measured at the observed nodes of the system. That is, $\{x(t)\}_K\in\mathbb{R}^{2\times 68\times K}$, in the case of the full observability and, $\{x(t)\}_K\in\mathbb{R}^{2\times |{\cal V}_o|\times K}$, in the case of the partial observability. We are experimenting with [5 \%, 10 \%, 20 \%, 40 \%, 70 \%, 100 \%] node observation levels.  We experiment with the dynamic models expressing different degree of physics, discussed in Section \ref{sec:dynamic},  but also test static models adapted to the (time-incremental) map. \footnote{All static models, but AlexNet are tested in the regime of limited observability. This is because AlexNet is fine tuned to the fixed size of the input, $x(0)\in\mathbb{R}^{2\times 68}$, and adapting it to partial observability would require reconstructing entire architecture of the NN. Moreover, AlexNet was not competitive in the case of the full observability.} In this dynamic state estimation setting we select observation nodes at random, and then repeat multiple experiments (collect statistics) for this particular set of the observed nodes. Our training consist in minimizing the $l_2$ norm given by Eq.~(\ref{eq:l2-Loss-NeuralODE-Path}), adapted respectively to different dynamical models considered \footnote{In the cases of $\text{path-HNN}_{\phi}$ and $\text{path-DIRODENN}_{\phi}$, and according to the discussions above,  some of the parameters will be physical and some physics-ignorant. Optimization in the respective version of Eq.~(\ref{eq:l2-Loss-NeuralODE-Path}) is over all the parameters.}. Actual optimization is implemented by the Adam \cite{kingma2014adam} gradient method over 1000 epochs (Under exception of the case of the HNN model under 100 \% of observability, where the training is over 200 epochs). Details on the specific implementations of our dynamic state estimation experiments are provided in Appendix \ref{append:exp2}. The results are presented in Fig.~\ref{fig:failure_dynamic}. Accuracy of prediction (in dB) is accessed according to log of the ration of the mismatch between predicted and observed, normalized to the observed: 
$\text{Accuracy} = 10\lg(\frac{P_{\text{error}}}{P_{\text{output}}})$, where $P_{\text{error}} = \sum_t ||x^{(\text{pred})}_t - x_t^{(\text{pred})}||^2_2$,  $P_{\text{output}} = ||x_{t}^{(\text{pred})}||^2_2$.

\subsection{Dynamic State Estimation which Extrapolates: Discussion of Experiments}

It is important to start discussion of our dynamic state estimation experimental results with a disclaimer (which also motivated material discussed in the next Section \ref{sec:placement}) -- the results are very sensitive to where the measurements are placed. Otherwise, we observe that under full observability the models which are most physics informed, e.g. DIRODENN, and especially HNN, perform better than physics agnostic models, of which the only linear one (LR) is the worst in performance. Systematic decrease of observability, from almost complete to modest, does not affect the qualitative assessment much. We interprete this prelimiinary conclusion (preliminary in view of the disclaimer above) as the confirmation of the general expectation that adding information about the structure of the power systems and especially about its dynamics, helps to extrapolate, i.e. in our context represent part of the system where no samples were observed. On the other hand, when the observability becomes poor, it seems that the models which are from the middle of the pack (in terms of their use of the power system physics), such as PINN and Graph-ODENN which are aware of rather rough physical structure of the power system (and not about details) are the winners of the performance competition. This suggests that planting too much of physics into the dynamic state estimation algorithm in the regime of low observability may also lead to a diminishing return. 

\section{ML Algorithms for Optimal Placement of PMUs}  \label{sec:placement}

As the first set of experiments (detection of failure in the static regime, reported and discussed in Section \ref{sec:failure_static}) show, accuracy of the ML model varies very significantly not only on the percentage of nodes where observations are available,  but also on where exactly within the system the observations are made. This dependency  motives the third set of experiments discussed below. Specifically, we focus in this Section on building ML schemes which are capable to discover efficiently, i.e. fast, locations for close to optimal placement of the Phasor Measurement Units (PMUs) for the given level of observability.  

Notice that this problem of searching for the optimal PMU placement was already addressed in \cite{2019WentingLi}.  However,  the algorithm suggested there was "passive", which means that the algorithm worked in parallel with training of the main model (in the setting of our first experiment).  Stating it differently, in the passive search the placement configurations do not utilize the information received so far. In theory, this passive sampling conducted without a feedback loop, should eventually find the optimal PMU placement,  however the passive search is normally taking long time.  
 
In the following we develop an active strategy which reinforce the search by taking advantage of the measurement made so far and  thus allowing a much faster discovery of the optimal PMU placement than in the passive case considered so far.

\begin{figure}[h!]
	\centering
	\includegraphics[width=\linewidth, page=1]{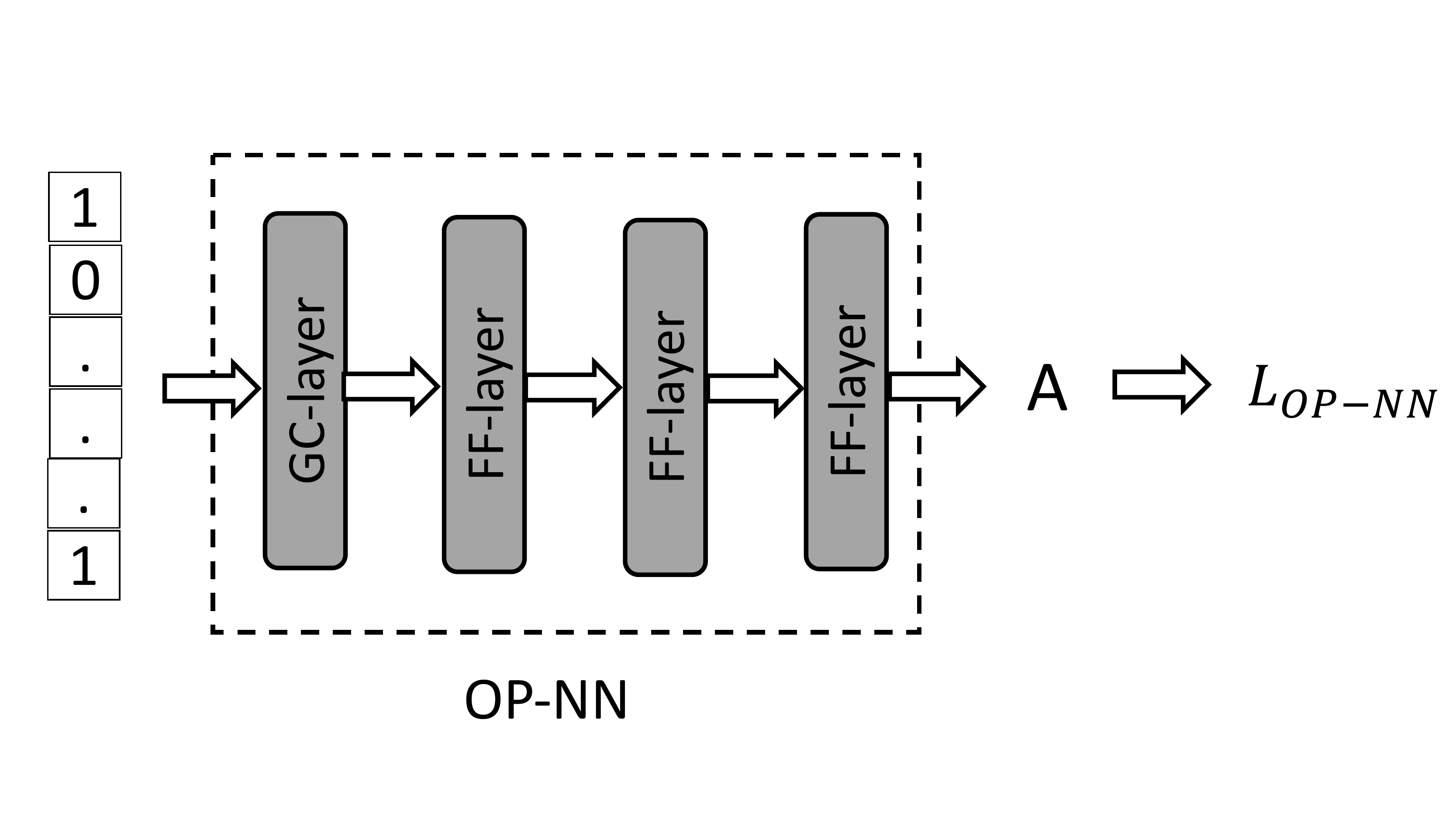}
	\includegraphics[width=\linewidth, page=2]{figs/AndreyOptimalPlacement.pdf}
	\caption{Architecture of the Optimal Placement NN. Top and bottom sub-figures illustrate the two-step training of the OP-NN. Gray shading highlights components of the architecture trained during the respective stage (parameters are tuned). See text for details.  }
	\label{fig:OP-NN}
\end{figure}

The main idea of the approach, illustrated in Fig.~(\ref{fig:OP-NN}), is in solving the OP problem in two steps: First,  find a function which maps each set of observed nodes to a score expressing the Accuracy, $A$, of the reconstruction: $f:{\cal V}_o\to A$, where $A\in [0,1]$, and $0$ and $1$ correspond to complete failure and success of the reconstruction, respectively.  Second, find the argument of the minimum of the function,  suggesting the desired Optimal Placement (OP). We construct function, $f$, by means of learning from multiple  Input-Output-Placement (IO-P) samples,  where each IO-P sample aggregates multiple samples correspondent to experiments discussed in Section \ref{sec:failure_static} conducted for the same placeemnt (i.e. the same set of observed nodes, ${\cal V}_o$) and for the same basic NN model,  for example LR model. Accuracy, $A$, of a particular OP-IO sample, corresponding to the asymptotic y-axis value of a curve in the Figure \ref{fig:failure_static} achieved at the the end of the training run,  becomes the Output of the OP-NN,  shown in the left sub-figure of Fig.~(\ref{fig:OP-NN}). Additional details on the structure of the OP-NN are given in Appendix \ref{append:exp3}. Parameters of the OP-NN,  built from four layers (a Graphical Convolutional layer, followed by three Feed-Forward layers), are trained during the first stage by minimizing $L_{\text{OP-NN}}$, chosen to be the $l_2$ norm between $A$-predictions and $A$-observations. The second stage consists in fixing parameters of the OP-NN and then finding arg-maximum of the resulting optimal function, $f$. It is achieved by finding optimal vector $\alpha=(\alpha_{a}\in \mathbb{R}|a\in{\cal V})$, built from $n=|{\cal V}|$ real valued components, mapped via $g^{(s)}(\alpha)*\text{OP-NN}$ to the accuracy, $A$.  Here the $g^{(s)}(\alpha)$ is the function mapping a real valued  $\alpha$ associated to a vector the same length $n$ having non-zero components at the nodes of suggested PMU placement, formally
\begin{gather*}
    g_a^{(s)}(\alpha)\!=\!\frac{\exp(\alpha_a)}{\sum\limits_{b\in{\cal V}}\exp(\alpha_b)}\times\Big\{\!\! \begin{array}{cc} 1,& \alpha_a \in \text{ top-}s\text{ comp. of }\alpha;\\ 0,& \text{otherwise.}\end{array}
\end{gather*}
This additional "softening" function allows to take advantage of the automatic differentiation for finding minimum of $f$ efficiently. 

We also use the transfer learning technique \cite{zhuang2019comprehensive} to speed up and improve the quality of the OP scheme. Specifically, we first implement the scheme on (by far) the fastest, but also least accurate, Linear Regression (LR) method and then use the pre-trained  LR-OP-NN as a warm-start for training other (more accurate, but slower) methods of the OP reconstruction.

\subsection{Experiments and Discussions}

We pre-train the OP function, $f$, illustrated in Fig.~\ref{fig:OP-NN} on the LR training data in the experimental setting of the Section \ref{sec:failure_static}, on 1600 samples, each characterized by the (Placement nodes, LR Accuracy) pair. The results is used as a warm start for training all other schemes (FFNN, GCNN, AlexNet, ODE Lin, ODE Graph) independently, and each on 350 samples of (Placement nodes, Method Accuracy).
(Specifically, in training the advanced methods we fix parameters of the first three layers according to the pre-trained LR-OP-NN and retrain the last layer.)  We use the Adam \cite{kingma2014adam} gradient method for 1200 epochs with an initial step of $ 0.08 $ and decrease it by factor 10 every 300 epochs at the pre-training (LR) stage. We use the same method for 300 epochs, with an initial step of $ 0.01 $ and decrease it by factor 10 times every 100 epochs at the post-training (advanced methods) stage. 

\begin{figure}
	\subfigure[\ LR 70\% observability]{
		\includegraphics[scale=0.2]{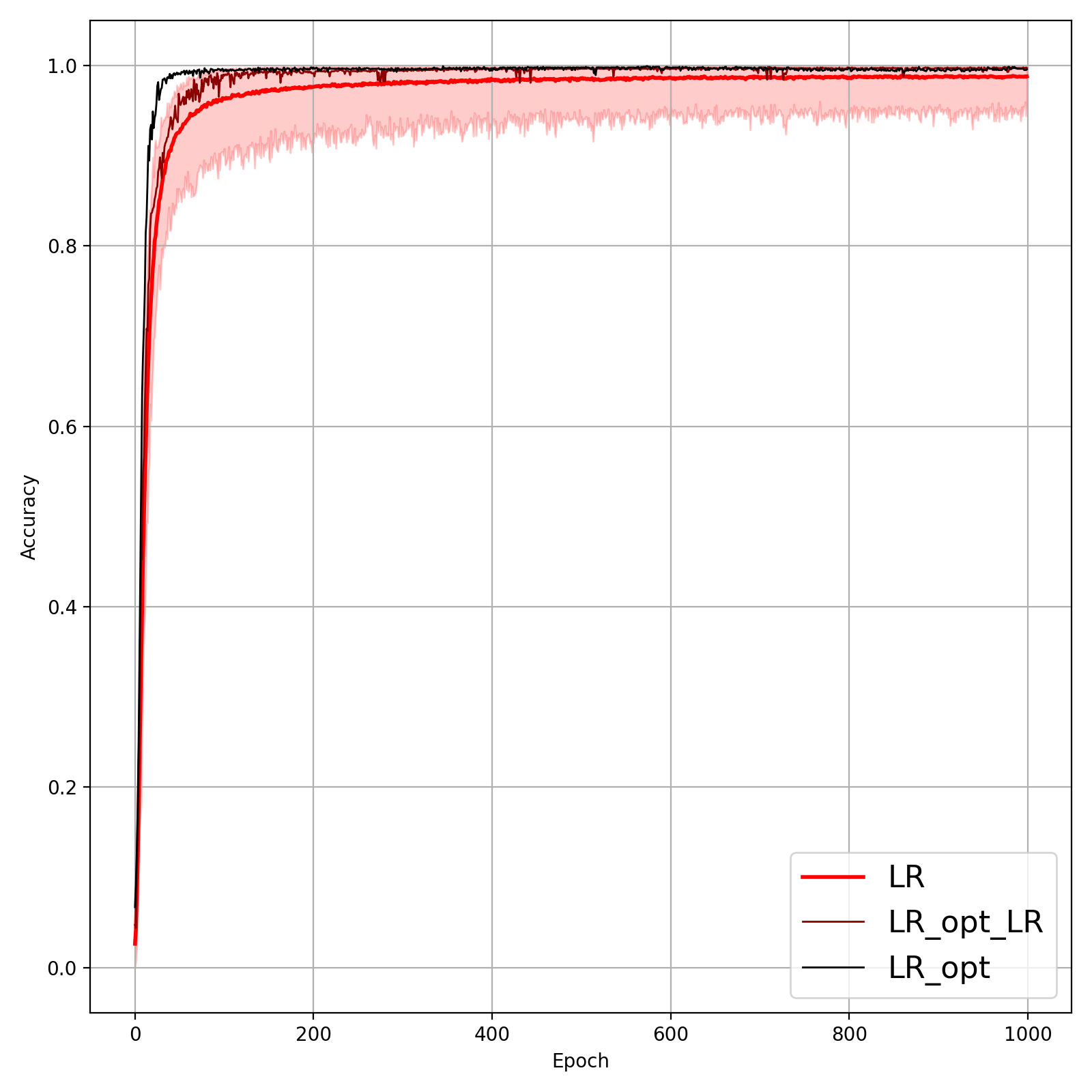}}
	\subfigure[\ LR 40\% observability]{
		\includegraphics[scale=0.2]{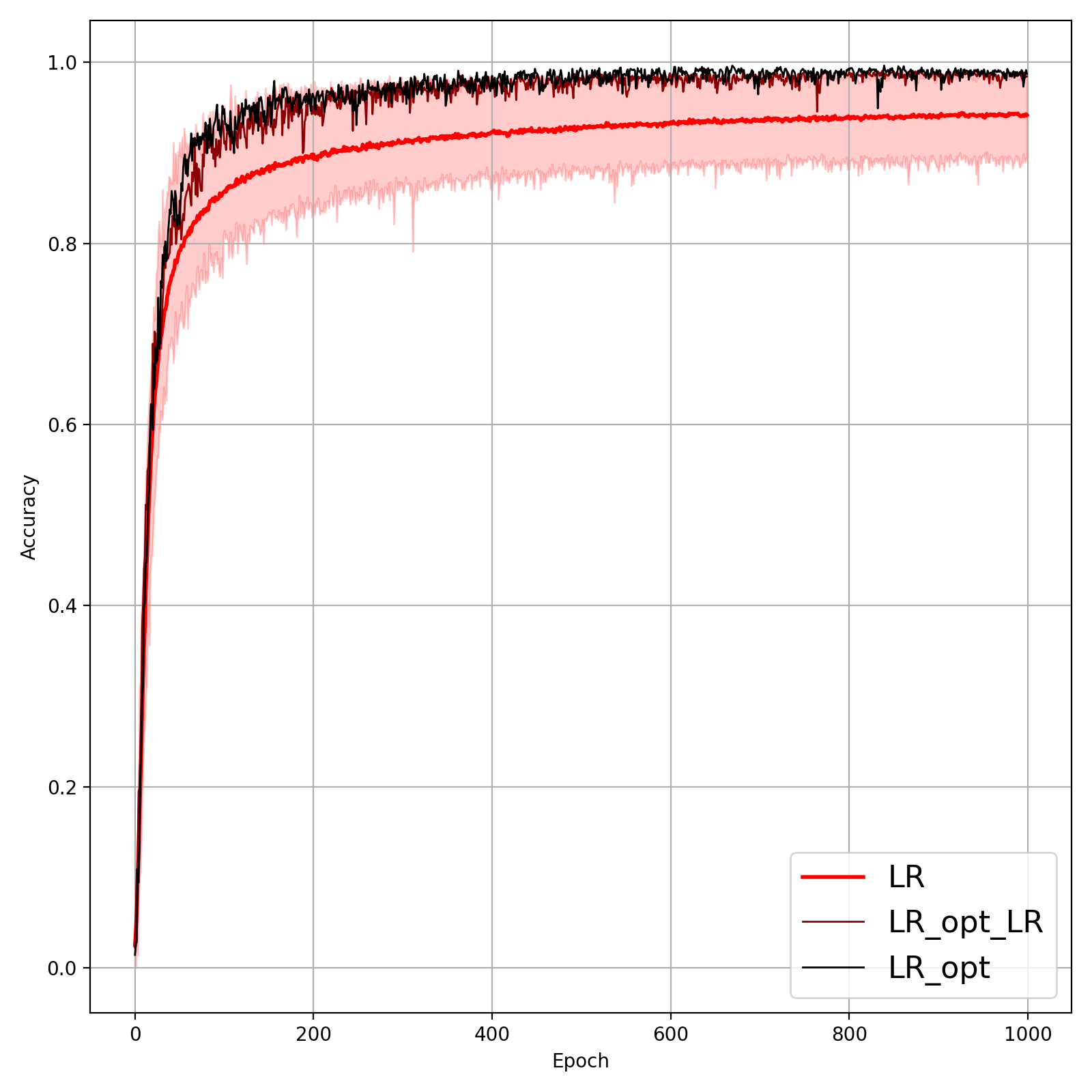}}
	\subfigure[\ LR 20\% observability]{
		\includegraphics[scale=0.2]{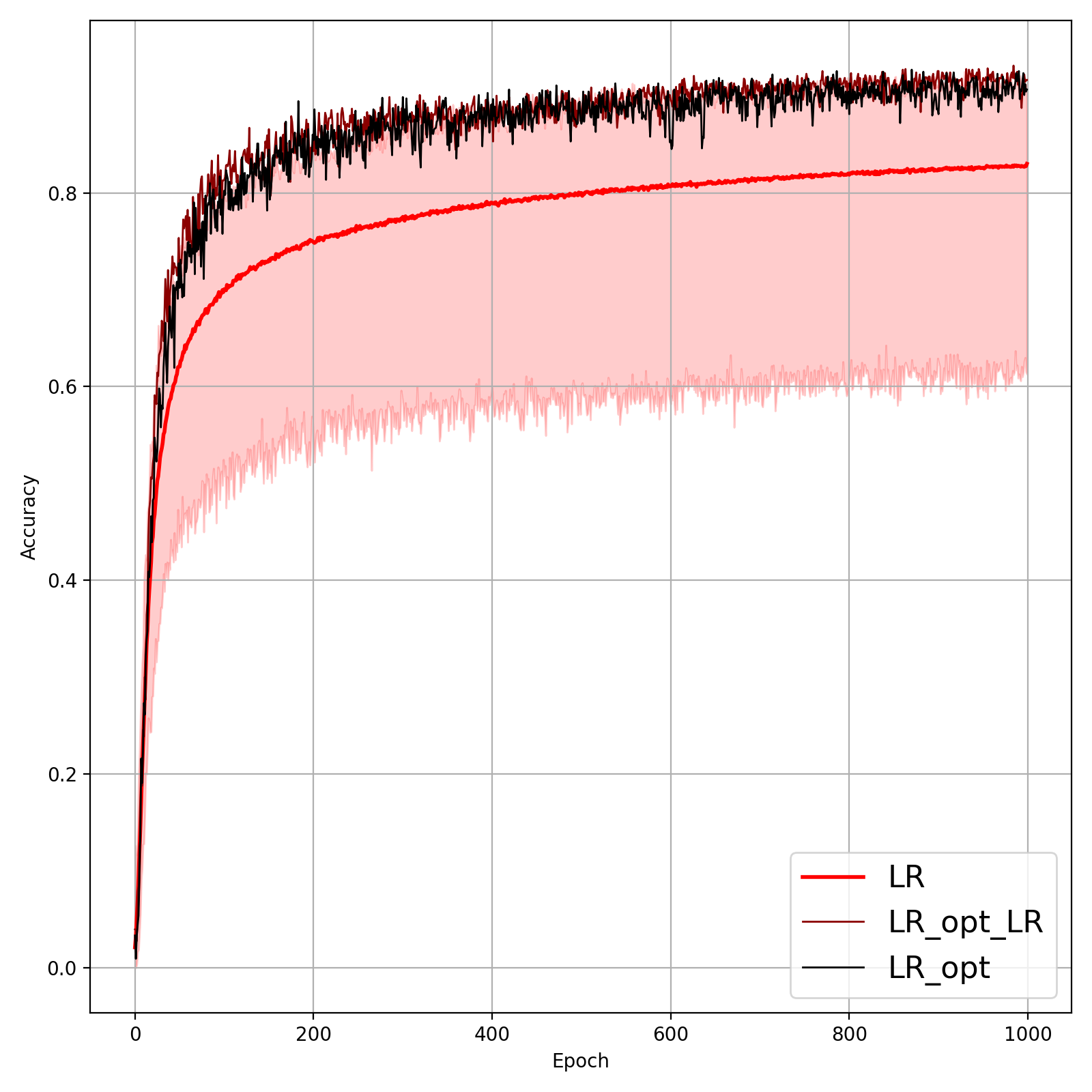}}
	\subfigure[\ LR 10\% observability]{
		\includegraphics[scale=0.2]{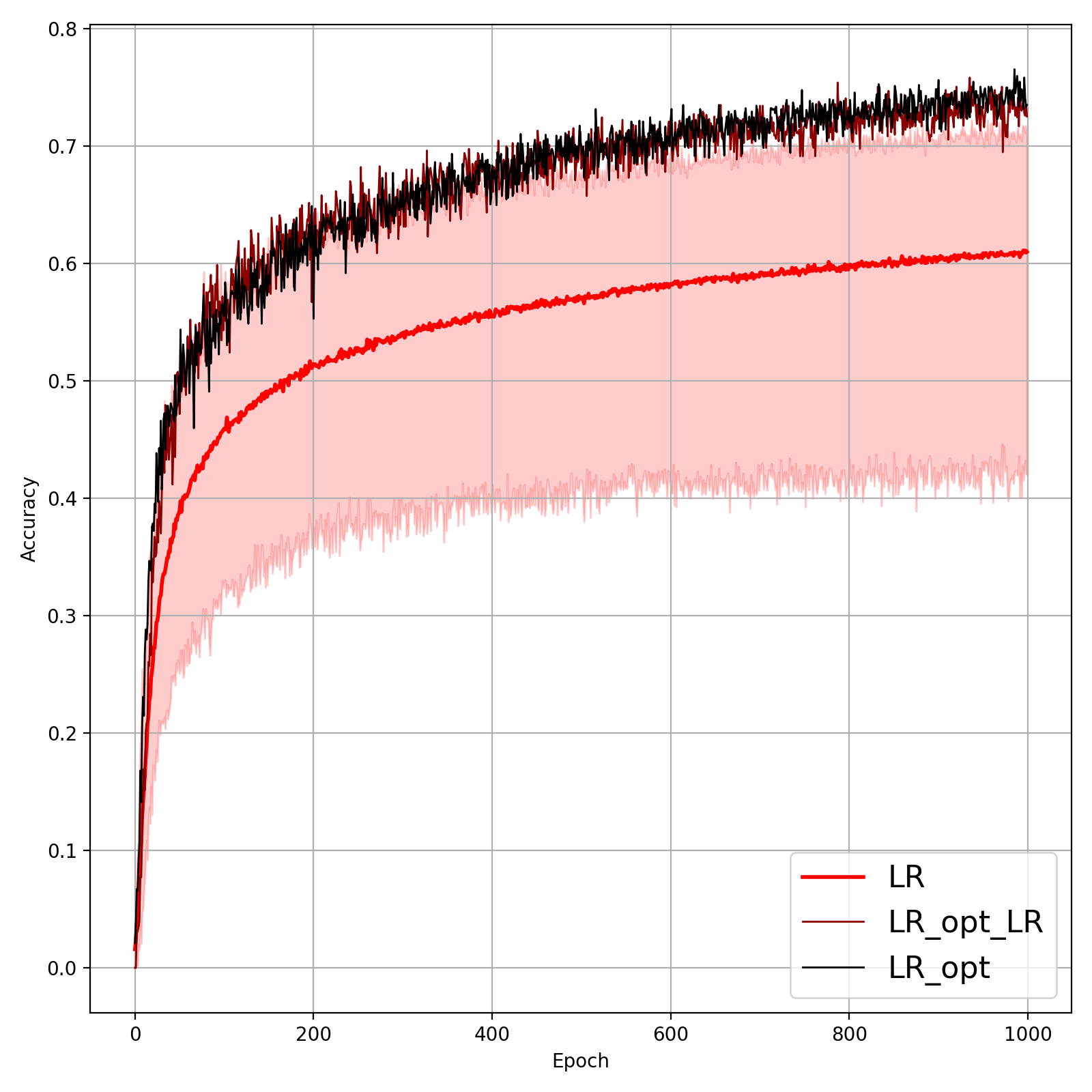}}
	\subfigure[\ LR 5\% observability]{
		\includegraphics[scale=0.2]{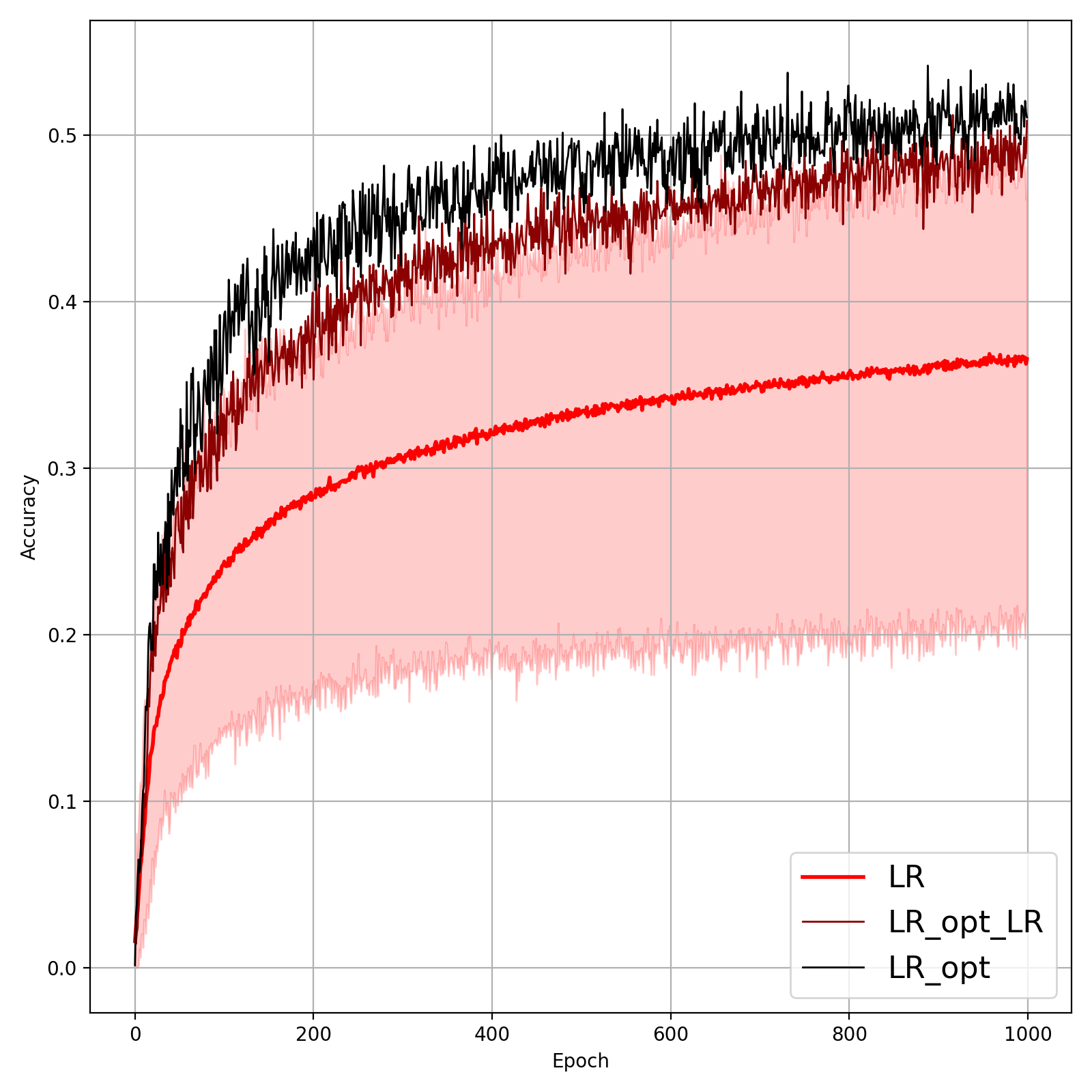}}
	\caption{Comparison of the placement learning with LR model for different observability. \label{fig:placement-LR}}
\end{figure}

\begin{figure}
	\subfigure[\ FFNN 70\% observability]{
		\includegraphics[scale=0.2]{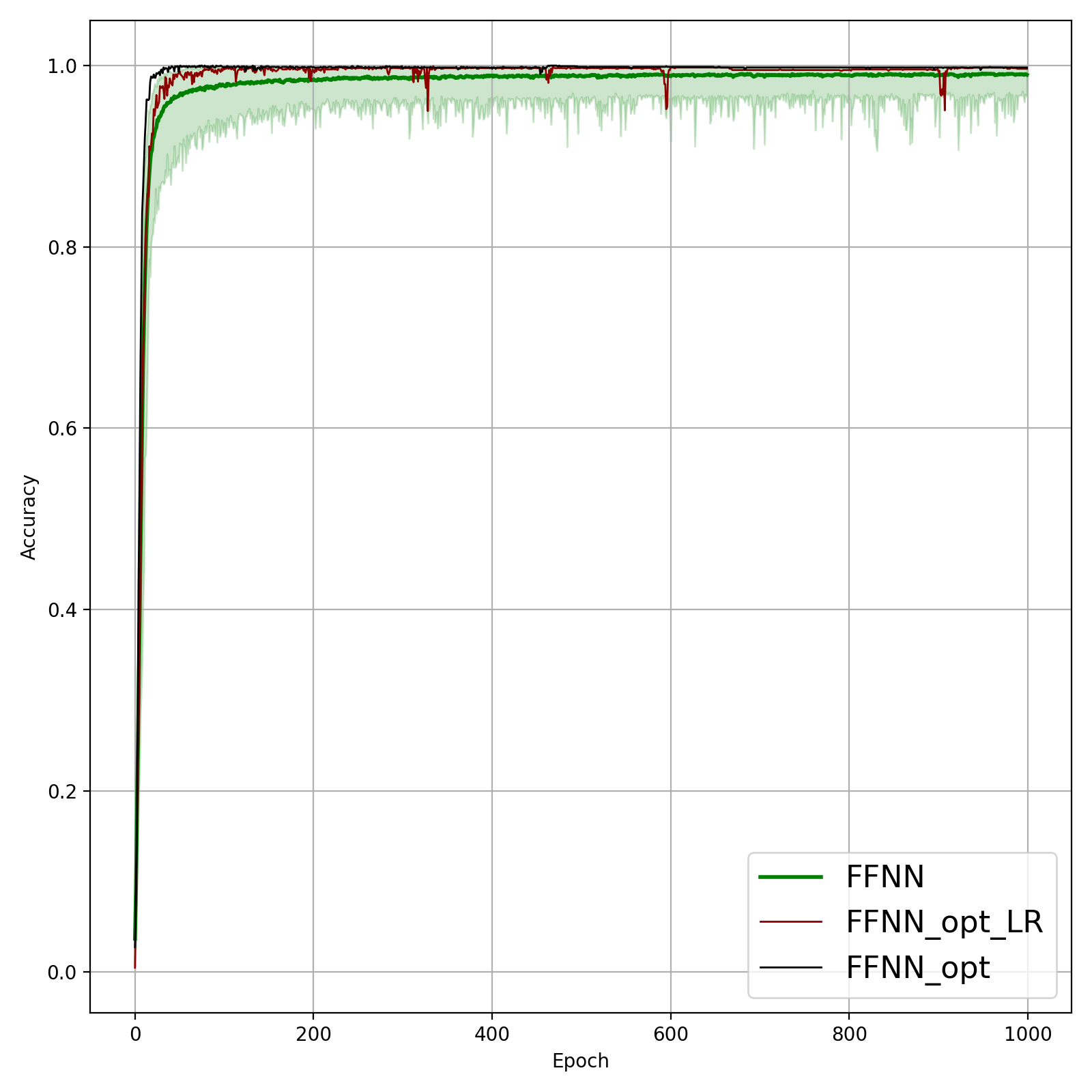}}
	\subfigure[\ FFNN 40\% observability]{
		\includegraphics[scale=0.2]{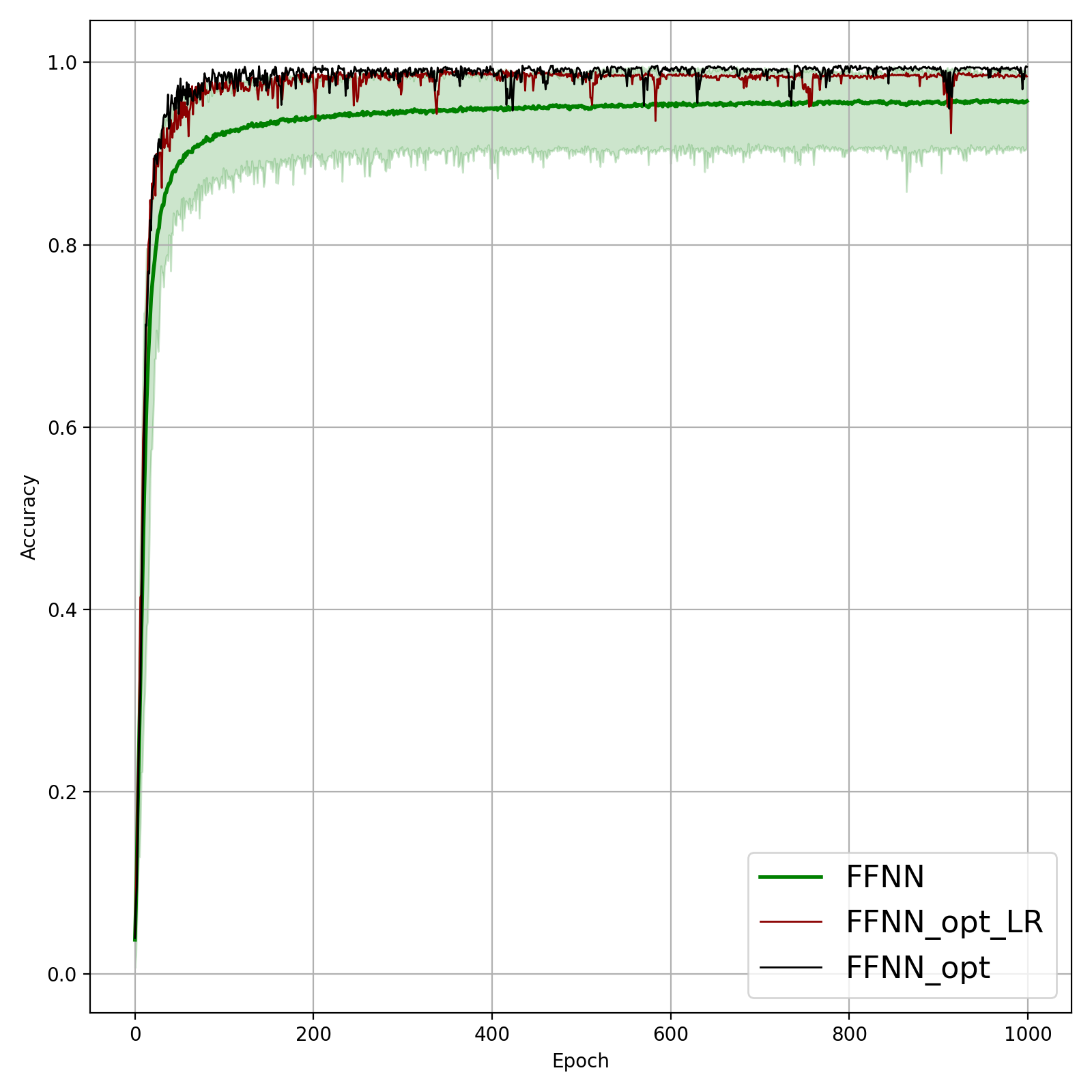}}
	\subfigure[\ FFNN 20\% observability]{
		\includegraphics[scale=0.2]{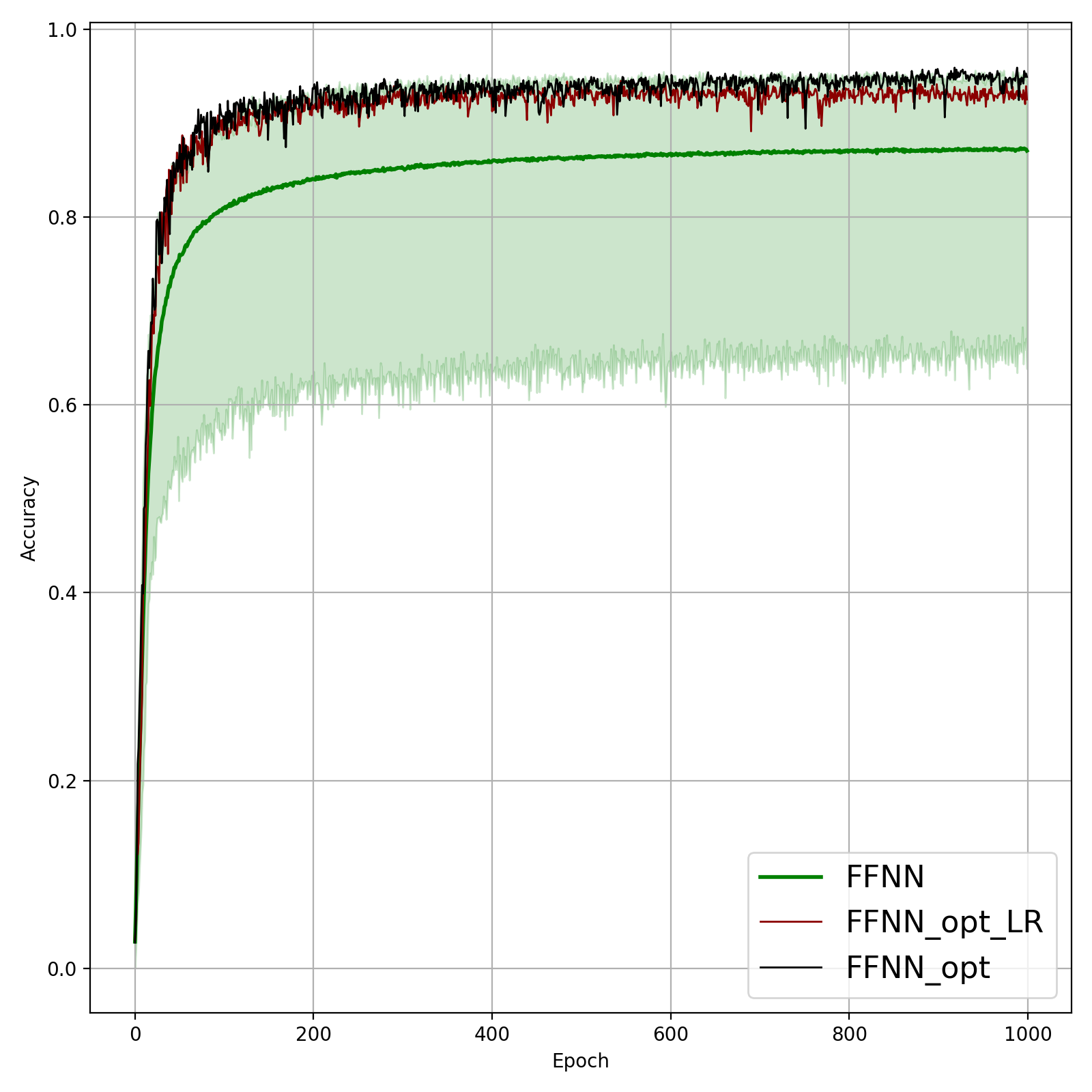}}
	\subfigure[\ FFNN 10\% observability]{
		\includegraphics[scale=0.2]{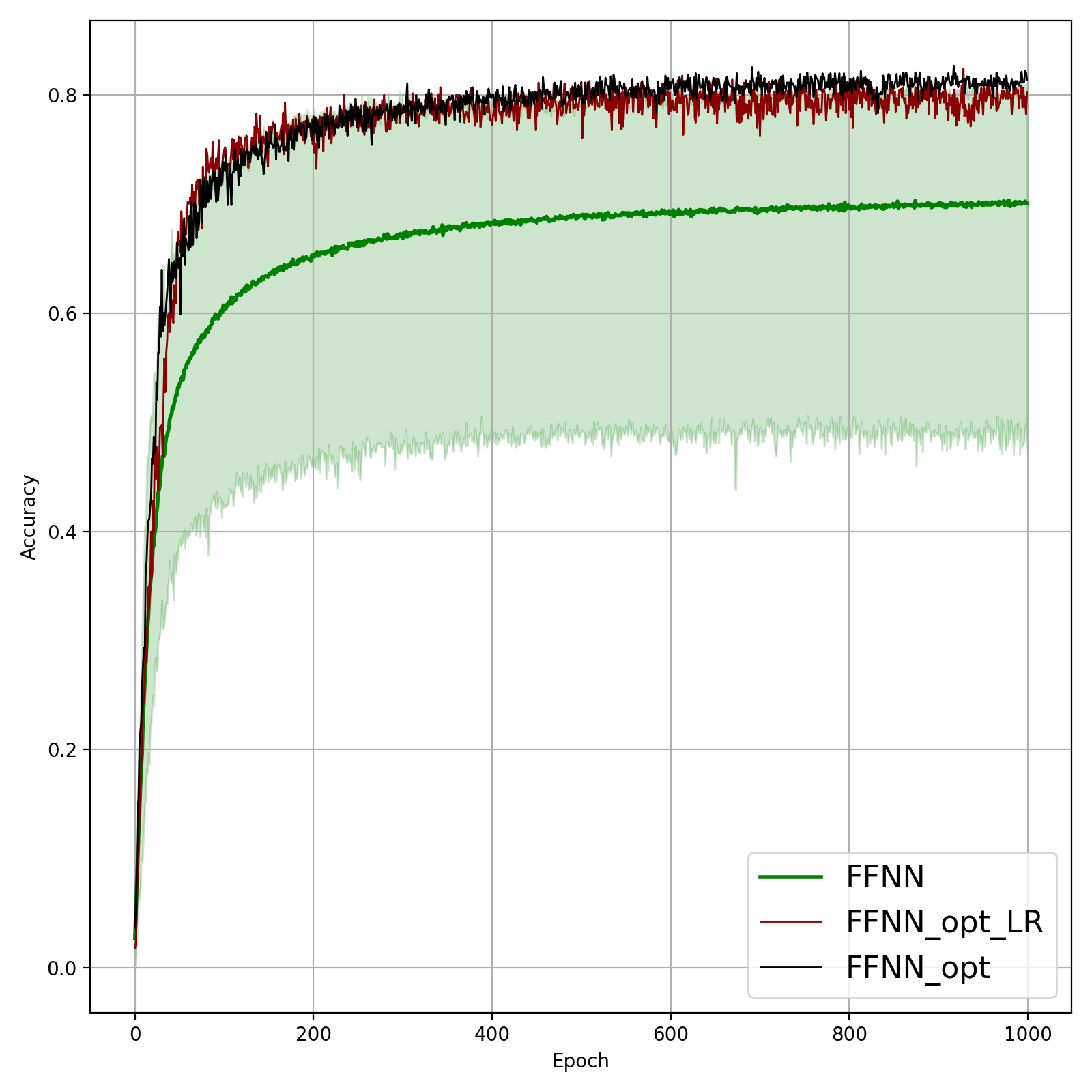}}
	\subfigure[\ FFNN 5\% observability]{
		\includegraphics[scale=0.2]{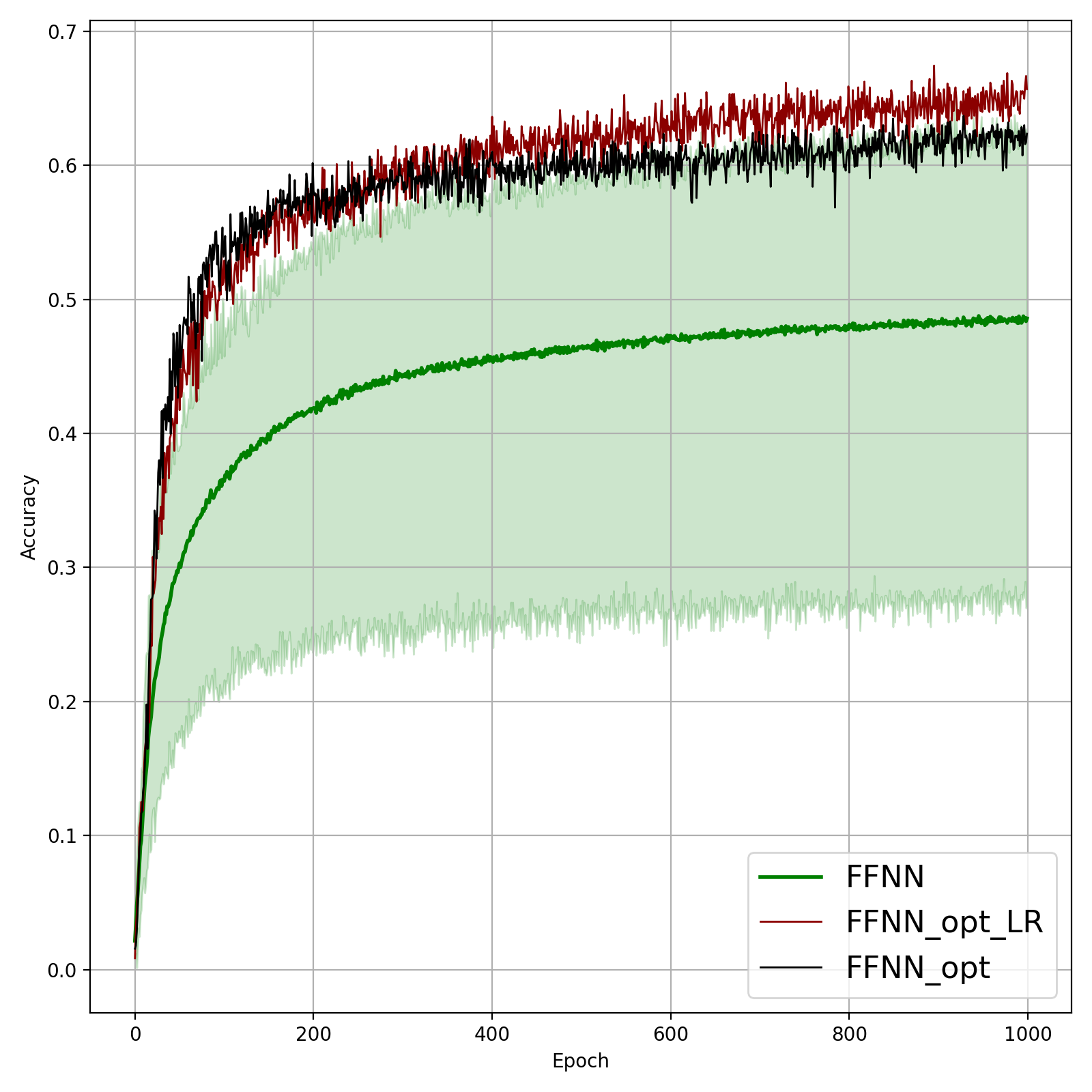}}
	\caption{Comparison of the placement learning with FFNN model for different observability. \label{fig:placement-FFNN}}
\end{figure}

\begin{figure}
	\subfigure[\ GCNN 70\% observability]{
		\includegraphics[scale=0.2]{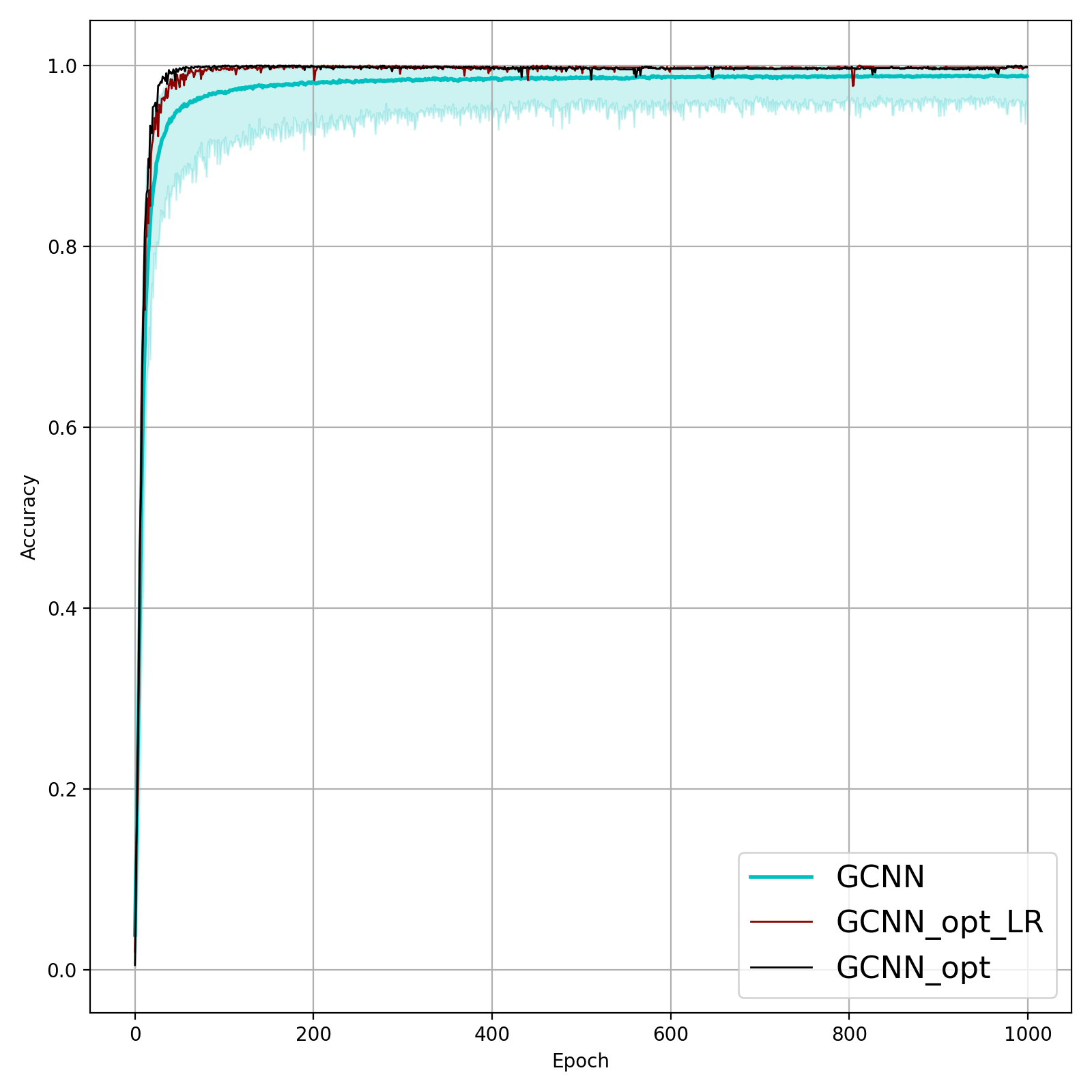}}
	\subfigure[\ GCNN 40\% observability]{
		\includegraphics[scale=0.2]{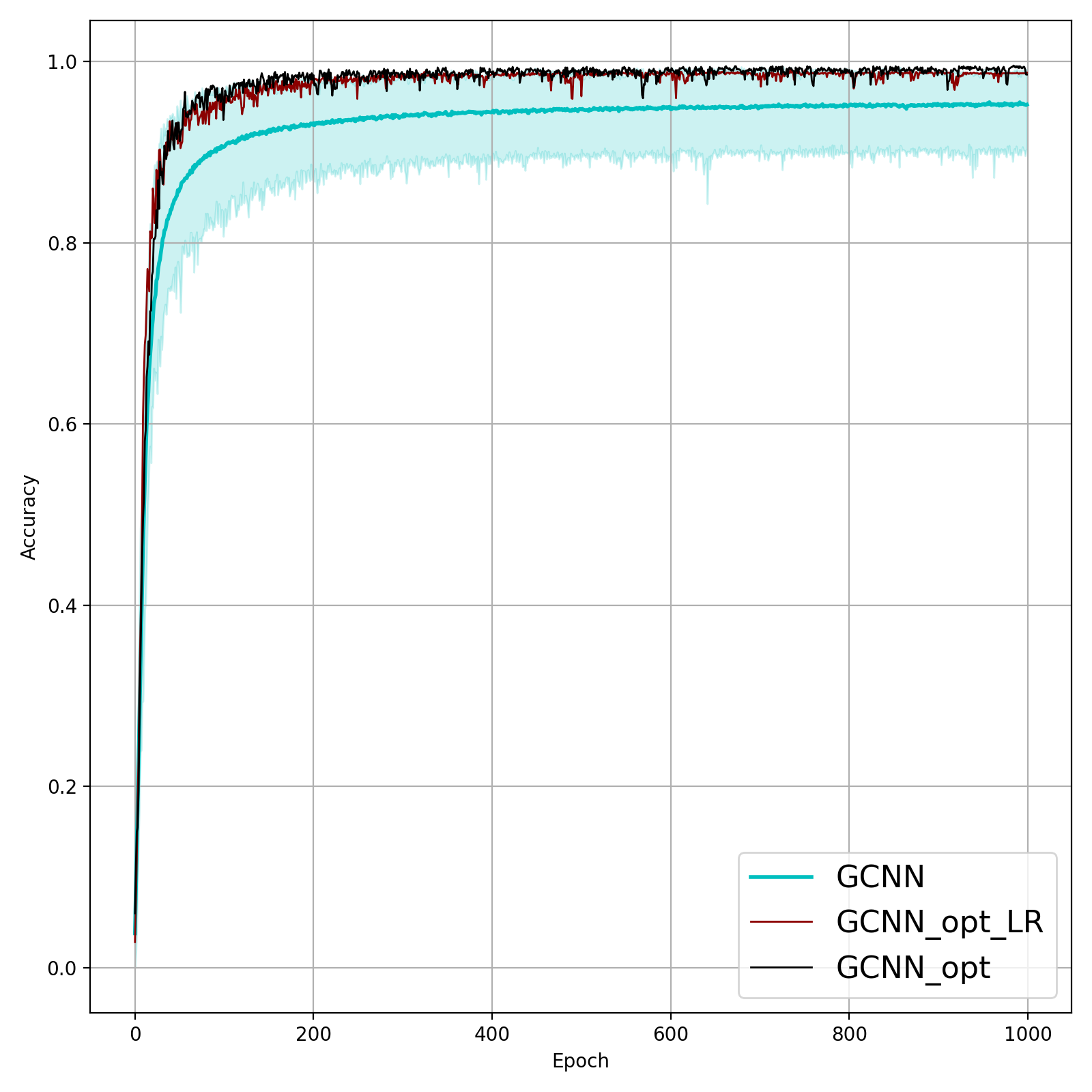}}
	\subfigure[\ GCNN 20\% observability]{
		\includegraphics[scale=0.2]{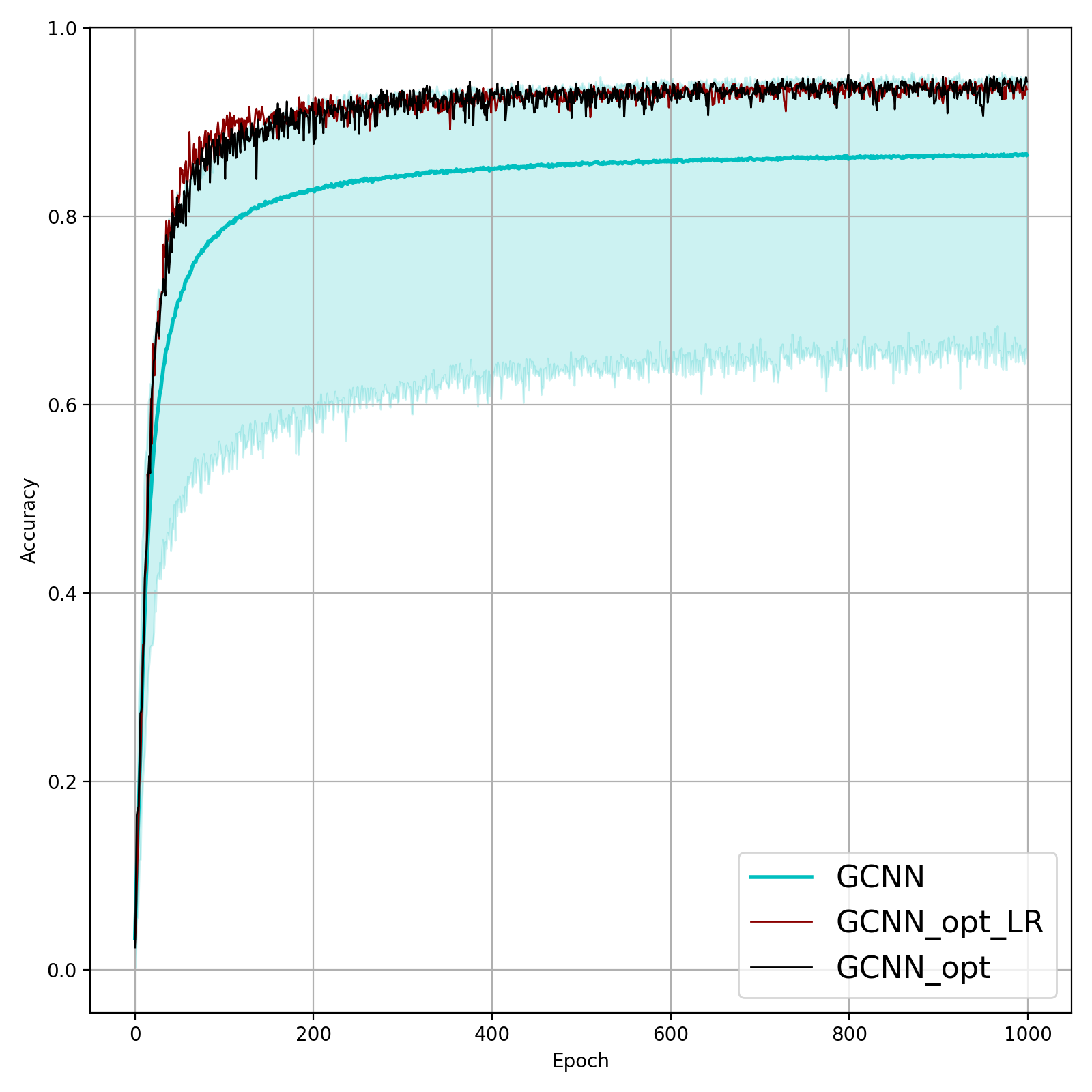}}
	\subfigure[\ GCNN 10\% observability]{
		\includegraphics[scale=0.2]{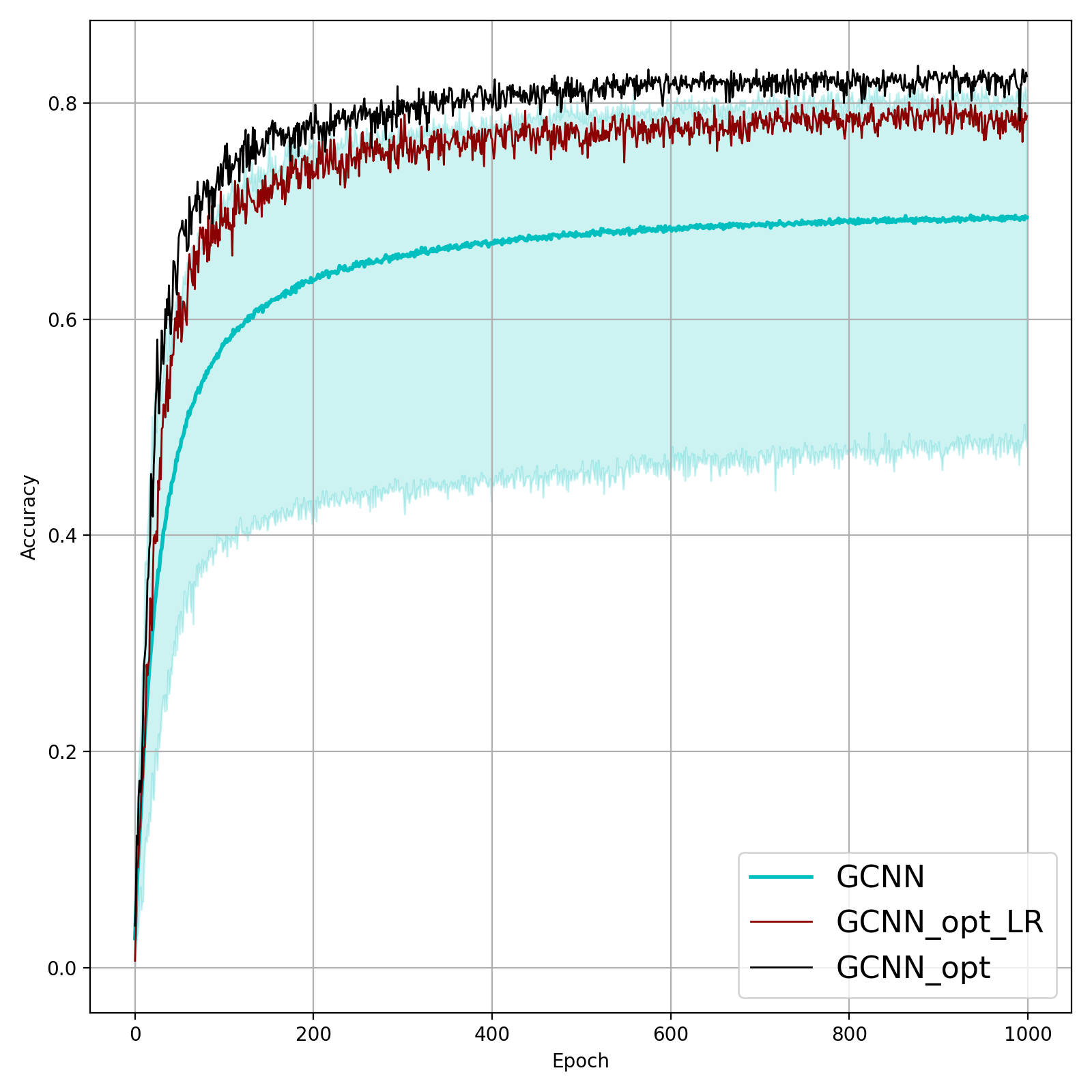}}
	\subfigure[\ GCNN 5\% observability]{
		\includegraphics[scale=0.2]{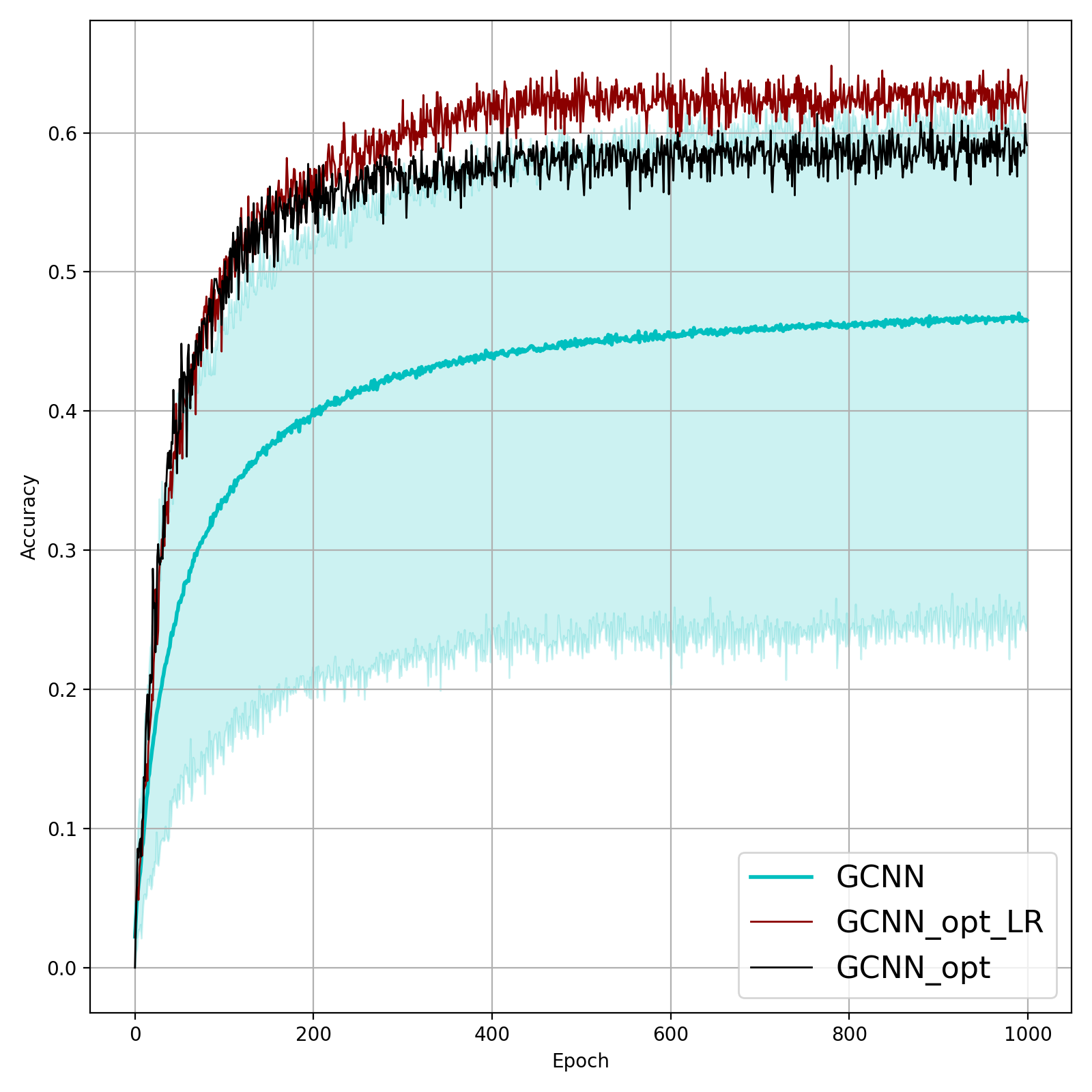}}
	\caption{Comparison of the placement learning with GCNN model for different observability. \label{fig:placement-GCNN}}
\end{figure}

\begin{figure}
	\subfigure[\ AlexNet 70\% observability]{
		\includegraphics[scale=0.2]{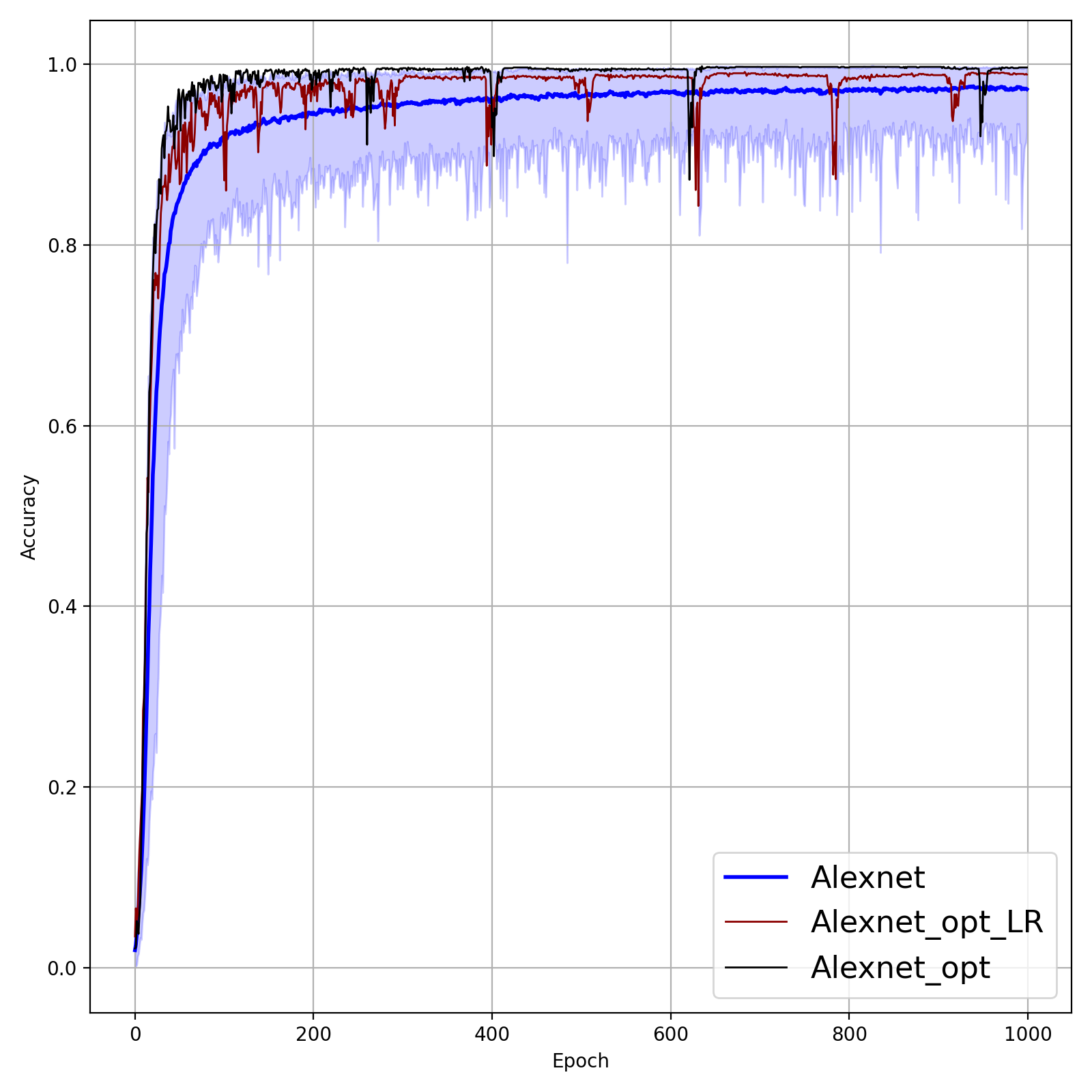}}
	\subfigure[\ AlexNet 40\% observability]{
		\includegraphics[scale=0.2]{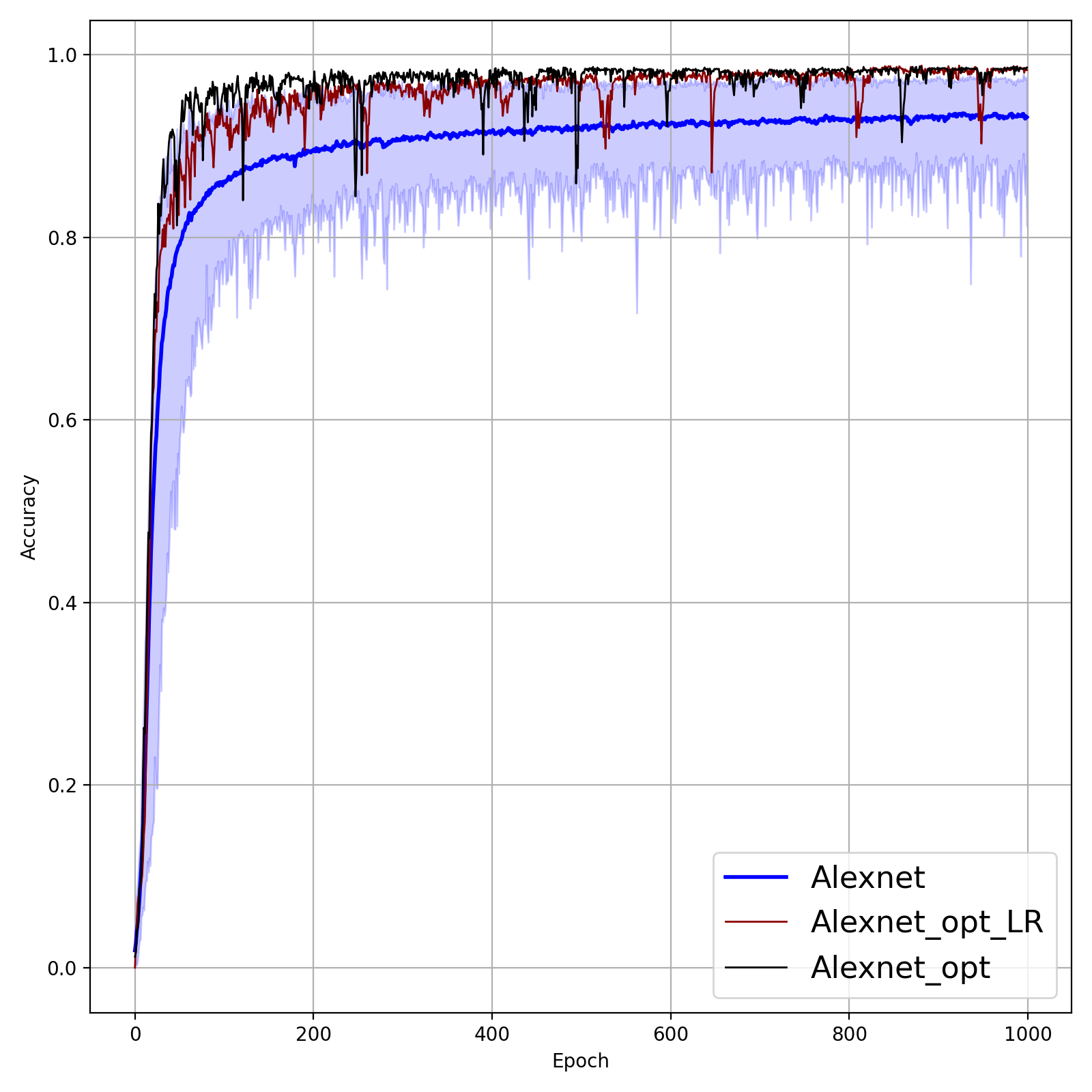}}
	\subfigure[\ AlexNet 20\% observability]{
		\includegraphics[scale=0.2]{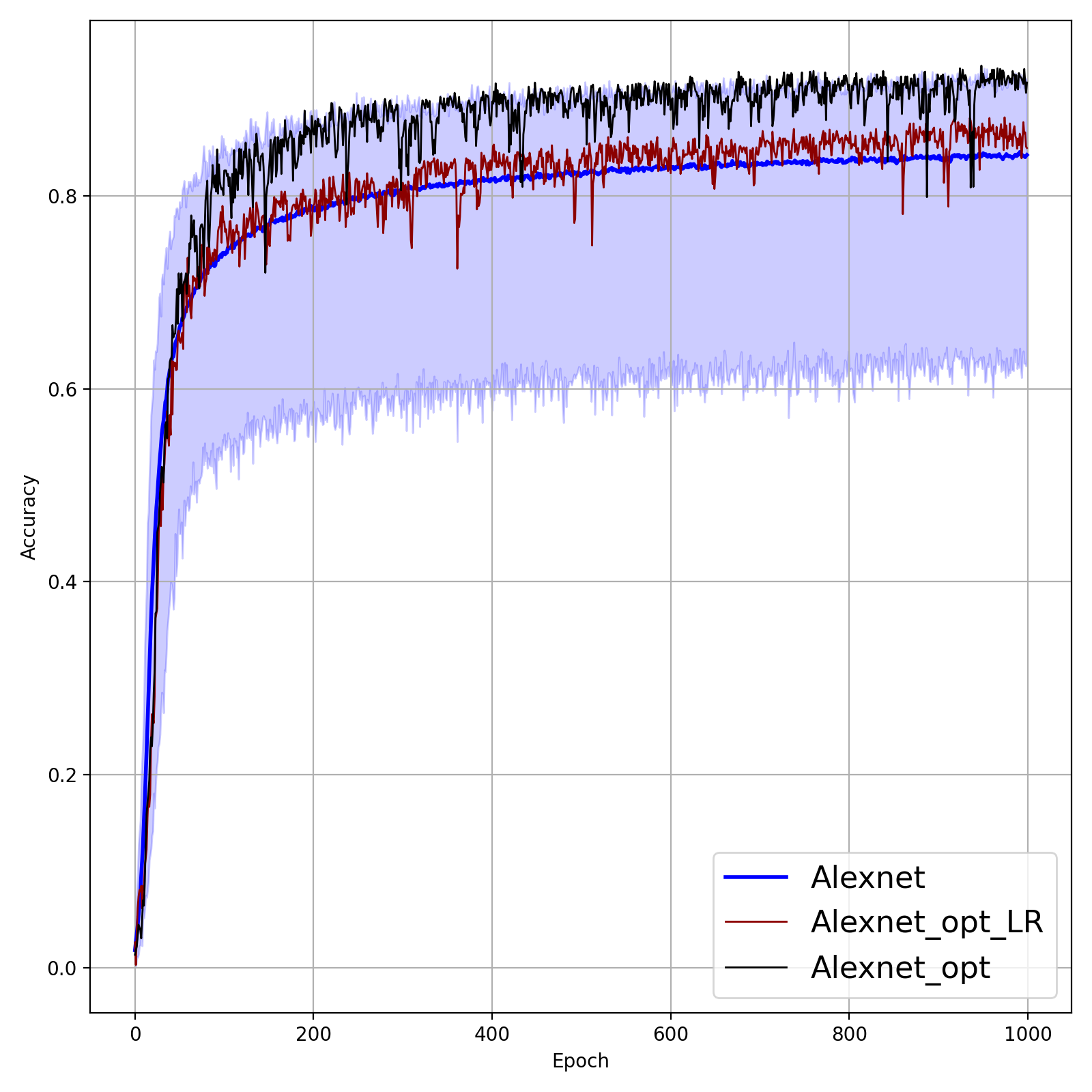}}
	\subfigure[\ AlexNet 10\% observability]{
		\includegraphics[scale=0.2]{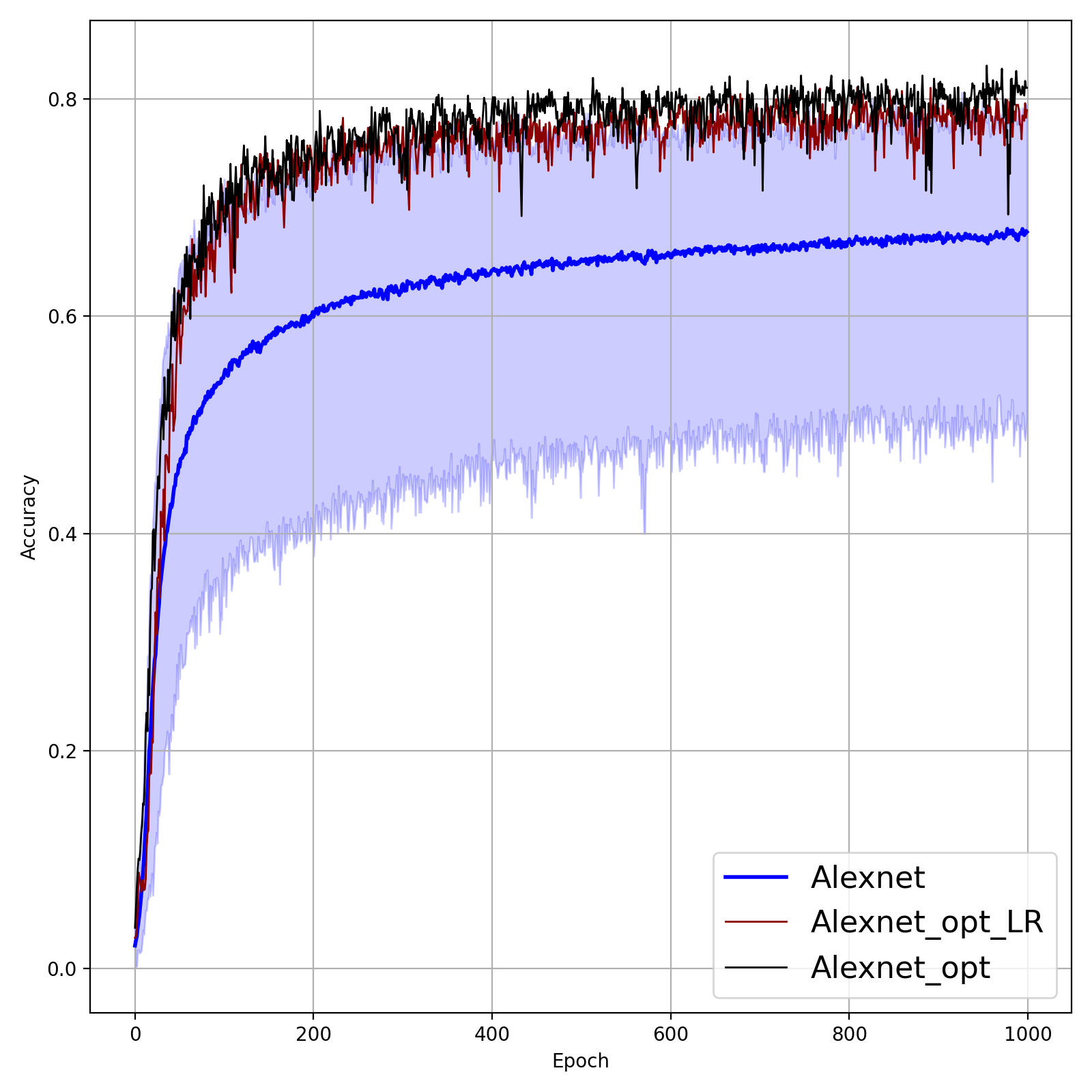}}
	\subfigure[\ AlexNet 5\% observability]{
		\includegraphics[scale=0.2]{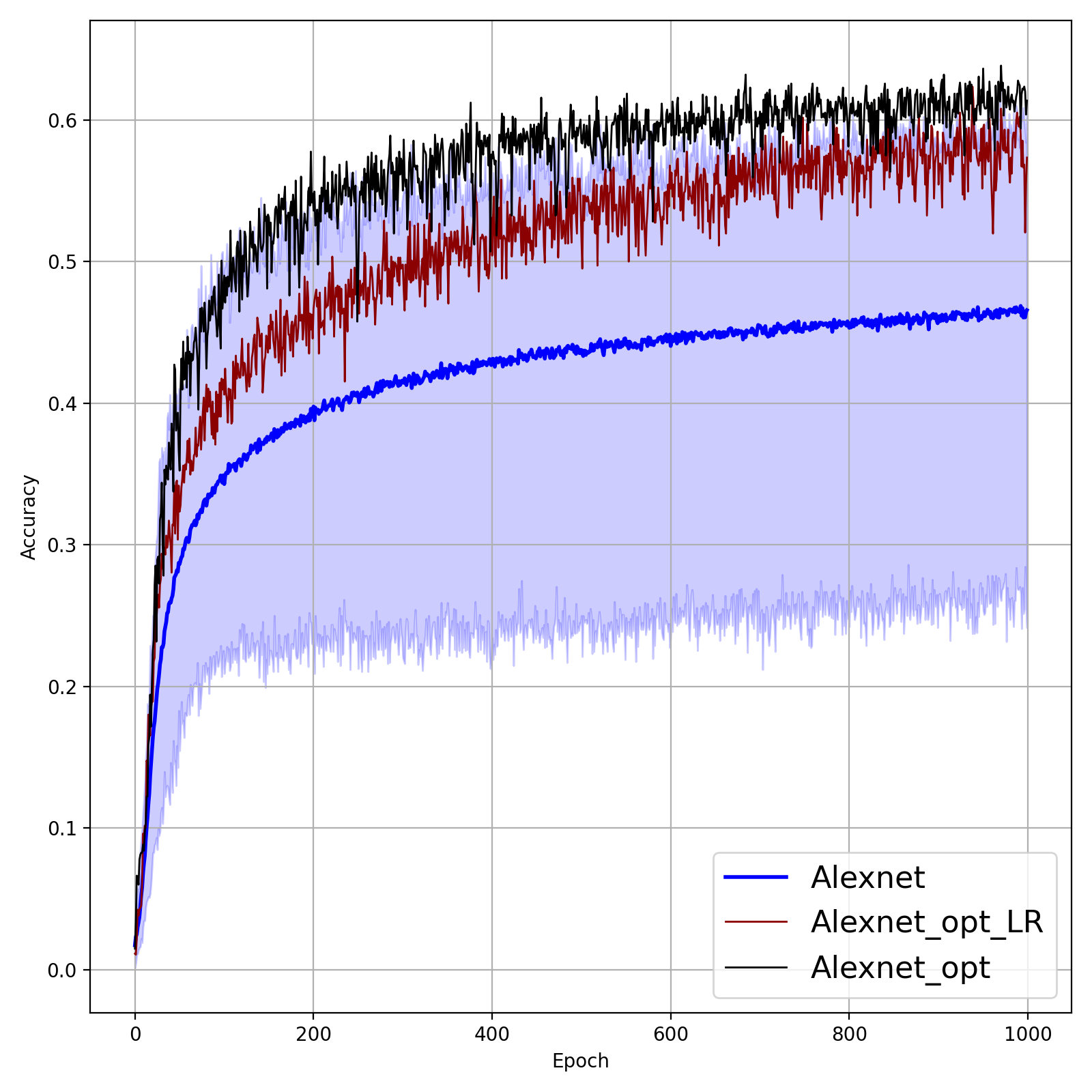}}
	\caption{Comparison of the placement learning with AlexNet model for different observability. \label{fig:placement-AlexNet}}
\end{figure}

\begin{figure}
	\subfigure[\ Linear ODE 70\% observability]{
		\includegraphics[scale=0.2]{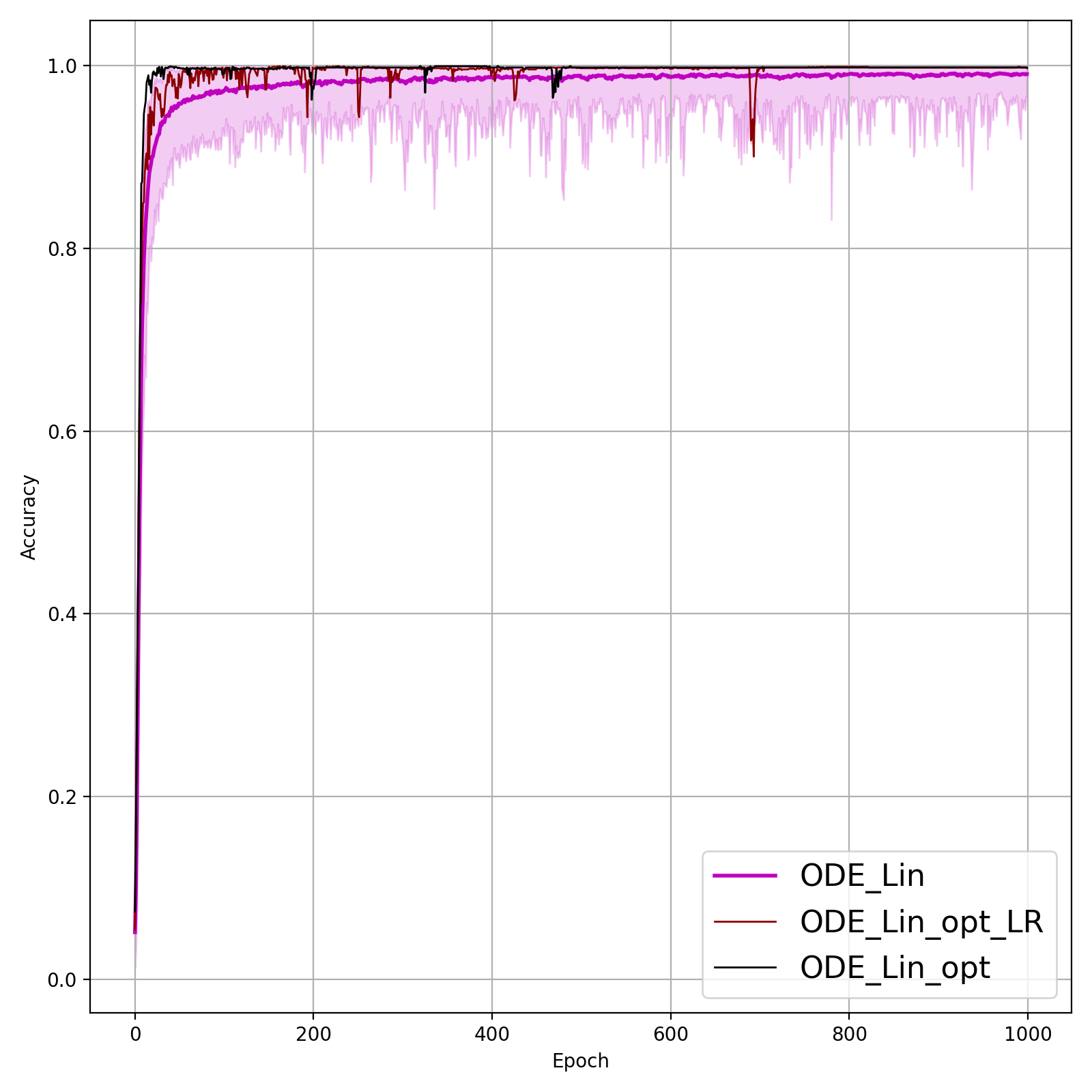}}
	\subfigure[\ Linear ODE 40\% observability]{
		\includegraphics[scale=0.2]{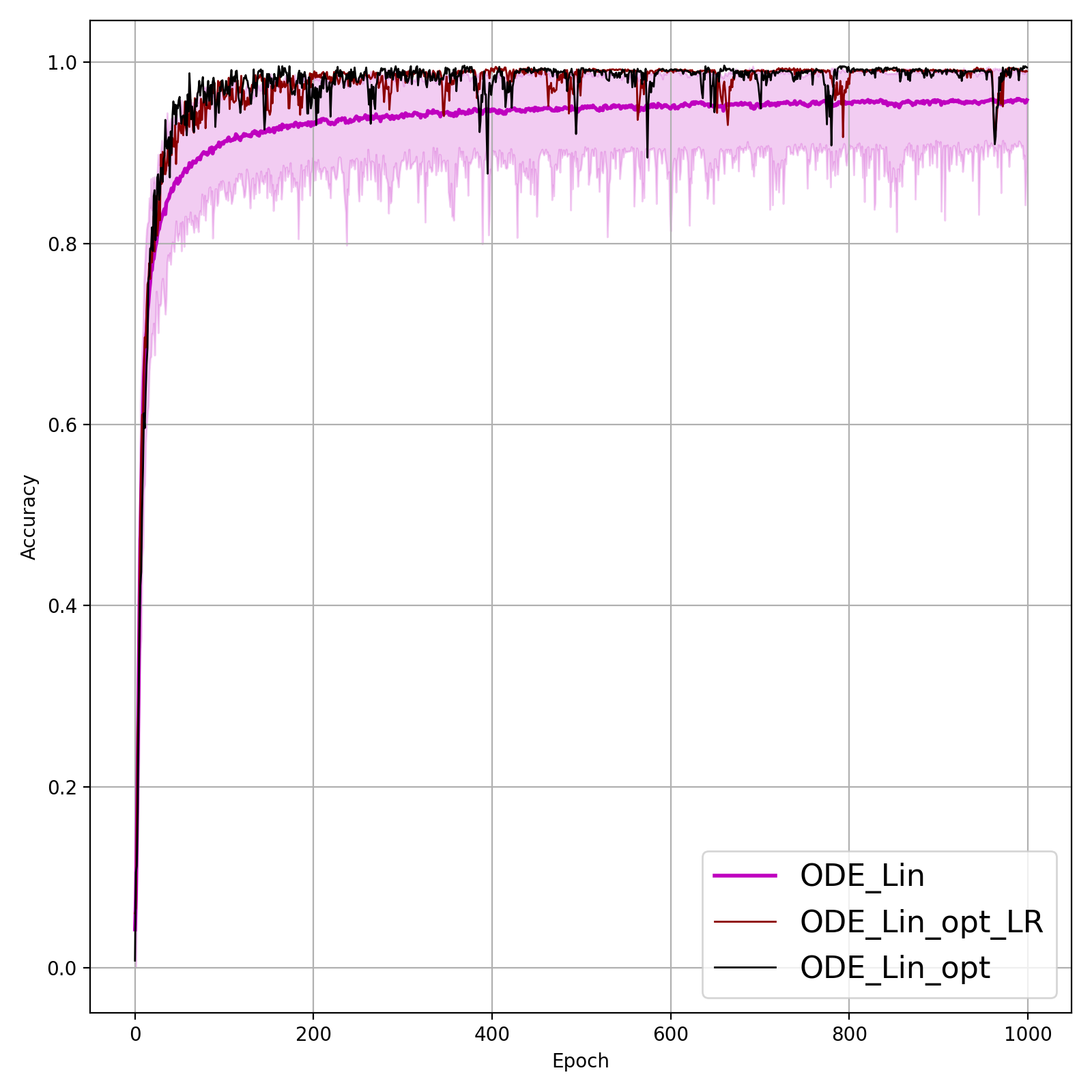}}
	\subfigure[\ Linear ODE 20\% observability]{
		\includegraphics[scale=0.2]{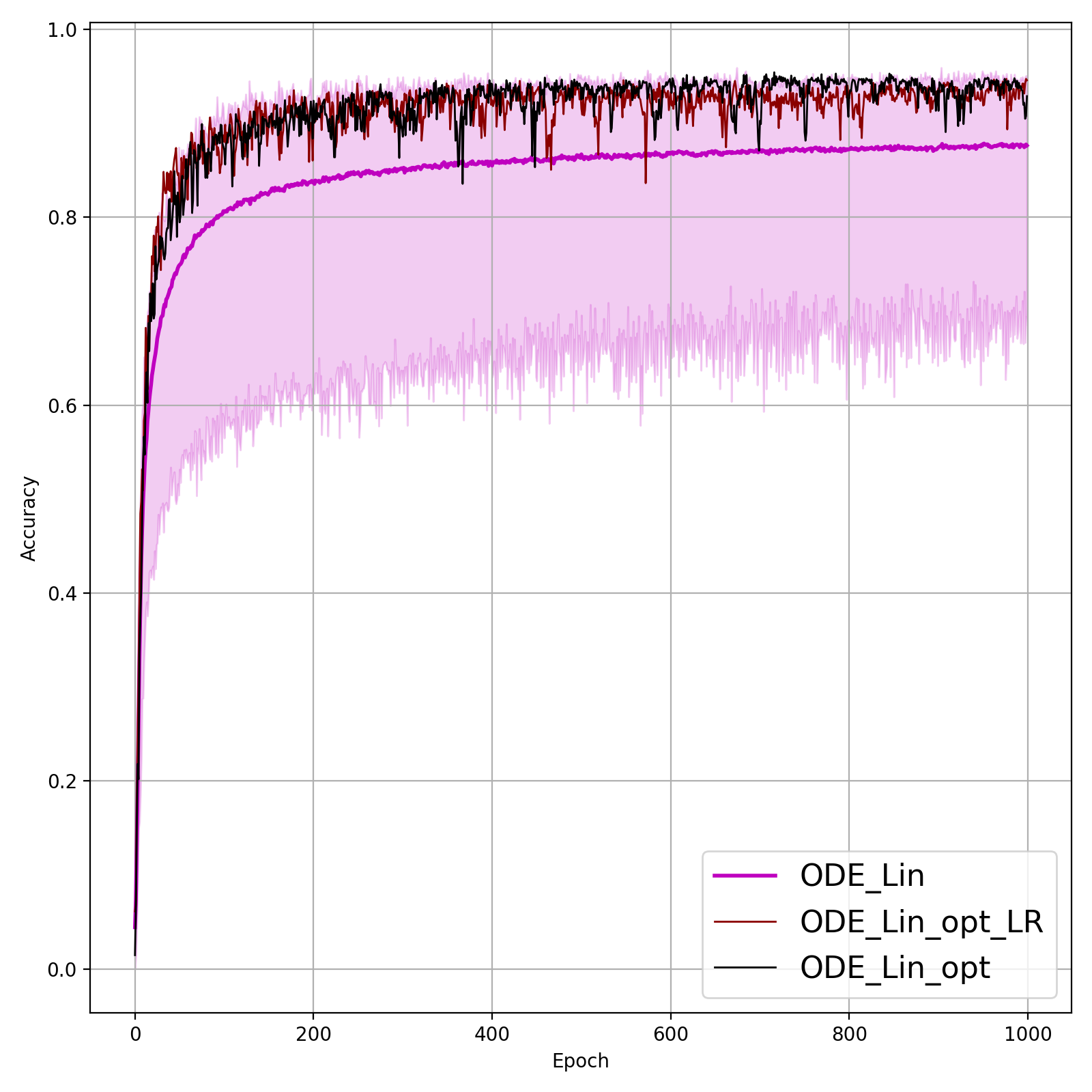}}
	\subfigure[\ Linear ODE 10\% observability]{
		\includegraphics[scale=0.2]{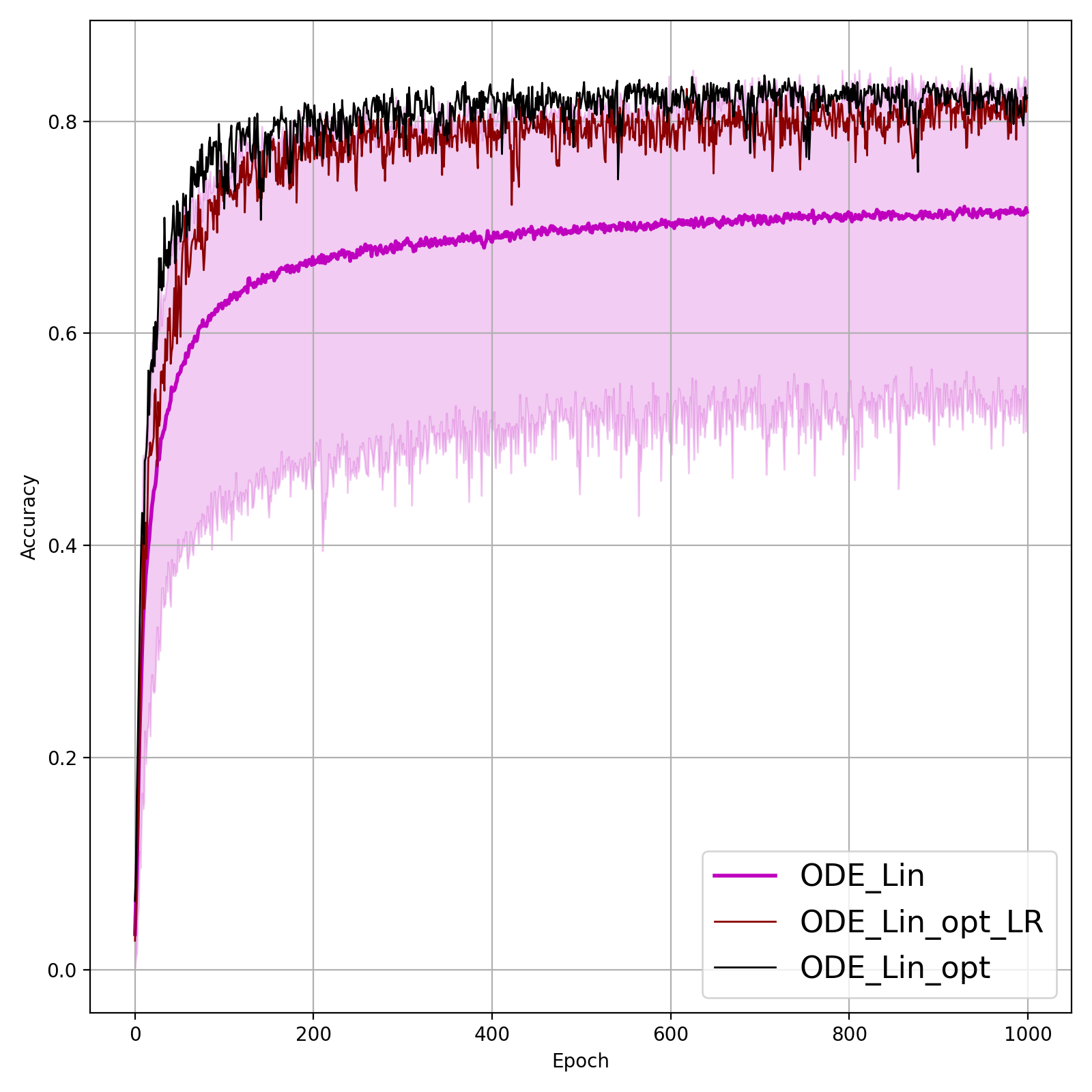}}
	\subfigure[\ Linear ODE 5\% observability]{
		\includegraphics[scale=0.2]{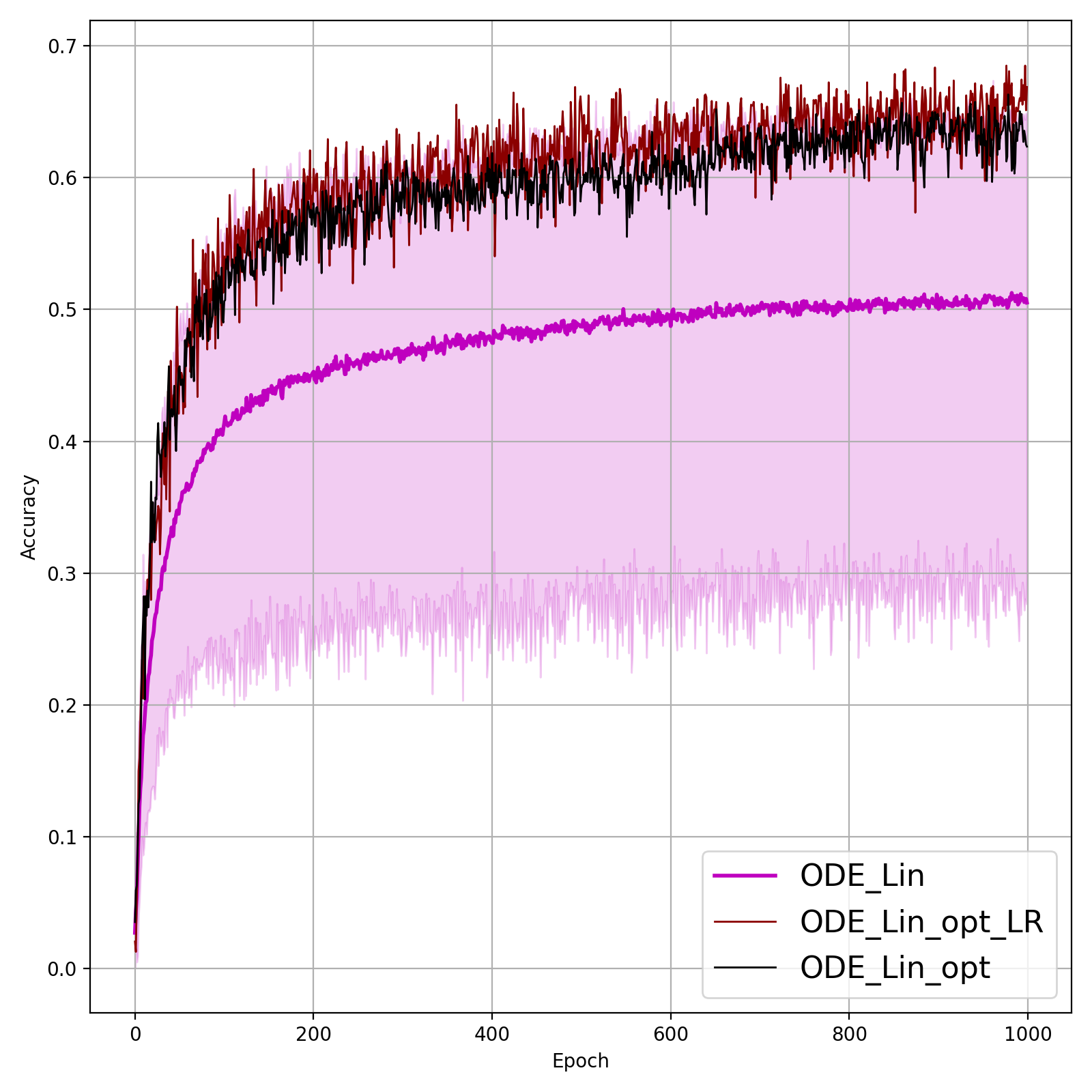}}
	\caption{Comparison of the placement learning with Linear ODE model for different observability. \label{fig:placement-ODE_Lin}}
\end{figure}

\begin{figure}
	\subfigure[\ Graph ODE 70\% observability]{
		\includegraphics[scale=0.2]{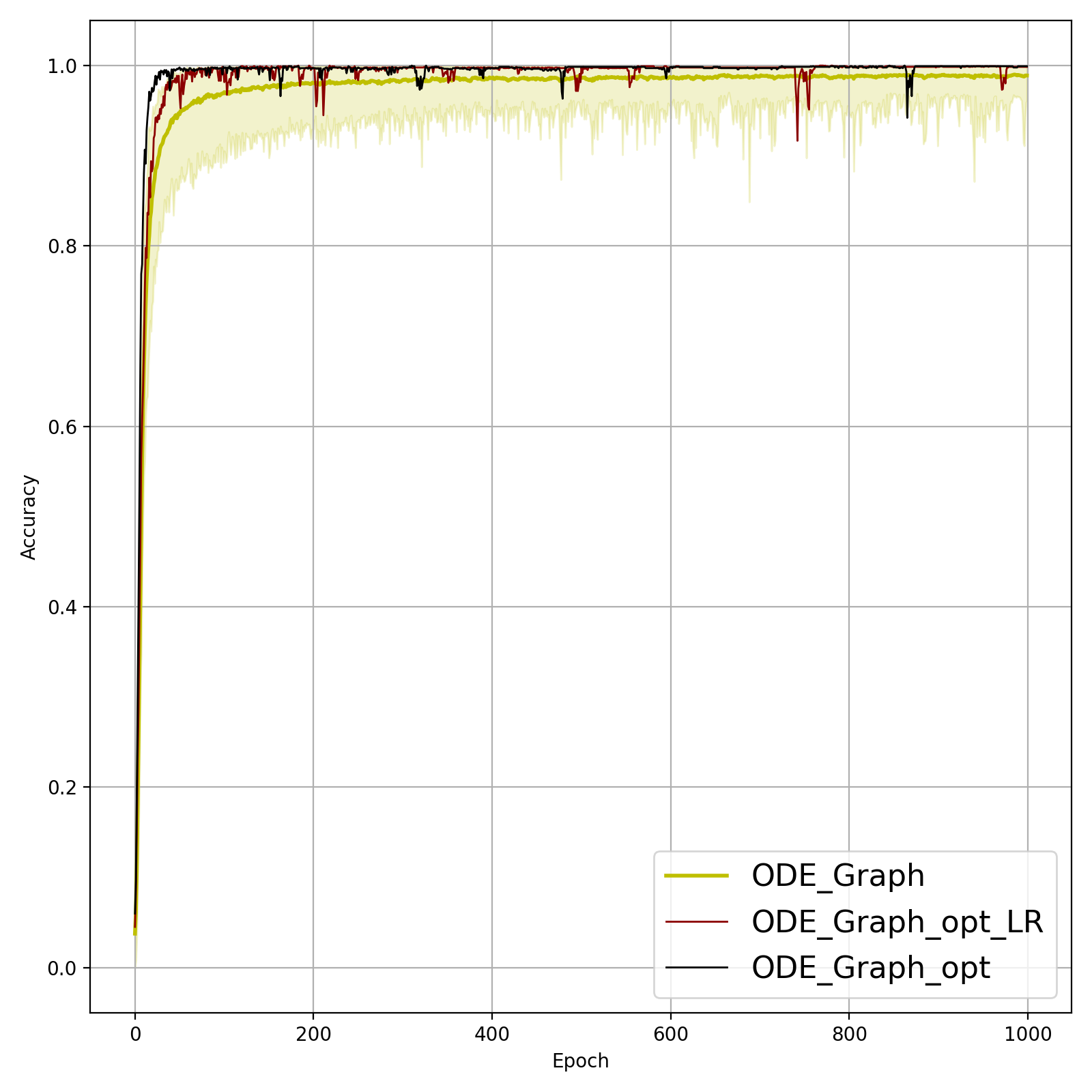}}
	\subfigure[\ Graph ODE 40\% observability]{
		\includegraphics[scale=0.2]{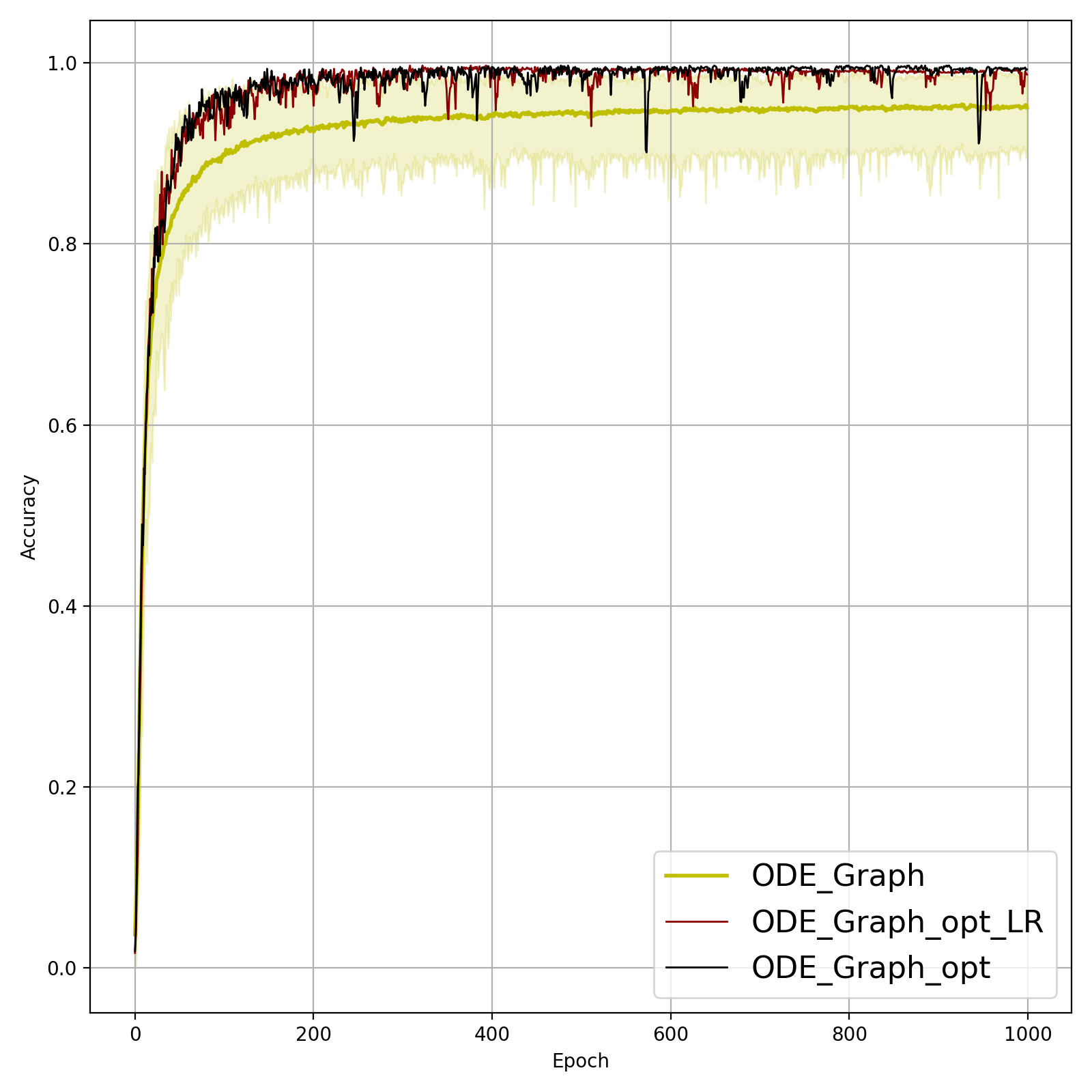}}
	\subfigure[\ Graph ODE 20\% observability]{
		\includegraphics[scale=0.2]{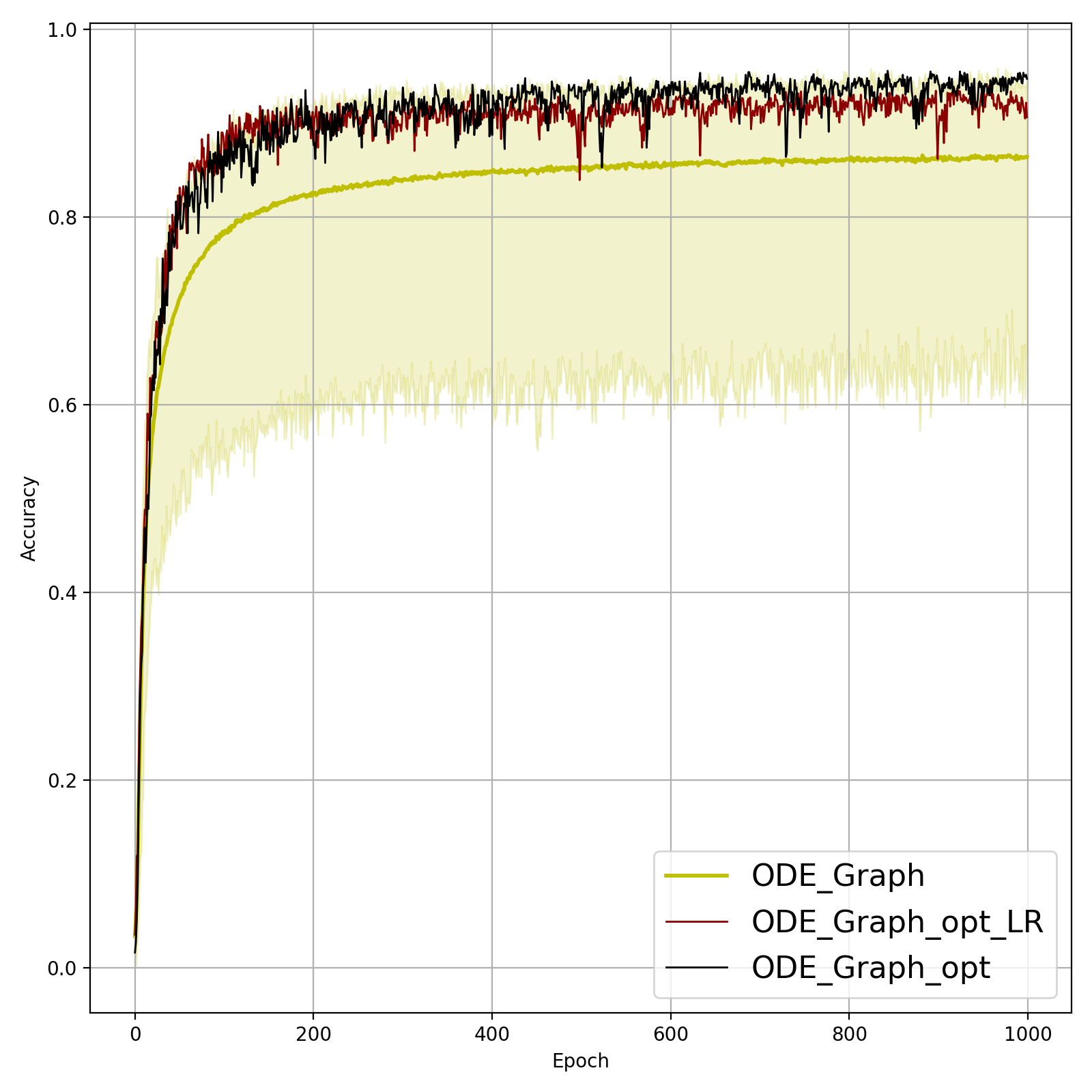}}
	\subfigure[\ Graph ODE 10\% observability]{
		\includegraphics[scale=0.2]{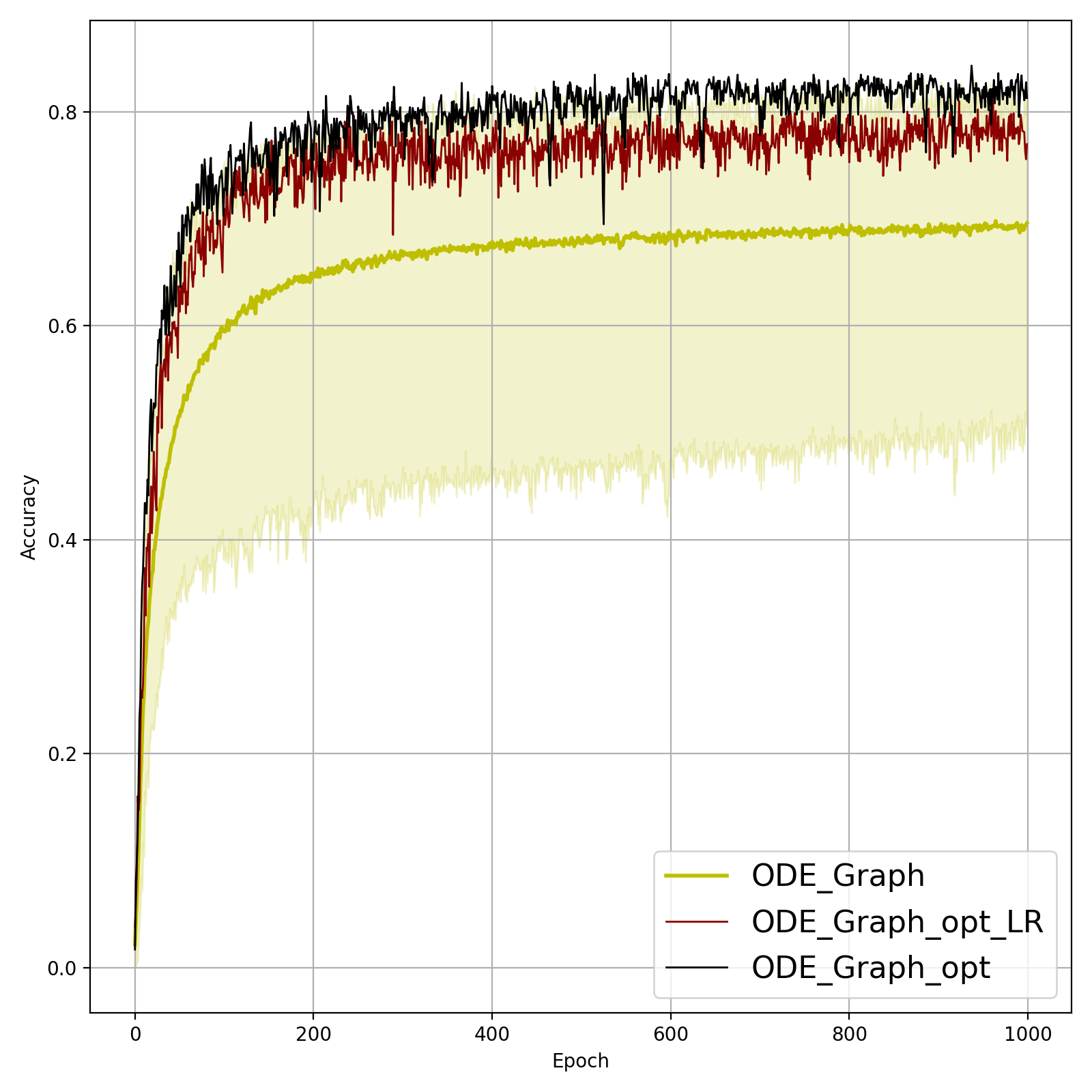}}
	\subfigure[\ Graph ODE 5\% observability]{
		\includegraphics[scale=0.2]{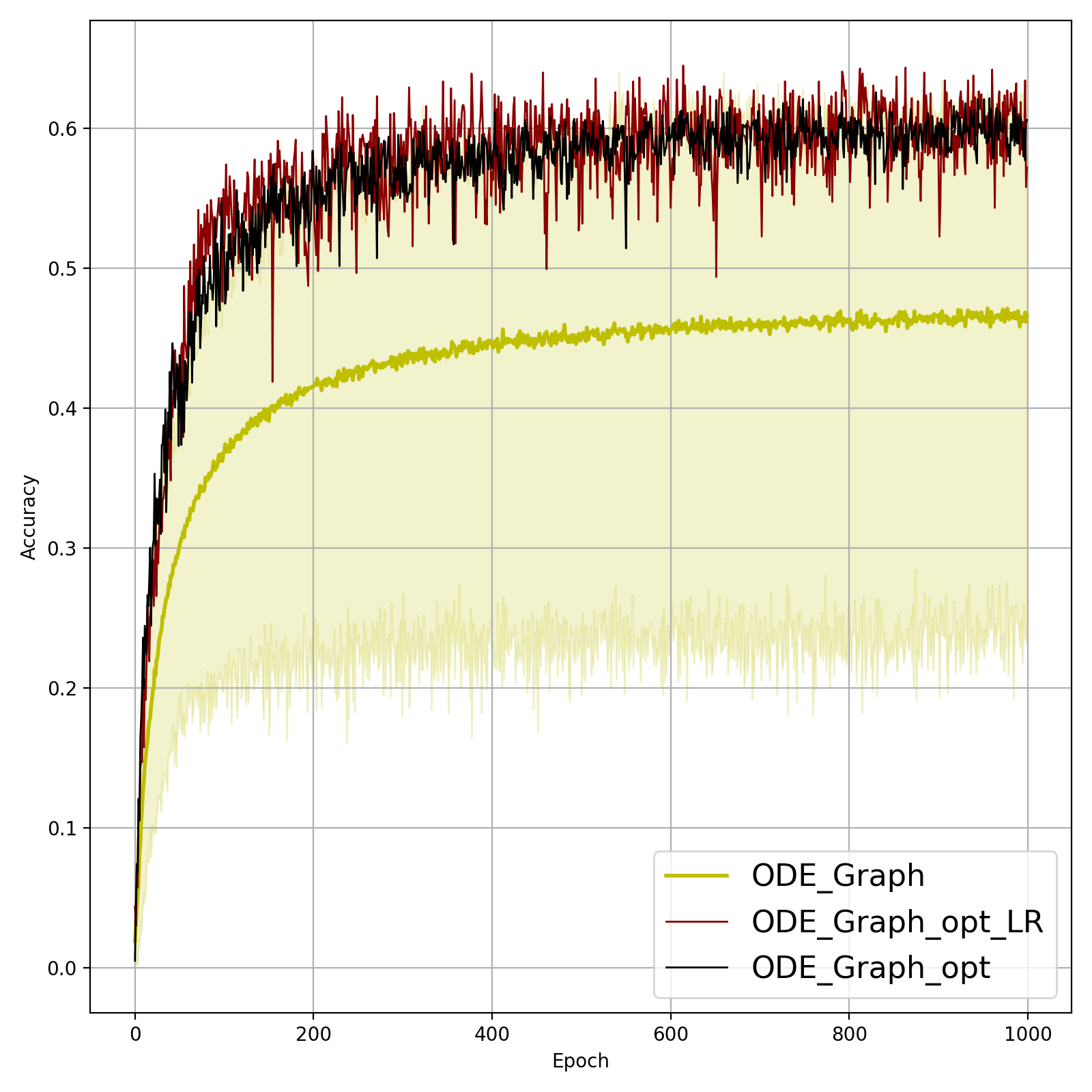}}
	\caption{Comparison of the placement learning with Graph ODE model for different observability. \label{fig:placement-ODE_Graph}}
\end{figure}

Results of the OP experiments are shown in Figures \ref{fig:placement-LR},\ref{fig:placement-FFNN},\ref{fig:placement-GCNN},\ref{fig:placement-AlexNet},\ref{fig:placement-ODE_Lin},\ref{fig:placement-ODE_Graph} for the LR, FFNN, GCNN, AlexNet, ODE Lin and ODE Graph, respectively, each showing in sub-figures performance under 100\%, 70\%, 40\%, 20\%, 10\% and 5\% of nodes.  Each sub-figure in the set corresponding to the advanced (all but LR) methods show comparative performance  of (a) multiple IO-P samples, (b) OP-LR configuration found with LR-based training only, and (c) OP configuration found with LR-based pre-training and follow-up training on corresponding model's data.

The experiments suggest that (a)  finding optimal placement improves performance of the fault detection dramatically; (b) optimal placement of PMU is a combinatorial optimization problem (of exponential complexity in the network size), which can be resolved efficiently (and, obviously, heuristically, i.e. approximately) with modern ML optimization software; (c) softening input and pre-training (with fast but not accurate LR method) are steps which are critical for making the Optimal Placement algorithms efficient.  

\appendices

\section{Static Failure Detection Experiments: Implementation Details} \label{appendix:exp1}

In this Appendix we provide details of how NN, tested in the static failure detection experiments, are constructed. We will be showing it for the exemplary IEEE bus-system we work with and also show architectures only for the case of full observability.  (Modification for the case of partial observability is straightforward --- only input signal should be adjusted to the respective lower dimensional signal.) We denote the input as $x\in \mathbb{R}^{|\mathcal{V}|}$,  where $|\mathcal{V}|=68$,  and the output as, $y\in \mathbb{R}^{|\mathcal{E}|}$, where $|\mathcal{E}|=87$ . Then architectures of the networks tested in our experiments are as follows: 
\begin{align*}
    &\text{LR}:\ x \rightarrow \text{Lin}(68, 87) \rightarrow y\\
    &\text{FFNN}:\ x\rightarrow \text{Lin}(68, 32) \rightarrow ReLU \rightarrow \text{Lin}(32, 87) \rightarrow y\\
    &\text{GCNN}: x \!\rightarrow\! \text{GraphConv}(68, 32) \!\rightarrow\! \text{ReLU} \!\rightarrow\! \text{Lin}(32, 87) \!\rightarrow\! y\\
    &\text{ODENN}\rightarrow \text{ODEBlock}(68, 68) \rightarrow \text{Lin}(68, 87) \rightarrow y
\end{align*}
where ODEBlock = [Linear(input size, output size)] followed by ReLU in the case of the LinODE and ODEBlock = [GraphConv(input size, output size) followed by ReLU in the case of the GraphODE. 

Table \ref{table:AlexNet} shows, mainly for self-sufficiency of the text, architecture of the AlexNet \cite{AlexNet}. 
\begin{table}[h!]
	\centering
	\begin{tabular}{ | c | }
		\hline
		Conv(inp. size =68, inp. chan.=1, out. chan.=4, ker. size=5, stride=1)\\
		\hline
		ReLU\\
		\hline
		MaxPool(ker. size=2, stride=2)\\
		\hline
		Conv(inp. chan.=4, out. chan.=8, ker. size=5, stride=1)\\
		\hline
		ReLU\\
		\hline
		MaxPool(ker. size=2, stride=2)\\
		\hline
		Conv(inp. chan.=8, out. chan.=8, ker. size=3, stride=1)\\
		\hline
		ReLU\\
		\hline
		MaxPool(ker. size=2, stride=2)\\
		\hline
		Conv(inp. chan.=8, out. chan.=8, ker. size=3, stride=1)\\
		\hline
		ReLU\\
		\hline
		MaxPool(ker. size=2, stride=2)\\
		\hline
		Linear(inp. size = 16, out. size = 87)\\
		\hline
	\end{tabular}
	\caption{AlexNet \cite{AlexNet} architecture. Abbreviations are inp.=input, out.=output, ker.=kernel, chan.=channels.}
	\label{table:AlexNet}
\end{table}

\section{Dynamic State Estimation Experiments: Implementation Details}\label{append:exp2}

Tables \ref{table:append_dynamic100},\ref{table:append_dynamic70},\ref{table:append_dynamic40},\ref{table:append_dynamic20},\ref{table:append_dynamic10},\ref{table:append_dynamic5} show details of the dynamic state estimation experiments under 100\%, 70\%, 40\%,20\% and 10\% of observability.

\begin{center}
	\begin{table}[h!]
		\centering
		\begin{tabular}{ |p{1.65cm}||p{1.5cm}|p{1.5cm}|p{1.5cm}|}
			\hline
			Model & Loss & Param. & CPU\\
			\hline
			LR   & -30.21  & 4,692 & 0.010 \\
			FFNN &  -28.16  & 4,452  & 0.012 \\
			GCNN  & -26.98 & 4,452 & 0.016 \\
			AlexNet  & -19.10 & 6,800 & 0.058 \\
			ODE Lin &  -34.30 & 4,692 & 0.301 \\
			ODE Graph & -34.98 & 4,692 & 0.378 \\
			PINN & -26.19  & 4,452 & 0.349  \\
			HNN & -36.39 & 37,469 & 1.860\\
			DIRODENN & -34.10  & 204 & 0.457 \\
			\hline
		\end{tabular}\\
		\caption{Comparison of the Dynamic State Estimation models  under 100\% observability. Loss is in decibels, otherwise nomenclature is the same as in Table \ref{table:failure_static}.}
		\label{table:append_dynamic100}
	\end{table}
	
	\begin{table}[h!]
		\centering
		\begin{tabular}{ |p{1.65cm}||p{1.5cm}|p{1.5cm}|p{1.5cm}|}
			\hline
			Model & Loss & Param. & CPU\\
			\hline
			LR   & -28.24  & 3,196 & 0.026\\
			FFNN &  -26.20  & 3,748 & 0.029\\
			GCNN  & -24.61 & 4,452 & 0.032\\
			ODE Lin &  -29.58 & 5,358 & 0.362\\
			ODE Graph & -30.02 & 6,370 & 0.564\\
			PINN & -27.63  & 3,748 & 0.376\\
			HNN & -32.34 & 29,899 & 2.140\\
			DIRODENN & -28.21  & 12,786 & 0.513\\
			\hline
		\end{tabular}\\
		\caption{Comparison of Dynamic State Estimation models  under 70\% observability. Loss is in decibels, otherwise nomenclature is the same as in Table \ref{table:failure_static}.}
		\label{table:append_dynamic70}
	\end{table}
	
	\begin{table}[h!]
		\centering
		\begin{tabular}{ |p{1.65cm}||p{1.5cm}|p{1.5cm}|p{1.5cm}|}
			\hline
			Model & Loss & Param. & CPU\\
			\hline
			LR   & -23.28  & 1,836 & 0.025\\
			FFNN &  -24.01  & 3,108 & 0.029\\
			GCNN  & -23.93 & 4,452 & 0.031\\
			ODE Lin &  -24.18 & 2,538 & 0.305\\
			ODE Graph & -24.29 & 3,630 & 0.456\\
			PINN & -23.52  & 3,108 & 0.381 \\
			HNN & -25.18 & 12,799 & 1.626\\
			DIRODENN & -23.52  & 7,286 & 0.412\\
			\hline
		\end{tabular}\\
		\caption{Comparison of Dynamic State Estimation models  under 40\%  observability. Loss is in decibels, otherwise nomenclature is the same as in Table \ref{table:failure_static}.}
		\label{table:append_dynamic40}
	\end{table}
	
	\begin{table}[h!]
		\centering
		\begin{tabular}{ |p{1.65cm}||p{1.5cm}|p{1.5cm}|p{1.5cm}|}
			\hline
			Model & Loss & Param. & CPU\\
			\hline
			LR   & -21.41  & 952 & 0.024\\
			FFNN &  -22.39  & 2,692 & 0.035\\
			GCNN  & -22.75 & 4,452 & 0.032\\
			ODE Lin &  -23.05 & 1,134 & 0.268\\
			ODE Graph & -23.11 & 1,849 & 0.378\\
			PINN & -22.10  & 2,692 & 0.381\\
			HNN & -21.99 & 5,116 & 1.427\\
			DIRODENN & -22.54  & 3,711 & 0.356\\
			\hline
		\end{tabular}\\
		\caption{Comparison of Dynamic State Estimation models  under 20\% observability. Loss is in decibels, otherwise nomenclature is the same as in Table \ref{table:failure_static}.}
		\label{table:append_dynamic20}
	\end{table}
	
	\begin{table}[h!]
		\centering
		\begin{tabular}{ |p{1.65cm}||p{1.5cm}|p{1.5cm}|p{1.5cm}|}
			\hline
			Model & Loss & Param. & CPU\\
			\hline
			LR   & -20.82  & 476 & 0.023\\
			FFNN &  -21.30  & 2,468  & 0.027\\
			GCNN  & -20.82 & 4,452 & 0.031\\
			ODE Lin &  -22.24 & 518 & 0.202\\
			ODE Graph & -22.69 & 890 & 0.242\\
			PINN & -21.77  & 2,468 & 0.398 \\
			HNN & -20.23 & 2,099 & 1.369\\
			DIRODENN & -22.42  & 1,786 & 0.330\\
			\hline
		\end{tabular}\\
		\caption{Comparison of Dynamic State Estimation models  under 10\% observability. Loss is in decibels, otherwise nomenclature is the same as in Table \ref{table:failure_static}.}
		\label{table:append_dynamic10}
	\end{table}
	
	\begin{table}[h!]
		\centering
		\begin{tabular}{ |p{1.65cm}||p{1.5cm}|p{1.5cm}|p{1.5cm}|}
			\hline
			Model & Loss & Param. & CPU\\
			\hline
			LR   & -18.92  & 272 & 0.023\\
			FFNN &  -21.24  & 2,372  & 0.027\\
			GCNN  & -20.40 & 2,372 & 0.029\\
			ODE Lin &  -20.06 & 479 & 0.181\\
			ODE Graph & -20.03 & 479 & 0.228\\
			PINN & -22.09  & 2,372 & 0.404 \\
			HNN & -19.30 & 1,046 & 1.268\\
			DIRODENN & -20.14  & 961 & 0.314\\
			\hline
		\end{tabular}\\
		\caption{Comparison of Dynamic State Estimation models  under 5\% observability. Loss is in decibels, otherwise nomenclature is the same as in Table \ref{table:failure_static}.}
		\label{table:append_dynamic5}
	\end{table}
\end{center}

\begin{table}[h!]
	\centering
	\begin{tabular}{|p{1.5cm}||p{1.1cm}|p{1.1cm}|p{1.1cm}|}
		\hline
		Model & 100\% & 70\% & 40\% 
		\\
		\hline
		LR   & 1e-3|3e-7  & 1e-3|3e-7 & 1e-3|3e-7 
		\\
		FFNN & 1e-2|1e-6  & 1e-2|1e-6  & 1e-2|1e-7 
		\\
		GCNN  & 1e-3|5e-8 & 5e-3|5e-8 & 5e-3|5e-9 
		\\
		AlexNet  & 1e-3|3e-7 & -|- & -|- 
		\\
		ODE Lin &  1e-2|1e-8 & 1e-2|1e-8 & 1e-2|1e-8 
		\\
		ODE Graph &  2e-2|3e-9 & 2e-2|3e-9 & 3e-2|3e-9 
		\\
		PINN & 1e-2|3e-9  & 1e-2|3e-8 & 5e-3|3e-8 
		\\
		HNN & 1e-2|0 & 3e-3|0 & 3e-3|0 
		\\
		DIRODENN & 5e-3|1e-8  & 5e-3|1e-8 & 1e-2|1e-8 
		\\
		\hline
	\end{tabular}
	\begin{tabular}{|p{1.5cm}||p{1.1cm}|p{1.1cm}|p{1.1cm}|}
		\hline
		Model 
		& 20\% & 10\% & 5\% 
		\\
		\hline
		LR   
		& 1e-3|3e-7 & 1e-3|3e-7 & 1e-3|3e-7
		\\
		FFNN 
		& 2e-2|5e-7 & 1e-2|5e-8 & 1e-2|5e-8
		\\
		GCNN  
		& 1e-2|5e-6 & 1e-2|3e-6 & 5e-2|5e-8
		\\
		AlexNet  
		& -|- & -|- & -|-
		\\
		ODE Lin 
		& 5e-2|5e-8 & 5e-2|5e-8 & 5e-2|5e-8
		\\
		ODE Graph 
		& 5e-2|5e-9 & 5e-2|5e-9 & 2e-2|0
		\\
		PINN 
		& 5e-3|8e-5 & 5e-3|8e-5 & 5e-3|8e-5 
		\\
		HNN 
		& 3e-3|0 & 5e-3|0 & 1e-2|0 
		\\
		DIRODENN 
		& 5e-2|1e-8 & 5e-2|1e-8 & 5e-2|1e-8
		\\
		\hline
	\end{tabular}
	\caption{Summary of the optimal hyper-parameters (optimal rate | value of the $l_2$ regularization) in the dynamic state estimation experiments. }
	\label{table:append_dynamic_opt_params}
\end{table}

\section{Optimal PMU Placement Experiments: Implementation Details}\label{append:exp3}

Architecture of the Optimal Placement NN, detailing the setting in Fig.~\ref{fig:OP-NN}, is shown in Table \ref{table:OP-NN}.

\begin{table}[h!]
	\centering
	\textbf{Optimal Placement Neural Network}:\\
	\begin{tabular}{ | c | } 
		\hline
		GraphConv layer(68, 16)\\
		\hline
		ReLU\\
		\hline
		FF layer(16, 16)\\
		\hline
		ReLU\\
		\hline
		FF layer(16, 16)\\
		\hline
		ReLU\\
		\hline
		FF layer(16, 6)\\
		\hline
		Sigmoid\\
		\hline
	\end{tabular}
	\caption{Additional details on the architecture of the Optimal Placement Neural Network (OP-NN) from Fig.~(\ref{fig:OP-NN}). 
	\label{table:OP-NN}}
\end{table}

\section{Description of Software and Hardware}
All the experiments were implemented on Python using Pytorch \cite{PyTorch} on the Google Colab with Intel Xeon CPU @ 2.20GHz and 12GB  NVIDIA  Tesla K80 GPU.

\bibliographystyle{plain}
\bibliography{bib/DynamicSystemLearning.bib,bib/GM.bib,bib/NeuralPIV.bib,bib/PSLearning.bib}

\end{document}